%% file: main.tex
\documentclass[10pt]{article}
\usepackage[numbers]{natbib}
\usepackage{fullpage}

\input{defines}

\usepackage[titletoc,page]{appendix}

\usepackage{moreverb,url} 
\usepackage{secdot}
\sectiondot{subsection}    
\sectiondot{subsubsection}
 
\begin{document} 
 
% \runninghead{Ross Hartley et al.}

\title{Contact-Aided Invariant Extended Kalman Filtering for \\Robot State Estimation\thanks{Submitted for journal publication. \url{rosshart@umich.edu}}}

\author{Ross Hartley\thanks{Robotics Institute and College of Engineering, University of Michigan, Ann Arbor, MI, USA.} \and Maani Ghaffari\footnotemark[2] \and Ryan M. Eustice\footnotemark[2] \and Jessy W. Grizzle\footnotemark[2]}

\maketitle

\begin{abstract}
    Legged robots require knowledge of pose and velocity in order to maintain stability and execute walking paths. Current solutions either rely on vision data, which is susceptible to environmental and lighting conditions, or fusion of kinematic and contact data with measurements from an inertial measurement unit (IMU). In this work, we develop a contact-aided invariant extended Kalman filter (InEKF) using the theory of Lie groups and invariant observer design. This filter combines contact-inertial dynamics with forward kinematic corrections to estimate pose and velocity along with all current contact points. We show that the error dynamics follows a log-linear autonomous differential equation with several important consequences: (a) the observable state variables can be rendered convergent with a domain of attraction that is independent of the system's trajectory; (b) unlike the standard EKF, neither the linearized error dynamics nor the linearized observation model depend on the current state estimate, which (c) leads to improved convergence properties and (d) a local observability matrix that is consistent with the underlying nonlinear system. Furthermore, we demonstrate how to include IMU biases, add/remove contacts, and formulate both world-centric and robo-centric versions. We compare the convergence of the proposed InEKF with the commonly used quaternion-based EKF though both simulations and experiments on a Cassie-series bipedal robot. Filter accuracy is analyzed using motion capture, while a LiDAR mapping experiment provides a practical use case. Overall, the developed contact-aided InEKF provides better performance in comparison with the quaternion-based EKF as a result of exploiting symmetries present in system.
\end{abstract}

% \keywords{Invariant estimation, invariant observer, Kalman filter, legged robots, inertial measurement unit, contact, contact-aided navigation, state estimation, SLAM, Cassie robot, matrix Lie groups, Lie algebra, exponential coordinates.}

%% Introduction
\input{intro}
  
%% Literature review
\input{literature}

%% Preliminaries
\input{preliminaries}

%% Motivating example; attitude estimation
\input{attitude}

%% Continuous Right-Invariant EKF
\input{riekf}

%% Simulation Results
\input{simulation_results}

%% IMU bias augmentation
\input{bias}

%% Switching contact points and state augmentation
\input{switch}

%% Experimental Results
\input{experimental_results}

%% Continuous Left-Invariant EKF
\input{liekf}

% Robo-centric right/left invariant EKF and connection to the original
\input{robocentric}

%% Other types of measurements
\input{additional_measurements}
\input{conclusion}

\input{appendix}

\section*{Acknowledgments}
The authors would like to thank Yukai Gong for the development of the feedback controller utilized in the Cassie experiments as well as Bruce Huang, Ray Zhang, Lu Gan, Zhenyu Gan, Omar Harib, Eva Mungai, and Grant Gibson for their help in collecting experimental data. Funding for R. Hartley and M. Ghaffari is given by the Toyota Research Institute (TRI), partly under award number N021515, however this article solely reflects the opinions and conclusions of its authors and not TRI or any other Toyota entity. Funding for J. Grizzle was in part provided by TRI and in part by NSF Award No.~1808051.

% \begin{acks}
% This class file was developed by Sunrise Setting Ltd,
% Paignton, Devon, UK. Website:\\
% \url{http://www.sunrise-setting.co.uk}
% \end{acks}

% Definition of any abbreviations used.
% \section{Acronyms}
\begin{acronym}
    \acro{IMU}{inertial measurement unit}
    \acro{EKF}{extended Kalman filter}
    \acro{ErEKF}{error-state extended Kalman filter}
    \acro{QEKF}{quaternion-based extended Kalman filter}
    \acro{InEKF}{invariant extended Kalman filter}
    \acro{RIEKF}{right-invariant extended Kalman filter}
    \acro{LIEKF}{lef- invariant extended Kalman filter}
    \acro{INS}{inertial navigation system}
    \acro{FK}{forward kinematic}
    \acro{DOF}{degrees of freedom}   
    \acro{SLAM}{simultaneous localization and mapping}
    \acro{ICP}{Iterative Closest Point}
\end{acronym}

\bibliographystyle{plainnat}
\bibliography{references}

\end{document}

%% file: defines.tex
\usepackage{graphicx}
\usepackage{caption}
\graphicspath{
    {graphics/}
}

\usepackage{amsmath,amssymb,enumerate}
\usepackage{amsthm}
\usepackage{setspace}
\usepackage{booktabs}
\usepackage[usenames,dvipsnames,svgnames,table]{xcolor}
\usepackage[bookmarks=true,colorlinks=true,pdfpagemode=UseNone,citecolor=black,linkcolor=black,urlcolor=BrickRed,pagebackref]{hyperref}
\usepackage{mathtools}
\usepackage{algorithm, algorithmicx, algpseudocode}
\usepackage{blindtext}
\usepackage{gensymb}
\usepackage{xparse}
\usepackage{lipsum}
\usepackage{graphicx}
\usepackage{subcaption}
\usepackage{soul} 
\usepackage{mathrsfs}
\usepackage[mathscr]{euscript}
\usepackage{times}

\usepackage{multicol}

\usepackage{amsfonts}
\usepackage{fixltx2e}
\usepackage{cleveref}
\usepackage[utf8]{inputenc}
\usepackage[T1]{fontenc}
\usepackage{textcomp}
\usepackage[nolist]{acronym}
\usepackage{tabularx}
\usepackage{adjustbox}
\usepackage{textcomp}
\usepackage{placeins}
\usepackage{afterpage}

\newtheorem{theorem}{Theorem}

\theoremstyle{definition}
\newtheorem{definition}{Definition}

\newtheorem{remark}{Remark}

\def\realnumbers{\mathbb{R}}

\definecolor{note}{rgb}{0.1,0.1,1}
\definecolor{rephase}{rgb}{0.15,0.7,0.15}
\definecolor{bag}{rgb}{0.6,0.6,0.2}

\newcommand\redefine{\mathrel{\overset{\makebox[0pt]{\mbox{\normalfont\tiny\sffamily redefine}}}{\triangleq}}}

\newcommand{\m}{\mathop{\mathrm{m}}}
\newcommand{\rad}{\mathop{\mathrm{rad}}}

\newcommand{\Hz}{\mathop{\mathrm{Hz}}}

\newcommand{\transpose}{\mathsf{T}}
\newcommand{\SO}{\mathrm{SO}}
\newcommand{\SE}{\mathrm{SE}}
\newcommand{\SEK}{\mathrm{SE}_K}

\newcommand{\sek}{\mathfrak{se}_K}

\newcommand{\deriv}{\dfrac{\mathrm{d}}{\mathrm{d}t}}
\newcommand{\dimension}{\mathrm{dim}}

\DeclareDocumentCommand{\stateTransition}{ O{} O{} }{\boldsymbol{\Phi}_{#1}^{#2}}

\DeclareDocumentCommand{\vectorToSkew}{ O{} }{\left(#1\right)_\times}
\DeclareDocumentCommand{\Gam}{ O{} O{} }{\boldsymbol{\Gamma}_{#1}^{#2}}

\DeclareDocumentCommand{\vector}{ O{} }{\mathrm{vec}(#1)} 
\DeclareDocumentCommand{\zeros}{ O{} }{\textbf{0}_{#1}}
\DeclareDocumentCommand{\X}{ O{} O{} }{\textbf{X}_{#1}^{#2}}
\DeclareDocumentCommand{\XE}{ O{} O{} }{\bar{\textbf{X}}_{#1}^{#2}}
\DeclareDocumentCommand{\U}{ O{} O{} }{\textbf{U}_{#1}^{#2}}
\DeclareDocumentCommand{\UE}{ O{} O{} }{\bar{\textbf{U}}_{#1}^{#2}}
\DeclareDocumentCommand{\E}{ O{} O{} }{\textbf{E}_{#1}^{#2}}
\DeclareDocumentCommand{\M}{ O{} O{} }{\textbf{M}_{#1}^{#2}}
\DeclareDocumentCommand{\L}{ O{} O{} }{\textbf{L}_{#1}^{#2}}
\DeclareDocumentCommand{\S}{ O{} O{} }{\textbf{S}_{#1}^{#2}}
\DeclareDocumentCommand{\P}{ O{} O{} }{\textbf{P}_{#1}^{#2}}
\DeclareDocumentCommand{\B}{ O{} O{} }{\textbf{B}_{#1}^{#2}}
\DeclareDocumentCommand{\Q}{ O{} O{} }{\textbf{Q}_{#1}^{#2}}
\DeclareDocumentCommand{\H}{ O{} O{} }{\textbf{H}_{#1}^{#2}}
\DeclareDocumentCommand{\J}{ O{} O{} }{\textbf{J}_{#1}^{#2}}
\DeclareDocumentCommand{\R}{ O{} O{} }{\textbf{R}_{#1}^{#2}}
\DeclareDocumentCommand{\RE}{ O{} O{} }{\bar{\textbf{R}}_{#1}^{#2}}
\DeclareDocumentCommand{\Q}{ O{} O{} }{\textbf{Q}_{#1}^{#2}}
\DeclareDocumentCommand{\T}{ O{} O{} }{\textbf{T}_{#1}^{#2}}
\DeclareDocumentCommand{\L}{ O{} O{} }{\textbf{L}_{#1}^{#2}}
\DeclareDocumentCommand{\K}{ O{} O{} }{\textbf{K}_{#1}^{#2}}
\DeclareDocumentCommand{\V}{ O{} O{} }{\textbf{V}_{#1}^{#2}}
\DeclareDocumentCommand{\N}{ O{} O{} }{\textbf{N}_{#1}^{#2}}
\DeclareDocumentCommand{\Y}{ O{} O{} }{\textbf{Y}_{#1}^{#2}}
\DeclareDocumentCommand{\Z}{ O{} O{} }{\textbf{Z}_{#1}^{#2}}
\DeclareDocumentCommand{\F}{ O{} O{} }{\textbf{F}_{#1}^{#2}}
\DeclareDocumentCommand{\G}{ O{} O{} }{\textbf{G}_{#1}^{#2}}
\DeclareDocumentCommand{\A}{ O{} O{} }{\textbf{A}_{#1}^{#2}}
\DeclareDocumentCommand{\TH}{ O{} O{} }{\boldsymbol{\Theta}_{#1}^{#2}}
\DeclareDocumentCommand{\x}{ O{} O{} }{\textbf{x}_{#1}^{#2}}
\DeclareDocumentCommand{\e}{ O{} O{} }{\textbf{e}_{#1}^{#2}}
\DeclareDocumentCommand{\c}{ O{} O{} }{\textbf{c}_{#1}^{#2}}
\DeclareDocumentCommand{\C}{ O{} O{} }{\textbf{C}_{#1}^{#2}}
\DeclareDocumentCommand{\CM}{ O{} O{} }{\tilde{\textbf{C}}_{#1}^{#2}}
\DeclareDocumentCommand{\RM}{ O{} O{} }{\tilde{\textbf{R}}_{#1}^{#2}}
\DeclareDocumentCommand{\I}{ O{} O{} }{\textbf{I}_{#1}^{#2}}
\DeclareDocumentCommand{\O}{ O{} O{} }{\textbf{O}_{#1}^{#2}}
\DeclareDocumentCommand{\r}{ O{} O{} }{\textbf{r}_{#1}^{#2}}
\DeclareDocumentCommand{\t}{ O{} O{} }{\textbf{t}_{#1}^{#2}}
\DeclareDocumentCommand{\u}{ O{} O{} }{\textbf{u}_{#1}^{#2}}
\DeclareDocumentCommand{\d}{ O{} O{} }{\textbf{d}_{#1}^{#2}}
\DeclareDocumentCommand{\dE}{ O{} O{} }{\bar{\textbf{d}}_{#1}^{#2}}
\DeclareDocumentCommand{\b}{ O{} O{} }{\textbf{b}_{#1}^{#2}}
\DeclareDocumentCommand{\a}{ O{} O{} }{\textbf{a}_{#1}^{#2}}
\DeclareDocumentCommand{\g}{ O{} O{} }{\textbf{g}_{#1}^{#2}}
\DeclareDocumentCommand{\l}{ O{} O{} }{\textbf{l}_{#1}^{#2}}
\DeclareDocumentCommand{\z}{ O{} O{} }{\textbf{z}_{#1}^{#2}}
\DeclareDocumentCommand{\dM}{ O{} O{} }{\tilde{\textbf{d}}_{#1}^{#2}}
\DeclareDocumentCommand{\SelectionMatrix}{ O{} O{} }{\boldsymbol{\Pi}_{#1}^{#2}}
\DeclareDocumentCommand{\params}{ O{} O{} }{\boldsymbol{\theta}_{#1}^{#2}}
\DeclareDocumentCommand{\paramsE}{ O{} O{} }{\bar{\boldsymbol{\theta}}_{#1}^{#2}}
\DeclareDocumentCommand{\paramError}{ O{} O{} }{\boldsymbol{\zeta}_{#1}^{#2}}
\DeclareDocumentCommand{\gyroscopeBias}{ O{} }{\textbf{b}_{#1}^{g}}
\DeclareDocumentCommand{\gyroscopeBiasE}{ O{} }{\bar{\textbf{b}}_{#1}^{g}}
\DeclareDocumentCommand{\accelerometerBias}{ O{} }{\textbf{b}_{#1}^{a}}
\DeclareDocumentCommand{\accelerometerBiasE}{ O{} }{\bar{\textbf{b}}_{#1}^{a}}
\DeclareDocumentCommand{\position}{ O{} O{} }{{}_\text{#2}\textbf{p}_{\text{#1}}}
\DeclareDocumentCommand{\positionDot}{ O{} O{} }{{}_\text{#2}\dot{\textbf{p}}_{\text{#1}}}
\DeclareDocumentCommand{\linearVelocity}{ O{} O{} }{{}_\text{#2}\textbf{v}_{\text{#1}}}
\DeclareDocumentCommand{\linearVelocityM}{ O{} O{} }{{}_\text{#2}\tilde{\textbf{v}}_{\text{#1}}}
\DeclareDocumentCommand{\orientation}{ O{} }{\textbf{R}_{\text{#1}}}
\DeclareDocumentCommand{\orientationDot}{ O{} }{\dot{\textbf{R}}_{\text{#1}}}
\DeclareDocumentCommand{\angularVelocity}{ O{} O{} }{{}_\text{#2}\boldsymbol{\omega}_{\text{#1}}}
\DeclareDocumentCommand{\angularVelocityM}{ O{} O{} }{{}_\text{#2}\tilde{\boldsymbol{\omega}}_{\text{#1}}}
\DeclareDocumentCommand{\acceleration}{ O{} O{} }{{}_\text{#2}\textbf{a}_{\text{#1}}}
\DeclareDocumentCommand{\accelerationM}{ O{} O{} }{{}_\text{#2}\tilde{\textbf{a}}_{\text{#1}}}
\DeclareDocumentCommand{\p}{ O{} O{} }{\textbf{p}_{#1}^{#2}}
\DeclareDocumentCommand{\pE}{ O{} O{} }{\bar{\textbf{p}}_{#1}^{#2}}
\DeclareDocumentCommand{\pM}{ O{} O{} }{\tilde{\textbf{p}}_{#1}^{#2}}
\DeclareDocumentCommand{\l}{ O{} O{} }{\textbf{l}_{#1}^{#2}}
\DeclareDocumentCommand{\lE}{ O{} O{} }{\bar{\textbf{l}}_{#1}^{#2}}
\DeclareDocumentCommand{\v}{ O{} O{} }{\textbf{v}_{#1}^{#2}}
\DeclareDocumentCommand{\vE}{ O{} O{} }{\bar{\textbf{v}}_{#1}^{#2}}
\DeclareDocumentCommand{\w}{ O{} O{} }{\boldsymbol{\omega}_{#1}^{#2}}
\DeclareDocumentCommand{\wM}{ O{} O{} }{\tilde{\boldsymbol{\omega}}_{#1}^{#2}}
\DeclareDocumentCommand{\wB}{ O{} O{} }{\bar{\boldsymbol{\omega}}_{#1}^{#2}}
\DeclareDocumentCommand{\a}{ O{} O{} }{\textbf{a}_{#1}^{#2}}
\DeclareDocumentCommand{\aM}{ O{} O{} }{\tilde{\textbf{a}}_{#1}^{#2}}
\DeclareDocumentCommand{\aB}{ O{} O{} }{\bar{\textbf{a}}_{#1}^{#2}}
\DeclareDocumentCommand{\y}{ O{} O{} }{\textbf{y}_{#1}^{#2}}
\DeclareDocumentCommand{\noise}{ O{} O{} }{\textbf{w}_{#1}^{#2}}
\DeclareDocumentCommand{\FK}{ O{} }{\;\textit{\textbf{h}}_{#1}}
\DeclareDocumentCommand{\FKswitch}{ O{} }{\;\textit{\textbf{g}}_{#1}}
\DeclareDocumentCommand{\angleTheta}{ O{} O{} }{\boldsymbol{\theta}_{#1}^{#2}}
\DeclareDocumentCommand{\anglePhi}{ O{} O{} }{\boldsymbol{\phi}_{#1}^{#2}}
\DeclareDocumentCommand{\encoders}{ O{} O{} }{\boldsymbol{\alpha}_{#1}^{#2}}
\DeclareDocumentCommand{\encodersM}{ O{} O{} }{\tilde{\boldsymbol{\alpha}}_{#1}^{#2}}
\DeclareDocumentCommand{\offsets}{ O{} O{} }{\boldsymbol{\epsilon}_{#1}^{#2}}
\DeclareDocumentCommand{\Cov}{ O{} O{} }{\boldsymbol{\Sigma}_{#1}^{#2}}
\DeclareDocumentCommand{\Axis}{ O{} O{} }{\textnormal{Axis}_{#1}^{#2}}
\DeclareDocumentCommand{\using}{ O{} O{} }{\stackrel{\mathmakebox[\widthof{=}]{\text{eq.} (#1)}}{#2} \enspace}
\DeclareDocumentCommand{\Adjoint}{ O{} }{\mathrm{Ad}_{#1}}
\DeclareDocumentCommand{\vectorToAlgebra}{ O{} }{{#1}^\wedge}
\DeclareDocumentCommand{\algebraToVector}{ O{} }{{#1}^\vee}
\DeclareDocumentCommand{\groupError}{ O{} O{} }{\boldsymbol{\eta}_{#1}^{#2}}
\DeclareDocumentCommand{\tangentError}{ O{} O{} }{\boldsymbol{\xi}_{#1}^{#2}}
\DeclareDocumentCommand{\lieGroup}{ }{\mathcal{G}}
\DeclareDocumentCommand{\lieAlgebra}{ }{\mathfrak{g}}

\newcommand{\squeezeup}{\vspace{-3mm}}

\makeatletter
\newcommand{\mathleft}{\@fleqntrue\@mathmargin0pt}
\newcommand{\mathcenter}{\@fleqnfalse}
\makeatother

%% file: intro.tex
\section{Introduction} \label{sec:intro}
Legged robots have the potential to transform the logistics and package delivery industries, become assistants in our homes, and aide in search and rescue~\citep{cnbccassie}. In order to develop motion planning algorithms and robust feedback controllers for these tasks, accurate estimates of the robot's state are needed. Some states, such as joint angles, can be directly measured using encoders, while other states, such as the robot's pose and velocity, require additional sensors. Most legged robots are  equipped with an \ac{IMU} that can measure linear acceleration and angular velocity, albeit with noise and bias perturbations. Consequently, nonlinear observers are typically used to fuse leg odometry and inertial measurements to infer trajectory, velocity, and calibration parameters~\citep{rotella2014state,benallegue2015estimation,eljaik2015multimodal,kuindersma2016optimization}. In view of a practical solution, designing a globally convergent observer is sacrificed for one with at best local properties, such as the \ac{EKF}~\citep{grizzle1995extended,krener2003convergence,trawny2005indirect,sola2017quaternion}. This \ac{EKF}-based approach is computationally efficient and easily customizable, allowing successful implementation on a number of legged robots with rigorous real-time performance requirements~\citep{bloesch2012state,bloesch2017state,fallon2014drift,nobiliheterogeneous}. 

Accurate pose estimation can be combined with visual data to build maps of the environment~\citep{fankhauser2014robot}. Then such maps can be used in gait selection to improve the stability of a robot while walking on uneven terrains and as a basis for high-level motion planning. Although there have been many recent advancements in visual-inertial-odometry and \ac{SLAM} \citep{qin2018vins,huai2018robocentric,forster2016manifold}, these algorithms often rely on visual data for pose estimation. This means that the observer (and ultimately the feedback controller) can be adversely affected by rapid changes in lighting as well as the operating environment. It is therefore beneficial to develop a low-level state estimator that fuses data only from proprioceptive sensors to form accurate high-frequency state estimates. This approach was taken by \citet{bloesch2012state} when developing a \ac{QEKF} that combines inertial, contact, and kinematic data to estimate the robot's base pose, velocity, and a number of contact states. In this article, we expand upon these ideas to develop an \ac{InEKF} that has improved convergence and consistency properties allowing for a more robust observer that is suitable for long-term autonomy.

The theory of invariant observer design is based on the estimation error being invariant under the action of a matrix Lie group~\citep{aghannan2002invariant,bonnabel2009non}, which has recently led to the development of the \ac{InEKF}\footnote{We use the \ac{InEKF} acronym to distinguish from an iterated-\ac{EKF} (IEKF).}~\citep{bonnabel2007left,barrau2015non,barrau2017invariant,barrau2018invariant} with successful applications and promising results in simultaneous localization and mapping~\citep{barrau2015non,zhang2017convergence} and aided inertial navigation systems~\citep{barczyk2011invariant,barczyk2013invariant,barrau2015non,wu2017invariant}. The invariance of the estimation error with respect to a Lie group action is referred to as a symmetry of the system~\citep{barczyk2013invariant}. Summarized briefly, \citet{barrau2017invariant} showed that if the state is defined on a Lie group, and the dynamics satisfy a particular ``group affine'' property, then the symmetry leads to the estimation error satisfying a ``log-linear'' autonomous differential equation on the Lie algebra. In the deterministic case, this linear system can be used to exactly recover the estimated state of the nonlinear system as it evolves on the group. The log-linear property therefore allows the design of a nonlinear observer or state estimator with strong convergence properties. 

%%%%%%%%%%%%%%%%%%%%%%%%%%%%%%%%%%%%%%%%%%%%%%%%%%%%%%%%%%%%%%%%%%%%%%%%%%%%%%%%%%%%%%%%%%%%%%%%%%%
\subsection{Contribution}
In this article, we derive an \ac{InEKF} for a system containing \ac{IMU} and contact sensor dynamics, with forward kinematic correction measurements. We show that the defined deterministic system satisfies the ``group affine'' property, and therefore, the error dynamics is exactly log-linear. In practice, with the addition of sensor noise and \ac{IMU} bias, this log-linear error system is only approximate; we show, however, that in many cases the proposed \ac{InEKF} is still preferred over standard \acp{QEKF} due to superior convergence and consistency properties. We demonstrate the utility and accuracy of the developed observer through a series of LiDAR mapping experiments performed on a Cassie-series biped robot. 

In summary, this work makes the following contributions:
\begin{enumerate}
\item Derivation of a continuous-time right-invariant \ac{EKF} for an \ac{IMU}/contact process model with a forward kinematic measurement model; the observability analysis is also presented;
\item State augmentation with \ac{IMU} biases;
\item Evaluations of the derived observer in simulation and hardware experiments using a 3D bipedal robot;
\item Alternative derivation of the observer using a left-invariant error definition;
\item Detailed explanation of the connection between the invariant error choice and the world-centric and robo-centric estimator formulations;
\item Equations provided for analytical discretization of the proposed observers; and
\item Development of an open-source C++ library for aided-inertial navigation using the \ac{InEKF} \url{https://github.com/RossHartley/invariant-ekf}.  
\end{enumerate}

\subsection{Outline}
\begin{figure}[t]
  \centering
  	\vspace{0.5em}
     \includegraphics[width=0.75\columnwidth]{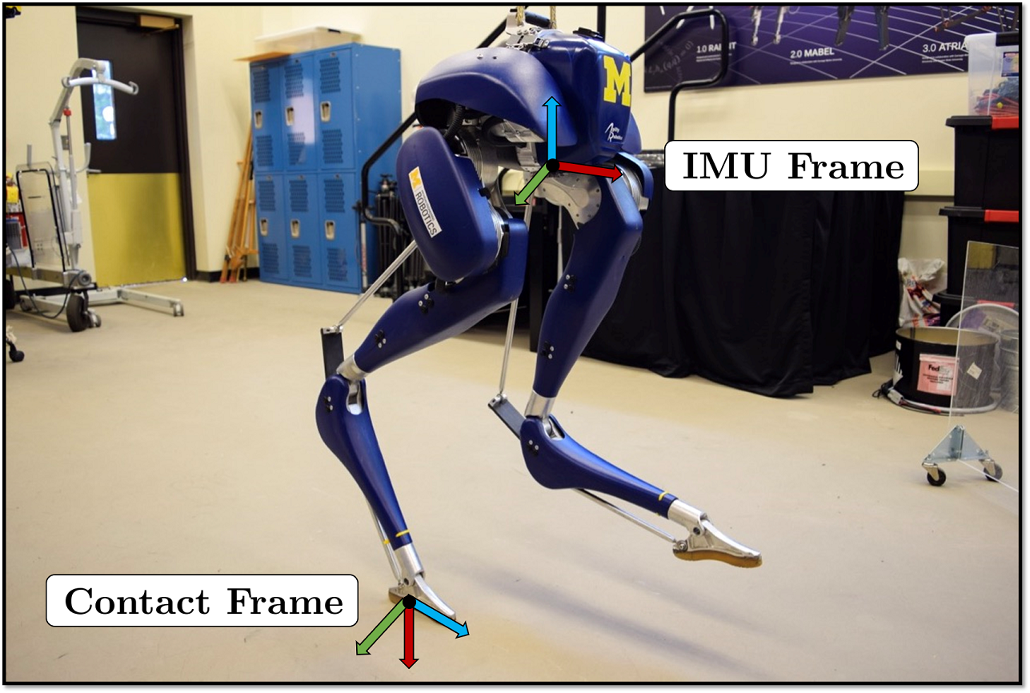}
    \caption{A Cassie-series biped robot is used for both simulation and experimental results. The robot was developed by Agility Robotics and has 20 degrees of freedom, 10 actuators, joint encoders, and an \acf{IMU}. The contact and \ac{IMU} frames used in this work are depicted above.} 
\label{fig:cassie}
\end{figure}

The remainder of this article is organized as follows. Background and related work are given in section~\ref{sec:literature}. Section~\ref{sec:prelim} provides the necessary preliminary material needed for understanding the \ac{InEKF} formulation, which is motived by an example from attitude dynamics in section~\ref{sec:attitide}. Section~\ref{sec:riekf} provides the derivation of a \ac{RIEKF} for contact-inertial navigation with a right-invariant forward kinematic measurement model. In Section~\ref{sec:sim}, we present simulation results comparing the convergence properties to a state-of-the-art \ac{QEKF}. Section~\ref{sec:bias} discusses the state augmentation of the previously derived \ac{InEKF} with \ac{IMU} bias. The consequences of adding and removing of contact points in the estimator are described in Section~\ref{sec:switchcontact}. Experimental evaluations on a 3D biped robot, shown in Figure~\ref{fig:cassie}, are presented in Section~\ref{sec:experimental_results} along with an application towards LiDAR-based terrain mapping. Section~\ref{sec:liekf} provides an alternative derivation of the proposed observer using the left-invariant dynamics along with an explanation of how to easily switch between the two formulations. Section~\ref{sec:robocentric} details how these equations can be modified to create a ``robot-centric'' estimator. Section~\ref{sec:additionalmeasurements} itemizes additional sensor measurements that fit within the \ac{InEKF} framework and describes the relation between the developed filter and landmark-based \ac{SLAM}. Finally, Section~\ref{sec:conclusion} concludes the article and suggests future directions. Details about time-discretization and useful Lie group expressions are given in the appendix.

%% file: literature.tex
\section{Background and Related work}
\label{sec:literature}
In this section, we first review the Kalman filtering literature to locate the proposed state estimator within the relate work. Then, we review the state estimation techniques for humanoid and legged robots.

\subsection{Kalman Filtering}

\emph{Filtering} methods involve estimating the robot's current state (and potentially landmarks) using the set of all measurements up to the current time \citep{anderson1979optimal,grewal2011kalman,smith2006approaches}. When the process model and measurements are linear, and the noise is white and Gaussian, Kalman filtering \citep{kalman1960new} provides an optimal method (minimum mean squared error) for state estimation. The general process for Kalman filtering involves two phases, propagation and correction. The state is typically represented using a Gaussian random vector, which is parameterized by a mean and a covariance. During the propagation phase, the previous state and covariance estimate are propagated forward in time using the system dynamics (alternatively known as the process model). When a measurement is obtained, the state and covariance estimate are corrected using the measurement model along with an associated  measurement noise covariance.

Although the Kalman filter provides a method for optimal linear filtering, most practical mobile robots have nonlinear system dynamics, and many useful sensor models are also nonlinear. For these cases, an \ac{EKF} can be designed, which utilizes Taylor series expansions to linearize the process and measurement models around the current state estimate \citep{thrun2005probabilistic}. Due to its low computational complexity and accurate performance, the \ac{EKF} quickly became the de facto standard of nonlinear filtering for many mobile robot applications, including wheeled vehicles, drones, and legged robots \citep{brossard2019rins,sebesta2014real,bloesch2012state}. The \acp{EKF} has also been proposed to solve the \ac{SLAM} problem \citep{smith1990estimating}. However, since the nonlinear system is linearized about the current state estimate, the \ac{EKF} is, at best, only a locally stable observer \citep{song1992extended,krener2003convergence}. The local convergence proofs are based on Lipschitz bounds of the nonlinear terms in a model, and hence ``the more nonlinear a system is, the worse an \ac{EKF} may perform''. Importantly, if the state estimate is initialized poorly, it is possible for the filter to diverge. In addition, because an \acp{EKF} uses a systems linearization about the current estimate, states that are unobservable can spuriously be treated as observable by the filter. While, this can be mitigated through the use of an observability-constrained \ac{EKF} developed by~\citet{huang2010observability}, it cannot be altogether avoided.
  
For many systems, the \emph{error-state} (or \emph{indirect}) \ac{EKF} offers superior performance to the standard (total state or direct) form. As the name implies, the \ac{ErEKF} is formulated using the errors, such as pose and velocity errors, as the filter variables, while the standard \ac{EKF} tracks the states themselves (pose and velocity directly) \citep{madyastha2011extended, roumeliotis1999circumventing,trawny2005indirect,sola2017quaternion}. Under small noise assumptions, this leads to linear error dynamics which are then used for covariance propagation in the error-state filter. The measurement model is also rewritten with respect to these errors. Although the error dynamics are linear in the error variables, they may still depend on the current state estimate, which if initialized poorly, will degrade the performance of the filter. However, the approximately linear nature of the error dynamics may respect the linear assumptions of the original Kalman filter better than the underlying system dynamics, which can lead to improved performance \cite{madyastha2011extended}. 

Perhaps the most important feature of the \ac{ErEKF} is the ability to circumvent dynamic modeling \citep{roumeliotis1999circumventing}. This is done by replacing the potentially complicated process model with a relatively simple \ac{IMU} integration model (also known as \emph{strapdown} modeling) \citep{titterton2004strapdown, merhav1991autonomously, woodman2007introduction}. Essentially, the \ac{IMU}'s angular velocity and linear acceleration measurements are integrated to propagate the state estimate, while the covariance is propagated using the error dynamics. Additional (independent) sensor measurements will correct the estimated error, which can then be used to update the state estimate. Using this method, there is no longer a need to formulate complicated, platform-specific dynamics models, which may require a large number of state variables and is likely to be exceedingly nonlinear. The ``strapdown'' \acp{ErEKF} has proven to yield highly accurate results (even with a low-cost \acp{IMU}) and continues to form the basis of many \acp{INS} \citep{barshan1995inertial,lefferts1982kalman,trawny2005indirect,sola2017quaternion,bloesch2012state}.

In the standard formulation of Kalman filtering theory, the system evolves on Euclidean spaces. However, in many cases, the state variables we are interested in lie on a \emph{manifold}. For example, the orientation of a 3D rigid body is represented by an element of the \emph{special orthogonal group}, $\SO(3)$. This matrix Lie group is defined by the set of orthogonal $3 \times 3$ matrices with a determinant of one. While the matrix contains nine variables, yet the dimension of the manifold is only three. One common approach is to parameterize $\SO(3)$ using local coordinates such as three Euler or Tait-Bryan angles \citep{gebre2004design,setoodeh2004attitude}. This allows the standard \ac{ErEKF} equations to be applied; however, these local parameterizations are often arbitrary (and therefore confusing) and contain singularities (the well-known Gimbal lock problem). Alternatively, it is possible to represent 3D orientation using quaternions, which are a four-dimensional double cover (a two to one diffeomorphism) of $\SO(3)$. Using quaternions eliminates the singularities; however, modifications to the standard \ac{ErEKF} equations have to be made \citep{sola2017quaternion, trawny2005indirect}. In brief, while the orientation is represented by a four-dimensional quaternion, the orientation error has to be defined by a $3-$vector (in the Lie algebra of $\SO(3)$) and the associated covariance by a $3 \times 3$ matrix in order to prevent degeneracy. Also, the orientation corrections are done through quaternion multiplication instead of vector addition. This \ac{QEKF} \footnote{The quaternion-based formulation of \acp{EKF} is also sometimes called ``multiplicative filtering'' (MEKF) due to the orientation correction being done through quaternion multiplication \citep{markley2004multiplicative}.} has been well studied and implemented on a number of platforms, ranging from spacecraft \citep{murrell1978precision,lefferts1982kalman} to legged robots \citep{bloesch2012state,rotella2014state,fallon2014drift}.
 
It turns out, many useful robot states can be characterized using matrix Lie groups. Examples include 2D orientation, $\SO(2)$, 3D orientation, $\SO(3)$, and 3D pose (orientation and position), $\SE(3)$. If the state to be estimated is a matrix Lie group, it is possible to further improve the \ac{EKF} filtering approach. \citet{bourmaud2013discrete,bourmaud2015continuous} developed versions of both discrete and continuous-time \acp{EKF} for systems where the state dynamics and measurements evolve on matrix Lie groups. In these formulations, noise is represented as a \emph{concentrated Gaussian on Lie groups} \citep{wang2006error,wang2008nonparametric}, which is a generalization of the multivariate Gaussian distribution. In essence, noise is represented as a Gaussian in the tangent space about a point on the manifold. This noise is then mapped to the Lie group through the use of the group's exponential map, resulting in a decidedly non-Gaussian distribution on the manifold. An improved state estimate is obtained due to the filter taking into account the geometry and structure of the problem \citep{bourmaud2015continuous}.

Most recently, a new type of \ac{EKF} has been developed that is rooted in the theory of invariant observer design, in which the estimation error is invariant under the action of a Lie group \citep{aghannan2002invariant,bonnabel2009non}. This invariance is referred to as the \emph{symmetries} of the system \citep{barczyk2013invariant}. This work led to the development of the \ac{InEKF} \citep{bonnabel2007left,barrau2015non,barrau2017invariant,barrau2018invariant}, with successful applications and promising results in \ac{SLAM} \citep{barrau2015non,zhang2017convergence} and aided \acp{INS} \citep{bonnable2009invariant,barczyk2011invariant,barczyk2013invariant,barrau2015non,wu2017invariant}. Similar to the above mentioned \acp{EKF} on matrix Lie groups, the state is again represented as a matrix Lie group and the noise as a concentrated Gaussian on the group. 

However, the \ac{InEKF} exploits available system symmetries to further improve filtering results. The culminating result of the \ac{InEKF} states that if a system satisfies a ``group-affine'' property, the estimation error satisfies a ``log-linear'' autonomous differential equation on the Lie algebra of the corresponding Lie group \citep{barrau2018invariant,barrau2015non}. In other words, the system linearization does not depend upon the estimated states. Therefore, one can design a nonlinear state estimator with strong convergence properties. Surprisingly, many mobile robot state estimation problems can be solved within the \ac{InEKF} framework. This includes attitude estimation \citep{bonnabel2007left}, inertial odometry \citep{barrau2018invariant,barrau2015non}, velocity-aided inertial navigation \citep{bonnable2009invariant}, landmark-aided navigation \citep{barrau2015non}, GPS and magnetometer-aided navigation \citep{barczyk2011invariant}, and even EKF-based \ac{SLAM}. In this article, we extend this class of solutions to contact-aided inertial navigation \citep{hartley2018contact}, where forward kinematics is used to correct an inertial and contact-based process model. This approach successfully allows an \ac{InEKF} to be used for legged robot state estimation.

%%%%%%%%%%%%%%%%%%%%%%%%%%%%%%%%%%%%%%%%%%%%%%%%%%%%%%%%%%%%%%%%%%%%%%%%
\subsection{Legged Robot and Humanoid State Estimation}

Legged robots are a subclass of mobile robots that locomote through direct and switching contact with the environment. These robots typically contain proprioceptive sensors, such as \ac{IMU}, joint encoders, and contact sensors. In addition, some legged robots, especially humanoids, also have access to exteroceptive sensors, namely cameras and LiDARs. As with all mobile robots, state estimation for legged robots is critical for mapping, planning, designing feedback controllers, and developing general autonomy. In this section, an overview of notable techniques for legged robot state estimation is given.
 
The simplest approach for estimation of a legged robot's spatial location and velocity is kinematic dead-reckoning, otherwise known as kinematic odometry. This involves estimating relative transformations using only kinematic and contact measurements. In particular, encoder measurements and the kinematics model are used to track the position, orientation, and velocity of the robot's base frame based on the assumption that a stance foot remains fixed to the ground. Although this method can be easily implemented, the state estimate is typically noisy due to kinematic modeling errors, encoder noise, and foot slip~\citep{roston1991dead}. When only one foot is in contact with the ground, this ``static contact assumption'' may be violated. For example, if the robot has point feet, the stance foot position may remain fixed, but the foot orientation is free to rotate (without changing the joint angles). Therefore, a gyroscope is often used to provide angular velocity measurements which allows the robot's body velocity to be recovered. 

Alternatively, if the terrain is known a priori and at least three noncollinear point feet are on the ground, \citet{lin2005leg} showed that the robot's instantaneous base pose can be computed through kinematics. These kinematic-based methods have been implemented on a number of legged robots including a planar one-legged hopper \citep{hodgins1989legged}, the CMU Ambler hexapod \citep{roston1991dead}, the RHex hexapod \citep{lin2005leg}, and the biped robot MARLO \citep{da20162d,da2017supervised}. However, due to the high amounts of noise coming from encoders and foot slip, the velocity estimate typically needs to be heavily filtered before becoming usable in the feedback controller \citep{hartley2017stabilization}. In addition, this noise causes the position and orientation estimates to drift substantially rendering the estimator impractical for mapping and autonomy tasks.  

Fortunately, legged robots are often equipped with additional sensors such as \acp{IMU}, GPS, cameras, or LiDARs which provide independent, noisy odometry measurements. Much of the literature on legged robot state estimation focuses on fusing these measurements (potentially with kinematic odometry) using filtering and smoothing methods. \citet{singh2006optical} combined inertial measurements with optical flow measurements in a four-phase hybrid EKF. This required explicit dynamic modeling of the robot in flight, landing, stance, and thrust phases. \citet{lin2006sensor} took a similar model-based approach and used an \ac{EKF} to fuse kinematic information with \ac{IMU} measurements to estimate the state of a hexapod. \citet{cobano2008location} developed an \ac{EKF} that fuses kinematic odometry and magnetometer readings with position measurements from a GPS to localize a SILO4 quadruped outdoors. This implementation fixes the issues with unbounded drift, but cannot operate in GPS-denied environments. If a prior terrain map is known, \citet{chitta2007proprioceptive} showed that it is possible to solve the localization problem for legged robots using only proprioceptive sensors and a particle filter. The key idea was that if the robot ``senses'' that a terrain change through kinematics, then this limits the potential locations the robot can be in a known map. The method was demonstrated on the LittleDog quadruped.

A breakthrough came in 2012 when \citet{bloesch2012state} combined inertial and kinematic measurements in an observability-constrained \ac{ErEKF} using the strapdown \ac{IMU} modeling approach. In this work, no \emph{a priori} knowledge of the terrain is assumed, and the \ac{IMU} integration model completely eliminates the need for dynamic modeling of the robot. Therefore, the derived filter equations are general enough to be used on any legged robot. The key idea was to augment the state vector with the set of all foot positions currently in contact with the environment. During the prediction phase, the foot contact dynamics are assumed to be Brownian motion, which can account for some foot slippage. In the correction phase, forward-kinematic position measurements are used to correct the estimated state. This work was conducted on the StarlETH quadruped robot. If the stance feet orientations also remain constant, as is the case for many humanoids, \citet{rotella2014state} showed that this \ac{ErEKF} can be extended to allow forward-kinematic orientation measurements. The same group also formulated a similar unscented Kalman filter that uses forward-kinematic velocity measurements to correct inertial predictions and to accurately detect foot slip~\citep{bloesch2013state}. A detailed analysis of these filtering techniques combined with methods for incorporating computer vision can be found in~\citet{bloesch2017state}. Due to the complexity involved in accurately formulating dynamic models, many groups have since adopted this \ac{IMU} motion model approach to legged robot state estimation \citep{xinjilefu2014decoupled,fallon2014drift,hartley2018contact}. 

This combined inertial and kinematic filtering approach yields an estimate of the robot's base pose and velocity. However, some legged robots require additional states to be estimated. \citet{hwangbo2016probabilistic} formulated a probabilistic contact estimator for cases when contact sensors are unavailable. \citet{xinjilefu2014decoupled} developed a decoupled \ac{EKF} that is able to estimate the full state of the humanoid robot ATLAS, including base states, joint angles, and joint velocities. Using proprioceptive sensing only, \citet{bloesch2012state} proved that the absolute positions and yaw angles are unobservable. Thus, over time, estimates of these quantities will drift unboundedly. This is unacceptable for global mapping and planning algorithms; however, local elevation maps can still be obtained \citep{fankhauser2014robot}. \citet{fallon2014drift} proposed a method for drift-free state estimation for the humanoid ATLAS. In their implementation, inertial and kinematic measurements were fused to yield accurate odometry. Point cloud data from a LiDAR sensor was used with a particle filter to localize the robot into a pre-built map. This approach provided corrections of position and yaw to obtain a drift-free estimate of the state. \citet{nobili2017heterogeneous} took a similar approach but used the \ac{ICP} algorithm to perform LiDAR-based point cloud matching. The algorithm was tested on the HyQ quadruped robot.

In this article, we propose using an \ac{InEKF} to estimate the base pose and velocity of a general legged robot. The approach we take is most similar to \citet{bloesch2012state}, however we model the entire state as a single matrix Lie group as opposed to a decoupled state approach. This allows us to take advantage of the geometry and symmetry of the estimation problem to formulate autonomous error dynamics. In addition with our formulation, the linearizations are independent of the state estimate resulting in improved convergence properties, especially for poor state initializations. We formulate both world-centric and robo-centric state estimators highlighting the relation between the left- and right-invariant error dynamics. In addition, we provide exact analytical time-discretizations of both filters. The implemented filters can be run at high speeds \mbox{($>2000\Hz$)} and can be directly used for accurate local odometry. We demonstrate this idea through a LiDAR terrain mapping application on a Cassie-series biped robot.

%% file: preliminaries.tex
\section{Theoretical background and preliminaries}
\label{sec:prelim}

We assume a matrix Lie group~\citep{hall2015lie,chirikjian2011stochastic} denoted $\lieGroup$ and its associated Lie Algebra denoted $\lieAlgebra$. If elements of $\lieGroup$ are $n \times n$ matrices, then so are elements of $\lieAlgebra$. When doing calculations, it is very convenient to let
$$\vectorToAlgebra[(\cdot)]:\realnumbers^{\dimension \lieAlgebra} \to \lieAlgebra$$
be the linear map that takes elements of the tangent space of $\lieGroup$ at the identity to the corresponding matrix representation so that the \textit{exponential map of the Lie group}, $\exp:\realnumbers^{\dimension \lieAlgebra} \to \lieGroup$,
is computed by
$$\exp(\tangentError) = \exp_m(\vectorToAlgebra[\tangentError]),$$ where $\exp_m(\cdot)$ is the usual exponential of $n \times n$ matrices.

A process dynamics evolving on the Lie group with state at time $t$, $\X[t] \in \lieGroup$, is denoted by
$$\deriv \X[t] = f_{u_t}(\X[t]), $$
and $\XE[t]$ is used to denote an estimate of the state. The state estimation error is defined using right or left multiplication of $\X[t][-1]$  as follows.
\begin{definition}[Left and Right Invariant Error] 
The right- and left-invariant errors between two trajectories $\X[t]$ and $\XE[t]$ are:
\begin{equation} \label{eq:invariant_error}
\begin{split}
\groupError[t][r] &= \XE[t] \X[t][-1] = (\XE[t] \textbf{L}) (\X[t] \textbf{L})^{-1} \quad \text{(Right-Invariant)}\\
\groupError[t][l] &= \X[t][-1] \XE[t] = (\textbf{L} \XE[t])^{-1} (\textbf{L} \X[t]), \quad \text{(Left-Invariant)}
\end{split}
\end{equation}
where $\textbf{L}$ is an arbitrary element of the group.
\end{definition}
The following two theorems are the fundamental results for deriving an \ac{InEKF} and show that by correct parametrization of the error variable,  a wide range of nonlinear problems can lead to linear error equations.
\begin{theorem}[Autonomous Error Dynamics \citep{barrau2017invariant}] \label{theorem:autonomous_error_dynamics}
A system is group affine if the dynamics, $f_{u_t}(\cdot)$, satisfies:
\begin{equation} \label{eq:group_affine}
f_{u_t}(\textnormal{\textbf{X}}_1 \textnormal{\textbf{X}}_2) = f_{u_t}(\textnormal{\textbf{X}}_1) \textnormal{\textbf{X}}_2 + \textnormal{\textbf{X}}_1 f_{u_t}(\textnormal{\textbf{X}}_2) - \textnormal{\textbf{X}}_1 f_{u_t}(\textbf{\textit{I}}_d) \textnormal{\textbf{X}}_2
\end{equation}
for all $t>0$ and $\textnormal{\textbf{X}}_1, \textnormal{\textbf{X}}_2 \in \lieGroup$. Furthermore, if this condition is satisfied, the right- and left-invariant error dynamics are trajectory independent and satisfy:
\begin{alignat*}{2}
\deriv \groupError[t][r] &= g_{u_t}(\groupError[t][r]) \quad \text{where} \quad
g_{u_t}(\groupError[][r]) &&= f_{u_t}(\groupError[][r]) - \groupError[][r] f_{u_t}(\textbf{\textit{I}}_d)  \\
\deriv \groupError[t][l] &= g_{u_t}(\groupError[t][l]) \quad \text{where} \quad
g_{u_t}(\groupError[][l]) &&=  f_{u_t}(\groupError[][l]) - f_{u_t}(\textbf{\textit{I}}_d) \groupError[][l] \\
\end{alignat*}
\end{theorem}
In the above, $\textbf{\textit{I}}_d \in \lieGroup$ denotes the group identity element; to avoid confusion, we use $\I$ for a $3 \times 3$ identity matrix, and $\I[n]$ for the $n \times n$ case. The following statements hold for both the left- and right-invariant error dynamics. 

Define $\A[t]$ to be a $\dimension \lieAlgebra \times \dimension \lieAlgebra$ matrix satisfying $$g_{u_t}(\exp(\tangentError[])) \triangleq \vectorToAlgebra[(\A[t] \tangentError[])] + \mathcal{O}(||\tangentError[]||^2).$$ 
For all $t \ge 0$, let $\tangentError[t]$ be the solution of  the linear differential equation
\begin{equation}
\label{eq:LTVODE}
\deriv \tangentError[t][] = \A[t] \tangentError[t][].
\end{equation}

\begin{theorem}[Log-Linear Property of the Error \citep{barrau2017invariant}] \label{theorem:log_linear_error}
Consider the right-invariant error, $\groupError[t][]$, between two trajectories (possibly far apart). For arbitrary initial error $\tangentError[0][] \in \realnumbers^{\dimension \lieAlgebra}$, if 
\mbox{$\groupError[0][] =\exp(\tangentError[0][])$}, then for all $t\ge 0$, 
\begin{equation*}
\groupError[t][] = \exp(\tangentError[t][]);
\end{equation*}
that is, the nonlinear estimation error $\groupError[t][]$ can be exactly recovered from the time-varying linear differential equation \eqref{eq:LTVODE}.
\end{theorem}
This theorem states that \eqref{eq:LTVODE} is not the typical Jacobian linearization along a trajectory because the (left- or) right-invariant error on the Lie group can be exactly recovered from its solution. This result is of major importance for the propagation (prediction) step of the InEKF~\citep{barrau2017invariant}, where in the deterministic case, the log-linear error dynamics allows for exact covariance propagation.

\begin{remark}
This indirect way of expressing the Jacobian of $g_{u_t}$ is from~\citet{barrau2017invariant}; it is used because we are working with a matrix Lie group viewed as an embedded submanifold of a set of $n \times n$ matrices.
\end{remark}

During the correction step of a Kalman filter, the error is updated using incoming sensor measurements. If these observations take a particular form, then the linearized observation model and the innovation will also be autonomous \citep{barrau2017invariant}. This happens when the measurement, $\Y[t]$, can be written as either
\begin{equation} \label{eq:invariant_observations}
\begin{alignedat}{2} 
\Y[t] &= \X[t] \b + \V[t]  \quad &&\text{(Left-Invariant Observation)}~~~\text{or} \\ 
\Y[t] &= \X[t][-1] \b + \V[t] \quad &&\text{(Right-Invariant Observation)},
\end{alignedat}
\end{equation}
where $\b$ is a constant vector and $\V[t]$ is a vector of Gaussian noise.

The adjoint representation plays a key role in the theory of Lie groups and through this linear map we can capture the non-commutative structure of a Lie group.
   \begin{definition}[The Adjoint Map, see \citet{hall2015lie} page 63]
Let $\lieGroup$ be a matrix Lie group with Lie algebra $\lieAlgebra$. For any $\X \in \lieGroup$ the adjoint map, $\Adjoint[\X]:\lieAlgebra \to \lieAlgebra$, is a linear map defined as \mbox{$\Adjoint[\X](\vectorToAlgebra[\tangentError]) = \X \vectorToAlgebra[\tangentError] \X[][-1]$}. Furthermore, we denote the matrix representation of the adjoint map by $\Adjoint[\X]$.
\end{definition}

For more details on the material discussed above, along with the theory and proofs about the \ac{InEKF}, we refer reader to~\citet{barrau2015non,barrau2017invariant,barrau2018invariant}.

%%%%%%%%%%%%%%%%%%%%%%%%%%%%%%%%%%%%%%%%%%%%%%%%%%%%%%%%%%%%%

%% file: attitude.tex
\section{A motivating example: 3D orientation propagation}
\label{sec:attitide}
Suppose we are interested in estimating the 3D orientation of a rigid body given angular velocity measurements in the body frame, $\wM[t] \triangleq \vector[\omega_x, \omega_y, \omega_z] \in \realnumbers^{3}$. This type of measurement can be easily obtained from a gyroscope. 

There are several different parameterizations of $\SO(3)$; Euler angles, quaternions, and rotation matrices being the most common. If we let $\textbf{q}_t \triangleq \vector[q_x, q_y, q_z]$ be a vector of Euler angles using the $R_zR_yR_x$ convention, then the orientation dynamics can be expressed as~\citep{diebel2006representing}
\begin{equation*}
    \small
    \deriv 
    \begin{bmatrix}
        q_x \\ q_y \\ q_z 
    \end{bmatrix} =
    \begin{bmatrix}
        1 & \sin(q_x) \tan(q_y) & \cos(q_x) \tan(q_y) \\
        0 & \cos(q_x) & -\sin(q_x) \\
        0 & \sin(q_x) \sec(q_y) & \cos(q_x) \sec(q_y)
    \end{bmatrix}
    \begin{bmatrix}
        w_x \\ w_y \\ w_z 
    \end{bmatrix}.
\end{equation*}
Let $\delta \textbf{q}_t \triangleq \textbf{q}_t - \bar{\textbf{q}}_t \in \realnumbers^3$ be the error between the true and estimated Euler angles. The error dynamics can be written as a nonlinear function of the error variable, the inputs, and the state
\begin{equation*}
    \deriv \delta \textbf{q}_t = g(\delta \textbf{q}_t, \wM[t], \textbf{q}_t).
\end{equation*}
In order to propagate the covariance in an EKF, we need to linearize the error dynamics at the current state estimate, $\textbf{q}_t = \bar{\textbf{q}}_t$ (i.e. zero error). This leads to a linear error dynamics of the form:
\begin{equation*}
    \small
    \begin{split}
    \deriv \delta \textbf{q}_t &\approx 
    \begin{bmatrix}
        0 & (\omega_z \bar{c}_x + \omega_y \bar{s}_x)/\bar{c}_y^{2} & \bar{t}_y (\omega_y \bar{c}_x - \omega_z \bar{s}_x) \\
        0 & 0 & \omega_z \bar{c}_x + \omega_y \bar{s}_x \\
        0 & (\bar{s}_y (\omega_z \bar{c}_x + \omega_y \bar{s}_x))/\bar{c}_y^{2} & (\omega_y \bar{c}_x - \omega_z \bar{s}_x)/\bar{c}_y 
    \end{bmatrix} \delta \textbf{q}_t \\ 
    &\triangleq \A(\wM[t],\bar{\textbf{q}}_t) \delta \textbf{q}_t,
    \end{split}
\end{equation*}
where $\bar{c}_x$, $\bar{s}_x$, and $\bar{t}_x$ are shorthand for $\cos(\bar{q}_x)$, $\sin(\bar{q}_x)$, and $\tan(\bar{q}_x)$. The linear dynamics matrix clearly depends on the estimated angles, therefore bad estimates will affect the accuracy of the linearization and ultimately the performance and consistency of the filter.

Now instead, let's use a rotation matrix to represent the 3D orientation, $\R[t] \in \SO(3)$. The dynamics can now be simply expressed as
\begin{equation*}
    \deriv \R[t] = \R[t] \vectorToSkew[\wM[t]],
\end{equation*}
where $\vectorToSkew[\cdot]$ denotes a $3\times3$ skew-symmetric matrix. If we define the error between the true and estimated orientation as $\groupError[t] \triangleq \R[t][\transpose] \RE[t] \in \SO(3)$, then the (left-invariant) error dynamics becomes
\begin{equation} \label{eq:so3_example}
    \begin{split}
    \deriv \groupError[t] &= \R[t][\transpose] \RE[t] \vectorToSkew[\wM[t]] + \left(\R[t] \vectorToSkew[\wM[t]]\right)^\transpose \RE[t] \\
    &= \R[t][\transpose] \RE[t] \vectorToSkew[\wM[t]] - \vectorToSkew[\wM[t]] \R[t][\transpose] \RE[t] \\
    &= \groupError[t] \vectorToSkew[\wM[t]]  - \vectorToSkew[\wM[t]] \groupError[t] \\
    &= g(\groupError[t], \wM[t]).
    \end{split} 
\end{equation}
Using this particular choice of state and error variable yields an autonomous error dynamics function (independent of the state directly). Since, $\SO(3)$ is a Lie Group, we can look at the dynamics of a redefined error that resides in the tangent space, $\groupError[t] \triangleq \exp(\tangentError[t])$.
\begin{equation*}
    \begin{split}
    \deriv (\exp(\tangentError[t])) &= \exp(\tangentError[t]) \vectorToSkew[\wM[t]]  - \vectorToSkew[\wM[t]] \exp(\tangentError[t]) \\
    \deriv \left(\I + \vectorToSkew[\tangentError[t]] \right)&\approx \left(\I + \vectorToSkew[\tangentError[t]] \right) \vectorToSkew[\wM[t]]  - \vectorToSkew[\wM[t]] \left(\I + \vectorToSkew[\tangentError[t]] \right) \\
    \implies \deriv \vectorToSkew[\tangentError[t]] &= \vectorToSkew[\tangentError[t]] \vectorToSkew[\wM[t]] - \vectorToSkew[\wM[t]] \vectorToSkew[\tangentError[t]] \\
    &= \vectorToSkew[ -\vectorToSkew[\wM[t]] \tangentError[t] ] \\
    \implies \deriv \tangentError[t] &= \vectorToSkew[-\wM[t]] \tangentError[t] 
    \end{split}
\end{equation*}
After making a first-order approximation of the exponential map, the tangent space error dynamics become linear. In addition, this linear system only depends on the error, not the state estimate directly. In other words, wrong state estimates will not affect the accuracy of the linearization, which leads to better accuracy and consistency of the filter. For $\SO(3)$, this effect is well studied and has been leveraged to design the commonly used \acp{QEKF}\footnote{The set of quaternions, along with quaternion multiplication, actually forms a Lie group.}, \citep{trawny2005indirect, sola2017quaternion}. However, the extension to general matrix Lie groups, called the \ac{InEKF}, was only recently developed by \citet{barrau2017invariant}.

In the above example, we utilized the first-order approximation for the exponential map of $\SO(3)$; $\exp(\tangentError[t]) \approx \I + \vectorToSkew[\tangentError[t]]$. In general, one may ask how much accuracy is lost when making this approximation. The surprising result by \citet{barrau2017invariant} is that this linearization is, in fact, exact. This is the basis of Theorem \ref{theorem:log_linear_error}. If the initial error is known, the nonlinear error dynamics can be exactly recovered from this linear system.

Let's demonstrate this theorem for the SO(3) example. Let $\groupError[0]=\exp(\tangentError[0])$ be the initial left invariant error. We can show that $\groupError[t] = \R[t][\transpose]\groupError[0]\R[t]$ is a solution to the error dynamics equation \eqref{eq:so3_example} through differentiation.
\begin{equation*}
\begin{split}
    \deriv \groupError[t] &= \R[t][\transpose] \groupError[0] \R[t] \vectorToSkew[\wM[t]] - \vectorToSkew[\wM[t]] \R[t][\transpose] \groupError[0] \R[t]\\
    &= \groupError[t] \vectorToSkew[\wM[t]]  - \vectorToSkew[\wM[t]] \groupError[t] \\
\end{split}
\end{equation*}
Once we replace the group error with the tangent space error and use the group's adjoint definition to shift the rotation inside the exponential, we arrive at a simple expression for the tangent space error.
\begin{equation*}
    \begin{split}
        \groupError[t] &= \R[t][\transpose]\groupError[0]\R[t] \\
        \implies \exp(\tangentError[t]) &= \R[t][\transpose]\exp(\tangentError[0])\R[t] = \exp(\R[t][\transpose] \tangentError[0]) \\
        \implies \tangentError[t] &= \R[t][\transpose] \tangentError[0] \\
    \end{split}
\end{equation*}
Our final log-linear error dynamics can now be obtained by differentiating this new error expression. 
\begin{equation*}
    \begin{split}
        \deriv \tangentError[t] &= -\vectorToSkew[\wM[t]] \R[t][\transpose] \tangentError[0] \\
        &= -\vectorToSkew[\wM[t]] \tangentError[t]
    \end{split}
\end{equation*}
Again, this result shows that if the initial error is known, the nonlinear error dynamics can be exactly recovered from this linear system. In this work, we leverage these ideas to develop a contact-aided inertial observer for legged robots.

%% file: riekf.tex
\section{$\SE_{N+2}(3)$ Continuous Right-Invariant EKF}
\label{sec:riekf}
In this section, we derive a \acf{RIEKF} using IMU and contact motion models with corrections made through forward kinematic measurements. This \ac{RIEKF} can be used to estimate the state of a robot that has an arbitrary (finite) number of points in contact with a static environment. While the filter is particularly useful for legged robots, the same theory can be applied for manipulators as long as the contact assumptions (presented in Section \ref{sec:riekf_dynamics}) are verified. 

In order to be consistent with the standard InEKF theory, IMU biases are neglected for now. Section \ref{sec:bias} provides a method for reintroducing the bias terms.

\subsection{State Representation}
As with typical aided inertial navigation, we wish to estimate the orientation, velocity, and position of the IMU (body) in the world frame \citep{lupton2012visual,forster2016manifold,yang2017monocular}. These states are represented by $\orientation[WB](t), \linearVelocity[B][W](t)$, and $\position[WB][W](t)$ respectively. In addition, we append the position of all contact points (in the world frame), ${}_\text{W}\p[\text{WC}_i](t)$, to the list of state variables. This is similar to the approach taken in \citet{bloesch2012state,bloesch2017state}.

The above collection of state variables forms a matrix Lie group, $\lieGroup$. Specifically, for $N$ contact points, \mbox{$\X[t] \in \SE_{N+2}(3)$} can be represented by the following matrix:
\begin{equation*}
\nonumber \textbf{X}_t \triangleq
\begin{bmatrix}
\orientation[WB](t) & \linearVelocity[B][W](t) & \position[WB][W](t) & {}_\text{W}\p[\text{WC}_1](t) & \cdots & {}_\text{W}\p[\text{WC}_N](t)  \\
\zeros[1,3] & 1 & 0 & 0 & \cdots & 0 \\
\zeros[1,3] & 0 & 1 & 0 & \cdots & 0 \\
\zeros[1,3] & 0 & 0 & 1 & \cdots & 0 \\
\vdots & \vdots & \vdots & \vdots & \ddots & \vdots \\ 
\zeros[1,3] & 0 & 0 & 0 & \cdots & 1 \\
\end{bmatrix}.
\end{equation*}
This Lie group is an extension of $\SE(3)$ and has been previously used to solve inertial navigation \citep{barrau2015non} and \ac{SLAM} problems \citep{bonnabel2012symmetries,barrau2015ekf}. In fact, the estimators derived in this work have a connection to the landmark-based \ac{SLAM} problem. This connection is detailed later in Section~\ref{sec:additionalmeasurements}.

Because the process and measurements models for each contact point, ${}_\text{W}\p[\text{WC}_i](t)$, are identical, without loss of generality, we will derive all further equations assuming only a single contact point. Furthermore, for the sake of readability, we introduce the following shorthand notation:
\begin{equation*}
\begin{split}
\X[t] \triangleq
\begin{bmatrix}
\R[t] & \v[t] & \p[t] & \d[t] \\
\zeros[1,3] & 1 & 0 & 0 \\
\zeros[1,3] & 0 & 1 & 0 \\
\zeros[1,3] & 0 & 0 & 1 \\
\end{bmatrix}
,\quad
\u[t] = 
\begin{bmatrix}
\angularVelocityM[WB][B](t) \\
\accelerationM[WB][B](t) \\
\end{bmatrix} 
\triangleq 
\begin{bmatrix}
\wM[t] \\
\aM[t] \\
\end{bmatrix}, 
\end{split}
\end{equation*}
where the input $\u[t]$ is formed from the angular velocity and linear acceleration measurements coming from the IMU. It is important to note that these measurements are taken in the body (or IMU) frame.
The Lie algebra of $\lieGroup$, denoted by $\lieAlgebra$, is an $N+5$ dimensional square matrix. We use the \textit{hat} operator, $\vectorToAlgebra[(\cdot)]: \realnumbers^{3N+9} \to \lieAlgebra$, to map a vector to the corresponding element of the Lie algebra. In the case of a single contact, for example, this function is defined by: 
\begin{equation}
\vectorToAlgebra[\tangentError] = 
\begin{bmatrix}
\vectorToSkew[\tangentError[][R]] & \tangentError[][v] & \tangentError[][p] & \tangentError[][d] \\
\zeros[1,3] & 0 & 0 & 0 \\
\zeros[1,3] & 0 & 0 & 0 \\
\zeros[1,3] & 0 & 0 & 0 \\
\end{bmatrix},
\end{equation}
where $\vectorToSkew[\cdot]$ denotes a $3\times 3$ skew-symmetric matrix. The inverse operation is defined using the \textit{wedge} operator, \mbox{$\algebraToVector[(\cdot)]:\lieAlgebra \to \realnumbers^{3N+9}$}. The matrix representation of the adjoint is given by:
\begin{equation} \label{eq:adjoint}
\Adjoint[\X[t]] = 
\begin{bmatrix}
\R & \zeros & \zeros & \zeros \\
\vectorToSkew[\v[t]] \R[t] & \R[t] & \zeros & \zeros \\
\vectorToSkew[\p[t]] \R[t] & \zeros & \R[t] & \zeros \\
\vectorToSkew[\d[t]] \R[t] & \zeros & \zeros & \R[t] \\
\end{bmatrix}.
\end{equation}
A closed form expression for the exponential map of $\SE_{N+2}(3)$ is given in Appendix \ref{appx:formulas}.

%%%%%%%%%%%%%%%%%%%%%%%%%%%%%%%%%%%%%%%%%%%%%%%%%%%%%%%%%%%%%
\subsection{Continuous-Time System Dynamics} \label{sec:riekf_dynamics}
The IMU measurements are modeled as being corrupted by additive Gaussian white noise processes, per
\begin{alignat*}{2}
\wM[t] &= \w[t] + \noise[t][g], \qquad &&\noise[t][g] \sim \mathcal{GP}\left(\zeros[3,1], \Cov[][g]\,\delta(t - t^\prime)\right) \\
\aM[t] &= \a[t] + \noise[t][a], \qquad &&\noise[t][a] \sim \mathcal{GP}\left(\zeros[3,1], \Cov[][a]\,\delta(t - t^\prime)\right), 
\end{alignat*}
where $\mathcal{GP}$ denotes a Gaussian process and $\delta(t - t^\prime)$ denotes the Dirac delta function. These are explicit measurements coming directly from a physical sensor. In contrast, the velocity of the contact point is implicitly \textit{inferred} through a contact sensor; specifically, when a binary sensor indicates contact, the position of the contact point is \textit{assumed to remain fixed in the world frame}, i.e. the measured velocity is zero. In order to accommodate potential slippage, the measured velocity is assumed to be the actual velocity plus white Gaussian noise, namely
\begin{equation*}
\linearVelocityM[C][W] = \zeros[3,1] = \linearVelocity[C][C] + \noise[t][v], \quad \noise[t][v] \sim \mathcal{GP}\left(\zeros[3,1], \Cov[][v]\,\delta(t - t^\prime)\right) . \\
\end{equation*}
Using the IMU and contact measurements, the individual terms of the system dynamics can be written as:
\begin{equation}\label{eq:world_dynamics}
\begin{split}
\deriv \R[t] &= \R[t]\vectorToSkew[\wM[t] - \noise[t][g]] \\ %, \quad 
\deriv \v[t] &= \R[t](\aM[t] - \noise[t][a]) + \g \\
\deriv \p[t] &= \v[t] \\ %, \qquad \qquad \qquad~~
\deriv \d[t] &= \R[t] \FK[R](\encodersM[t])(-\noise[t][v]),
\end{split}
\end{equation}
where $\g$ is the gravity vector and $\FK[R](\encodersM[t])$ is the measured orientation of the contact frame with respect to the IMU frame as computed through encoder measurements, $\encodersM[t] \in \realnumbers^M$, and forward kinematics. Therefore, $\R[t] \FK[R](\encodersM[t])$ is a rotation matrix that transforms a vector from the contact frame to the world frame.

In matrix form, the dynamics can be expressed as
\begin{equation} 
\begin{split}
\deriv \X[t] &= 
\begin{bmatrix}
  \R[t]\vectorToSkew[\wM[t]] 
& \R[t]\aM[t] + \g  
& \v[t]
& \zeros[3,1] \\
\zeros[1,3] & 0 & 0 & 0  \\
\zeros[1,3] & 0 & 0 & 0  \\
\zeros[1,3] & 0 & 0 & 0  \\
\end{bmatrix} - 
\begin{bmatrix}
\R[t] & \v[t] & \p[t] & \d[t] \\
\zeros[1,3] & 1 & 0 & 0 \\
\zeros[1,3] & 0 & 1 & 0 \\
\zeros[1,3] & 0 & 0 & 1 \\
\end{bmatrix}
\begin{bmatrix}
\vectorToSkew[\noise[t][g]] 
& \noise[t][a]
& \zeros[3,1]
& \FK[R](\encodersM[t]) \noise[t][v]  \\
\zeros[1,3] & 0 & 0 & 0  \\
\zeros[1,3] & 0 & 0 & 0  \\
\zeros[1,3] & 0 & 0 & 0  \\
\end{bmatrix} \\
&\triangleq f_{u_t}(\X[t]) - \X[t] \vectorToAlgebra[\noise[t]],
\end{split}
\end{equation}
with \mbox{$\noise[t] \triangleq \vector[\noise[t][g],\; \noise[t][a],\; \zeros[3,1], \FK[R](\encodersM[t]) \noise[t][v]]$}. The deterministic system dynamics, $f_{u_t}(\cdot)$, can be shown to satisfy the group affine property, \eqref{eq:group_affine}. Therefore, following Theorem~\ref{theorem:autonomous_error_dynamics}, the left- and right-invariant error dynamics will evolve independently of the system's state.

Using Theorem~\ref{theorem:autonomous_error_dynamics}, the right-invariant error dynamics is 
\begin{equation}
\begin{split}
\deriv \groupError[t][r] &= f_{u_t}(\groupError[t][r]) - \groupError[t][r] f_{u_t}(\textbf{\textit{I}}_d) + (\XE[t]  \vectorToAlgebra[\noise[t]] \XE[t][-1]) \groupError[t][r] \\
&\triangleq g_{u_t}(\groupError[t][r]) + \vectorToAlgebra[\bar{\noise}_t] \groupError[t][r] \\
\end{split}
\end{equation}
where the second term arises from the additive noise. The derivation follows the results in \citet{barrau2017invariant} and is not repeated here.

Furthermore, Theorem~\ref{theorem:log_linear_error} specifies that the invariant error satisfies a log-linear property. Namely, if $\A[t]$ is defined by $g_{u_t}(\exp(\tangentError)) \triangleq \vectorToAlgebra[(\A[t] \tangentError)] + \mathcal{O}(||\tangentError||^2)$, then the log of the invariant error, $\tangentError \in \realnumbers^{\dimension \lieAlgebra}$, approximately satisfies\footnote{With input noise, Theorem \ref{theorem:log_linear_error} no longer holds, and the linearization is only approximate.} the linear system
\begin{equation} \label{eq:tangent_error_dynamics}
\begin{split}
\deriv \tangentError[t] &= \A[t][r] \tangentError[t] + \bar{\noise}_t = \A[t] \tangentError[t] + \Adjoint[\XE[t]] \noise[t] \\
\groupError[t][r] &= \exp(\tangentError[t]).
\end{split}
\end{equation}
To compute the matrix $\A[t]$, we linearize the invariant error dynamics, $g_{u_t}(\cdot)$, using the first order approximation \mbox{$\groupError[t][r] = \exp(\tangentError[t]) \approx \textbf{\textit{I}}_d + \vectorToAlgebra[\tangentError[t]]$} to yield
\begin{equation} \label{eq:linear_error_dynamics}
  \resizebox{\hsize}{!}{$
  \begin{aligned} 
  g_{u_t}&(\exp(\tangentError[t])) \approx \\
  &\begin{bmatrix}  
  \left( \I+ \vectorToSkew[\tangentError[t][R]] \right) \vectorToSkew[\wM[t]]  
  &\left( \I+ \vectorToSkew[\tangentError[t][R]] \right) \aM[t] + \g & \tangentError[t][v] 
  & \zeros[3,1]  \\
  \zeros[1,3] & 0 & 0 & 0  \\
  \zeros[1,3] & 0 & 0 & 0  \\
  \zeros[1,3] & 0 & 0 & 0  \\
  \end{bmatrix} -
  \begin{bmatrix}
  \I + \vectorToSkew[\tangentError[t][R]] & \tangentError[t][v] & \tangentError[t][p] & \tangentError[t][d] \\
  \zeros[3,1] & 1 & 0 & 0 \\
  \zeros[3,1] & 0 & 1 & 0  \\
  \zeros[3,1] & 0 & 0 & 1 \\
  \end{bmatrix}
  \begin{bmatrix}
  \vectorToSkew[\wM[t]] & \aM + \g & \zeros[3,1] & \zeros[3,1]  \\
  \zeros[1,3] & 0 & 0 & 0  \\
  \zeros[1,3] & 0 & 0 & 0  \\
  \zeros[1,3] & 0 & 0 & 0  \\
  \end{bmatrix} \\
  &\qquad= 
  \begin{bmatrix}
  \zeros[3,3] & \vectorToSkew[\g] \tangentError[t][R]  & \tangentError[t][v] & \zeros[3,1] \\
  \zeros[1,3] & 0 & 0 & 0  \\
  \zeros[1,3] & 0 & 0 & 0  \\
  \zeros[1,3] & 0 & 0 & 0  \\
  \end{bmatrix}
  = \vectorToAlgebra[
  \begin{bmatrix} 
  \zeros[3,1] \\
  \vectorToSkew[\g] \tangentError[t][R] \\
  \tangentError[t][v] \\
  \zeros[3,1]
  \end{bmatrix} ].
  \end{aligned}$} 
  \end{equation}

With the above, we can express the prediction step of the \ac{RIEKF}. The state estimate, $\XE[t]$, is propagated though the deterministic system dynamics, while the covariance matrix, $\P[t]$, is computed using the Riccati equation \citep{maybeck1982stochastic}, namely,
\begin{equation} \label{eq:propagation}
\deriv \XE[t] = f_{u_t}(\XE[t])~\text{and}~
\deriv \P[t] = \A[t] \P[t] + \P[t] \A[t][\transpose] + \bar{\Q}_t,
\end{equation}
where the matrices $\A[t]$ and $\bar{\Q}_t$ are obtained from \eqref{eq:linear_error_dynamics} and \eqref{eq:tangent_error_dynamics},
\begin{align}
% \small
\label{eq:A_right_no_bias}
\A[t] = 
\begin{bmatrix} 
\zeros & \zeros & \zeros & \zeros  \\
\vectorToSkew[\g] & \zeros & \zeros & \zeros  \\
\zeros & \I & \zeros & \zeros \\
\zeros & \zeros & \zeros & \zeros   
\end{bmatrix} \text{and}~
\bar{\Q}_t = \Adjoint[\XE[t]] \text{Cov}\left( \noise[t] \right) \Adjoint[\XE[t]]^\transpose.
\end{align}
\begin{remark}
For the right-invariant case, expression \eqref{eq:A_right_no_bias}, $\A[t]$ is time-invariant and the time subscript could be dropped. However, in general it can be time-varying, therefore, we use $\A[t]$ throughout the paper.
\end{remark}

%%%%%%%%%%%%%%%%%%%%%%%%%%%%%%%%%%%%%%%%%5
\subsection{Right-invariant Forward Kinematic Measurement Model}
Let $\encoders[t] \in \realnumbers^M$ denote the vector of joint positions (prismatic or revolute) between the body and the contact point. We assume that the encoder measurements at time $t$ are corrupted by additive white Gaussian noise.
\begin{equation}
\encodersM[t] = \encoders[t] + \noise[t][\alpha], \quad \noise[t][\alpha] \sim \mathcal{N}(\zeros[M,1], \Cov[][\alpha])
\end{equation}
Using forward kinematics, we measure the relative position of the contact point with respect to the body,
\begin{equation} \label{eq:forward_kinematics_position_measurement}
\position[BC][B](t) \triangleq \FK[p](\encodersM[t] - \noise[t][\alpha]) \approx \FK[p](\encodersM[t]) - \J[p](\encodersM[t]) \noise[t][\alpha],
\end{equation}
where $\J[p]$ denotes the analytical Jacobian of the forward kinematics function. Using the state variables, the forward-kinematics position measurement becomes
\begin{align} \label{eq:fk_measurement}
\FK[p](\encodersM[t]) = \R[t][\transpose](\d[t] - \p[t]) + \J[p](\encodersM[t]) \noise[t][\alpha].
\end{align}
Re-written in matrix form, this measurement has the right-invariant observation form \eqref{eq:invariant_observations}, 
\begin{equation*} 
\underbrace{
\begin{bmatrix}
\FK[p](\encodersM[t]) \\
0 \\
1 \\
-1
\end{bmatrix}}_{\Y[t]}
= 
\underbrace{
\begin{bmatrix}
\R[t][\transpose] & -\R[t][\transpose]\v[t] & -\R[t][\transpose]\p[t] & -\R[t][\transpose]\d[t] \\
\zeros[1,3] & 1 & 0 & 0 \\
\zeros[1,3] & 0 & 1 & 0 \\
\zeros[1,3] & 0 & 0 & 1 \\
\end{bmatrix}}_{\X[t][-1]}
\underbrace{
\begin{bmatrix}
\zeros[3,1] \\
0 \\
1 \\
-1
\end{bmatrix}}_{\b} 
+
\underbrace{
\begin{bmatrix}
\J[p](\encodersM[t]) \noise[t][\alpha] \\
0 \\
0 \\
0 \\
\end{bmatrix}}_{\V[t]}.
\end{equation*}
Therefore, the innovation depends solely on the invariant error and the update equations take the form \cite[Section 3.1.2]{barrau2017invariant}
\begin{align} \label{eq:riekf_update}
% \small
\begin{split}
\XE[t][+] &= \exp\left( \L[t] \left( \XE[t] \Y[t] - \b \right) \right) \XE[t] \\
\groupError[t][r+] &= \exp\left( \L[t] \left( \groupError[t][r] \b - \b + \XE[t] \V[t] \right)  \right) \groupError[t][r],  \\
\end{split}
\end{align}
where $\exp(\cdot)$ is the exponential map corresponding to the state matrix Lie group, $\lieGroup$, $\L[t]$ is a gain matrix to be defined later, \mbox{\small$\b[][\transpose] = \begin{bmatrix} \zeros[1,3] & 0 & 1 & -1 \end{bmatrix}$}, and \mbox{\small$\Y[t][\transpose] = \begin{bmatrix} \FK[p]^\transpose(\encodersM[t]) & 0 & 1 & -1 \end{bmatrix}$}. Because the last three rows of \mbox{\small$\XE[t] \Y[t] - \b$} are identically zero, we can express the update equations using a reduced dimensional gain, $\K[t]$, and an auxiliary selection matrix \mbox{\small$\SelectionMatrix \triangleq \begin{bmatrix} \I & \zeros[3,3] \end{bmatrix}$}, so that \mbox{\small$\L[t] \left( \XE[t] \Y[t] - \b \right) = \K[t] \SelectionMatrix \XE[t] \Y[t] $} as detailed in \citet{barrau2015non}.
  
Using the first order approximation of the exponential map, \mbox{\small$\groupError[t][r] = \exp(\tangentError[t]) \approx \textbf{\textit{I}}_d + \vectorToAlgebra[\tangentError[t]]$}, and dropping higher-order terms, we can linearize the update equation \eqref{eq:riekf_update},
\begin{equation*}
  % \resizebox{\hsize}{!}{$
  \begin{aligned}
  & \groupError[t][r+] \approx \I[d] + \vectorToAlgebra[\tangentError[t][+]] \approx \I[d] + \vectorToAlgebra[\tangentError[t]] + \vectorToAlgebra[\left({ \K[t] \SelectionMatrix \left( \left(\I[d] + \vectorToAlgebra[\tangentError[t]]\right) 
  \begin{bmatrix}
  \zeros[3,1] \\ 0 \\ 1 \\ -1 
  \end{bmatrix} 
  + \XE[t]
  \begin{bmatrix}
  \J[p](\encodersM[t]) \noise[t][\alpha] \\ 0 \\ 0 \\ 0
  \end{bmatrix}
  \right) }\right)]. 
  \end{aligned}
  % $}
  \end{equation*}
Therefore,
\begin{align*}
  \begin{split}
  \vectorToAlgebra[\tangentError[t][+]] &= \vectorToAlgebra[\tangentError[t]] +  \scriptsize{\vectorToAlgebra[\left({ 
  \K[t] \SelectionMatrix \left( 
  \begin{bmatrix}
  \I + \vectorToSkew[\tangentError[t][R]] & \tangentError[t][v] & \tangentError[t][p] & \tangentError[t][d] \\
  \zeros[3,1] & 1 & 0 & 0 \\
  \zeros[3,1] & 0 & 1 & 0  \\
  \zeros[3,1] & 0 & 0 & 1 \\
  \end{bmatrix}
  \begin{bmatrix}
  \zeros[3,1] \\ 0 \\ 1 \\ -1 
  \end{bmatrix}
  + \XE[t]
  \begin{bmatrix}
  \J[p](\encodersM[t]) \noise[t][\alpha] \\ 0 \\ 0 \\ 0
  \end{bmatrix}
  \right) }\right)]} \\
  &=  \vectorToAlgebra[\tangentError[t]] + \vectorToAlgebra[\left({ \K[t] \SelectionMatrix \left(
  \begin{bmatrix}
  \tangentError[t][p] - \tangentError[t][d] \\ 0 \\ 1 \\ -1 
  \end{bmatrix}
  + \XE[t]
  \begin{bmatrix}
  \J[p](\encodersM[t]) \noise[t][\alpha] \\ 0 \\ 0 \\ 0
  \end{bmatrix}
  \right) }\right)].
  \end{split}
  \end{align*}
Taking $\algebraToVector[(\cdot)]$ of both sides yields the linear update equation,
\begin{equation} \label{eq:linear_update}
\begin{split}
\nonumber \tangentError[t][+] &= \tangentError[t] - \K[t] \left(
\begin{bmatrix}
\zeros[3,3] & \zeros[3,3] & -\I & \I
\end{bmatrix}
\tangentError[t] - \RE[t] (\J[p](\encodersM[t]) \noise[t][\alpha])
\right) \\
&\triangleq \tangentError[t] - \K[t] \left( \H[t] \tangentError[t] - \RE[t] \left(\J[p](\encodersM[t]) \noise[t][\alpha]\right) \right).
\end{split} 
\end{equation}
Finally, we can write down the full state and covariance update equations of the \ac{RIEKF} using the derived linear update equation and the theory of Kalman filtering~\citep{maybeck1982stochastic,anderson1979optimal,bar2001estimation} as
\begin{equation}
\begin{split}
\XE[t][+] &= \exp\left( \K[t] \SelectionMatrix \XE[t] \Y[t] \right) \XE[t] \\
\P[t][+] &= (\I - \K[t] \H[t]) \P[t] (\I - \K[t] \H[t])^\transpose + \K[t] \bar{\N}_t \K[t][\transpose], 
\end{split}
\end{equation}
where the gain $\K[t]$ is computed using
\begin{align*}
\S[t] = \H[t] \P[t] \H[t][\transpose] + \bar{\N}_t
\qquad
\K[t] = \P[t] \H[t][\transpose] \S[t][-1]
\end{align*}
and from  \eqref{eq:linear_update}, the matrices $\H[t]$ and $\bar{\N}_t$ are given by
\begin{equation} \label{eq:right_invariant_kinematics_linearization}
\begin{split}
\H[t] &= 
\begin{bmatrix}
\zeros[3,3] & \zeros[3,3] & -\I & \I 
\end{bmatrix}, \\
\bar{\N}_t &= \RE[t] \; \J[p](\encodersM[t]) \; \text{Cov}(\noise[t][\alpha]) \; \J[p][\transpose](\encodersM[t]) \; \RE[t][\transpose].
\end{split}
\end{equation}

\subsection{Observability Analysis} \label{sec:observability}
Because the error dynamics are log-linear (c.f., Theorem~\ref{theorem:log_linear_error}), we can determine the unobservable states of the filter without having to perform a nonlinear observability analysis \citep{barrau2015non}. Noting that the linear error dynamics matrix in our case is time-invariant and nilpotent (with a degree of 3), the discrete-time state transition matrix is a polynomial in $\A[t]$,
\begin{equation*}
\boldsymbol{\Phi} = \exp_m(\A[t] \Delta t) = 
\begin{bmatrix}
\I & \zeros & \zeros & \zeros \\
\vectorToSkew[\g] \Delta t & \I & \zeros & \zeros \\
\dfrac{1}{2} \vectorToSkew[\g] \Delta t^2 & \I \Delta t& \I & \zeros \\
\zeros & \zeros & \zeros & \I \\
\end{bmatrix}.
\end{equation*}
It follows that the discrete-time observability matrix is
\begin{equation*}
\mathcal{O} = 
\begin{bmatrix}
\H \\
\H \boldsymbol{\Phi} \\
\H \boldsymbol{\Phi}^2 \\
\vdots
\end{bmatrix}
=
\begin{bmatrix}
\zeros & \zeros & -\I & \I \\
- \dfrac{1}{2} \vectorToSkew[\g] \Delta t^2 & -\I \Delta t & -\I & \I \\
-2 \vectorToSkew[\g] \Delta t^2 & -2 \I \Delta t^2 & -\I & \I \\
\vdots & \vdots & \vdots & \vdots
\end{bmatrix}.
\end{equation*}
The last six columns (i.e., two matrix columns) of the observability matrix are clearly linearly dependent, which indicates the absolute position of the robot is unobservable. In addition, since the gravity vector only has a $z$ component, the third column of $\mathcal{O}$ is all zeros. Therefore, a rotation about the gravity vector (yaw) is also unobservable. This linear observability analysis agrees with the nonlinear observability results of \citet{bloesch2012state}, albeit with much less computation. Furthermore, as the error dynamics do not depend on the estimated state, there is no chance of the linearization spuriously increasing the numerical rank of the observability matrix \citep{barrau2015non}. This latter effect was previously known and studied by~\citet{bloesch2012state}, and in order to resolve this problem, an observability-constrained EKF~\citep{huang2010observability} was developed. In our proposed framework, by default, the discrete \ac{RIEKF} has the same unobservable states as the underlying nonlinear system; hence, the developed discrete \ac{RIEKF} intrinsically solves this problem.

%% file: simulation_results.tex
%%%%%%%%%%%%%%%%%%%%%%%%%%%%%%%%%%%%%%%%%%%%%%%%%%%%%%%%%%%%%%%%%%%%%%%%%%%%%%%%%%%%%%%%%%%%%
\section{Simulation Results} \label{sec:sim}
\noindent
To investigate potential benefits or drawbacks of the proposed filter, we compare it against a state-of-the-art \acf{QEKF}, similar to those described by \citet{bloesch2012state,rotella2014state}. For implementation, the filter equations were discretized; see Appendix \ref{appx:discretization} for more details.

\subsection{Quaternion-Based Filter Equations:}
The choice of error variables is the main difference between the \ac{InEKF} and the \ac{QEKF}. Instead of the right-invariant error \eqref{eq:invariant_error}, a \ac{QEKF} typically uses decoupled error states
\begin{equation} \label{eq:quaternion_error_states}
\begin{split}
\exp(\delta \boldsymbol{\theta}_t) &\triangleq \R[t][\transpose]\RE[t] \\
\delta \v[t] &\triangleq \v[t] - \vE[t] \\
\delta \p[t] &\triangleq \p[t] - \pE[t]. \\
\delta \d[t] &\triangleq \d[t] - \dE[t]. \\
\end{split}
\end{equation}
Using this definition of error, the \ac{QEKF} deterministic error dynamics can be approximated as
\begin{equation*}
  \deriv 
  \begin{bmatrix}
    \delta \boldsymbol{\theta}_t \\
    \delta \v[t] \\
    \delta \p[t] \\
    \delta \d[t] \\
  \end{bmatrix} = 
  \begin{bmatrix}
    -\vectorToSkew[\wM[t]] & \zeros & \zeros & \zeros \\
    -\RE[t]\vectorToSkew[\aM[t]] & \zeros & \zeros & \zeros \\
    \zeros & \I & \zeros & \zeros \\
    \zeros & \zeros & \zeros & \zeros \\
    \end{bmatrix}
  \begin{bmatrix}
    \delta \boldsymbol{\theta}_t \\
    \delta \v[t] \\
    \delta \p[t] \\
    \delta \d[t] \\
  \end{bmatrix},
\end{equation*}
while the linearized observation matrix becomes
\begin{equation*}
  \H[t] = 
  \begin{bmatrix}
    \vectorToSkew[\RE[t][\transpose](\dE[t]-\pE[t])] & \zeros & -\RE[t][\transpose] & \RE[t][\transpose]
  \end{bmatrix}.
\end{equation*}

The above linearizations are clearly dependent on the state estimate. Therefore, when the estimated state deviates from the true state, the linearizations are potentially wrong, reducing accuracy and consistency in the \ac{QEKF}. In contrast, the deterministic right-invariant error dynamics are exactly log-linear \eqref{eq:A_right_no_bias}. In addition, the linearized observation matrix for our \ac{InEKF} \eqref{eq:right_invariant_kinematics_linearization} is also independent of the state estimate.

\begin{table}[t]
	\centering
    \caption{Experiment Discrete Noise Statistics and Initial Covariance}
    \resizebox{0.75\columnwidth}{!}{%
	{\renewcommand{\arraystretch}{1.0}%
    \begin{tabular}{cc}
      \begin{tabular}{l|l}
            \toprule
			Measurement	Type        & noise st. dev.\\
            \midrule
			Linear Acceleration	    & $0.04 ~\m/\sec^2$\\
			Angular Velocity        & $0.002 ~\rad/\sec$\\
			Accelerometer Bias      & $0.001 ~\m/\sec^3$\\
			Gyroscope Bias          & $0.001 ~\rad/\sec^2$\\
			Contact Linear Velocity & $0.05 ~\m/\sec$  \\
			Joint Encoders  & $1.0 ~\deg$  \\
            & \\
            \bottomrule
	  \end{tabular} &
      \begin{tabular}{l|l}
              \toprule
              State Element           & initial st. dev.\\
              \midrule
              Orientation	of IMU     & $30.0 ~\deg$ \\
              Velocity of IMU   	   & $1.0 ~\m/\sec$\\
              Position of IMU        & $0.1 ~\m$\\
              Position of Right Foot & $0.1 ~\m$\\
              Position of Left Foot  & $0.1 ~\m$  \\
              Gyroscope Bias         & $0.005 ~\rad/\sec$\\
              Accelerometer Bias     & $0.05 ~\m/\sec^2$\\
              \bottomrule
      \end{tabular}
    \end{tabular}
    }}
\label{tab:params}
\end{table}

%%%%%%%%%%%%%%%%%%%%%%%%%%%%%%%%%%%%%%%%%%%%%%%%%%%%%%%%%%%%%%%%
\subsection{Convergence Comparison:}
\begin{figure}[h!]
  \centering
    \includegraphics[width=0.75\columnwidth]{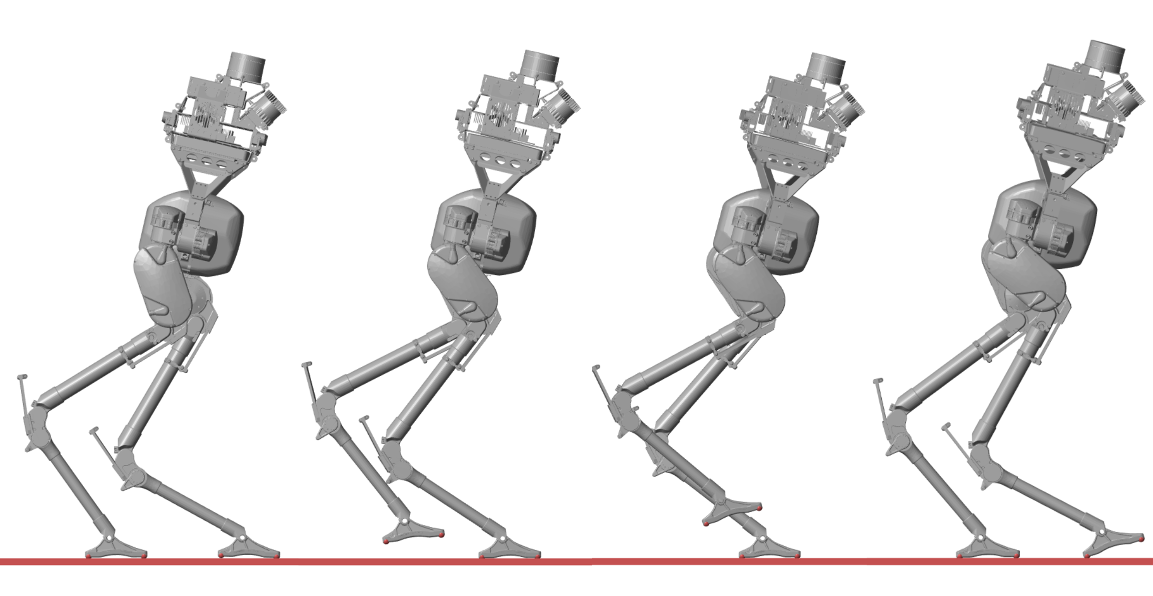}
      \caption{A typical walking gait that is used for filter comparisons. The Cassie bipedal robot is simulated using Simscape Multibody\texttrademark.}
\label{fig:cassie_sim_walking}
\end{figure}

A dynamic \textit{simulation} of a Cassie-series bipedal robot (described in Section \ref{sec:experimental_results}) was performed in which the robot slowly walked forward after a small drop, accelerating from $0.0$ to $0.3~\m/\sec$. The discrete, simulated measurements were corrupted by additive white Gaussian noise, which are specified in Table \ref{tab:params} along with the initial state covariance values. The same values were used in both simulation and experimental convergence evaluations of the filters. The IMU bias estimation was turned off for these simulations. The simulation was performed with MATLAB and Simulink (Simscape Multibody\texttrademark) where the simulation environment models ground contact forces with a linear force law (having a stiffness and damping term) and a Coulomb friction model. A typical walking gait is shown in Figure \ref{fig:cassie_sim_walking}.

\begin{figure*}[t!]
  \centering
    \includegraphics[width=\textwidth]{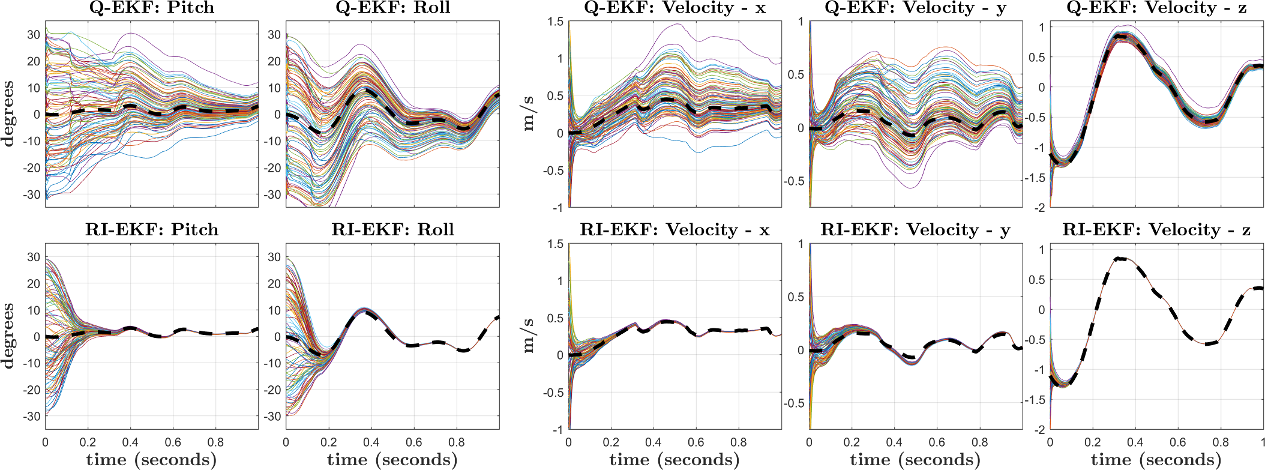}
      \caption{A \acf{QEKF} and the proposed \acf{RIEKF} were run 100 times using the same measurements, noise statistics, and initial covariance, but with random initial orientations and velocities. The noisy measurements came from a dynamic simulation of a Cassie-series biped robot where the robot walks forwards after a small drop, accelerating from $0.0$ to $0.3~\m/\sec$. The above plots show the state estimate for the initial second of data, where the dashed black line represents the true state. The \ac{RIEKF} (bottom row) converges considerably faster than the \ac{QEKF} (top row) for all observable states. The estimated yaw angle (not shown) does not converge for either filter because it is unobservable. Therefore, to compare convergence, the velocities shown are represented in the estimated IMU (body) frame.}
\label{fig:sim_comparison}
\end{figure*}

To compare the convergence properties of the two filters, 100 simulations of each filter were performed using identical measurements, noise statistics, initial covariance, and various random initial orientations and velocities. The initial Euler angle estimates were sampled uniformly from $-30\deg$ to $30\deg$. The initial velocity estimates were sampled uniformly from $-1.0~\m/\sec$ to $1.0~\m/\sec$. The pitch and roll estimates as well as the (body frame) velocity estimates for both filters are shown in Figure~\ref{fig:sim_comparison}. Although both filters converge for this set of initial conditions, the proposed \ac{RIEKF} converges considerably faster than the standard quaternion-based EKF. 

%%%%%%%%%%%%%%%%%%%%%%%%%%%%%%%%%%%%%%%%%%%%%%%%%%%%%%%%%%%%%%%%
\subsection{Accuracy of Linearized Dynamics}
The superior performance of the \ac{InEKF} over the \ac{QEKF} comes from the improved accuracy of the linearized error dynamics. As indicated by Theorem~\ref{theorem:log_linear_error}, the deterministic error dynamics of the \ac{InEKF} are actually exact, while the \ac{QEKF} version is only an approximation. To demonstrate this, a simulation was performed where propagation of the true error is compared to the propagation of the linearized error dynamics.

\begin{figure}[!h]
  \centering
  \begin{minipage}{.49\textwidth}
      \centering
      \vspace{0.4cm}
      \includegraphics[width=0.99\textwidth]{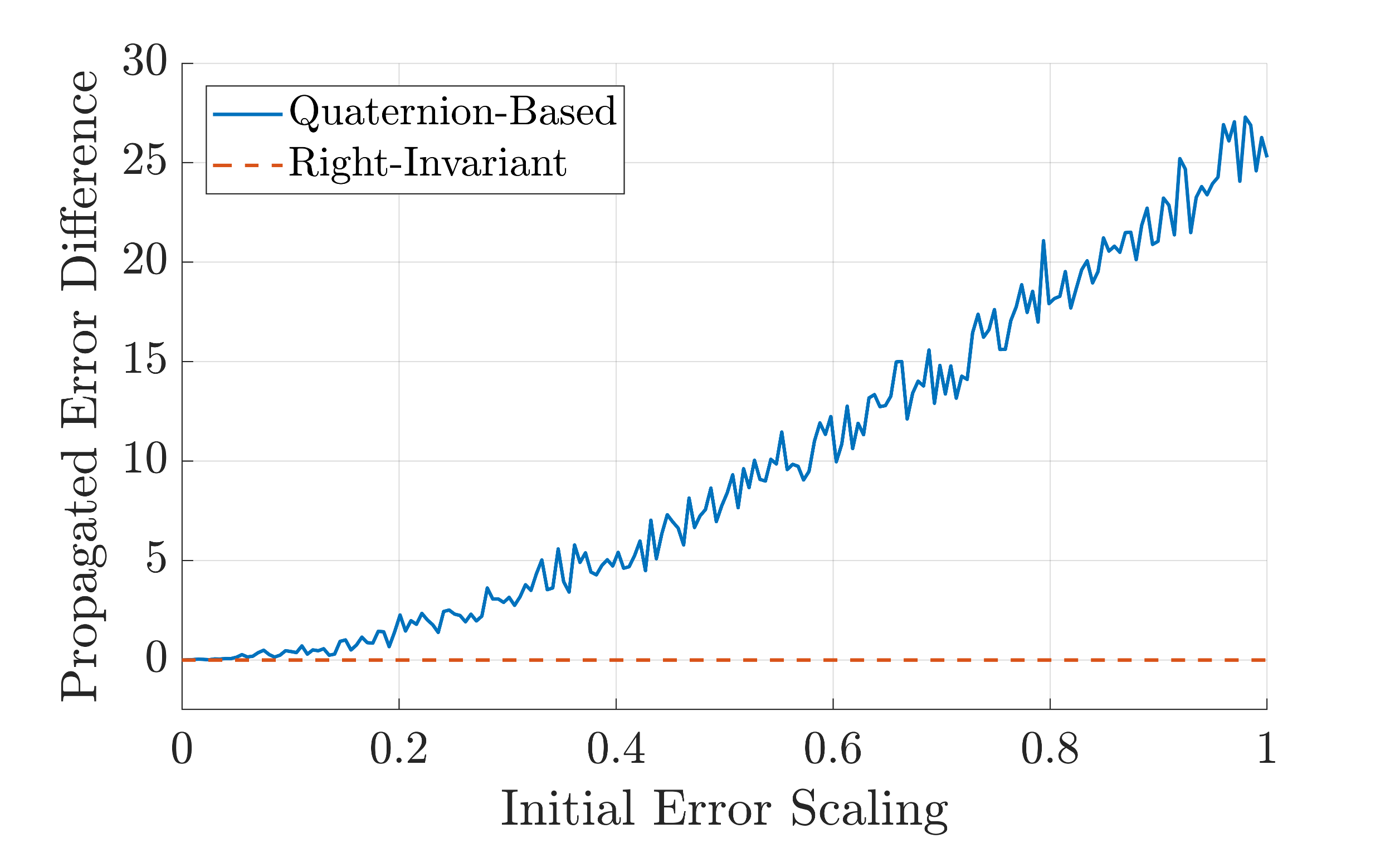}
      \caption{Analyzing accuracy of the deterministic error dynamics. This Figure shows the difference between the true error and the propagated error as the initial true error increases. The state and errors were propagated for 1 second using randomly sampled \ac{IMU} measurements.}
      \label{fig:deterministic_error_propagation}
  \end{minipage}%
  \hfill
  \begin{minipage}{0.49\textwidth}
      \centering
      \includegraphics[width=0.99\textwidth]{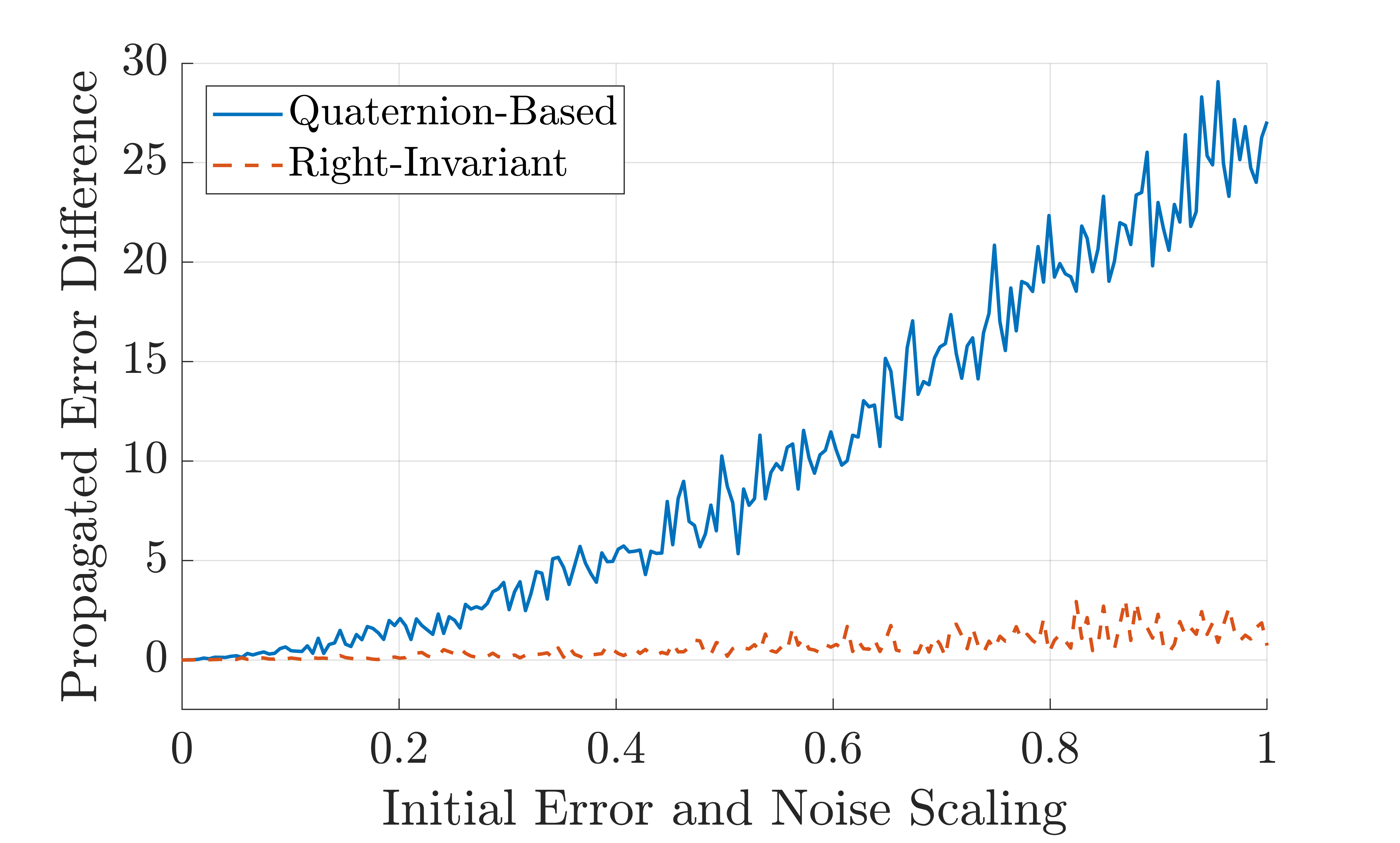}
      \caption{Difference between the true and propagated errors when measurements contain noise. The log-linear error dynamics of the \ac{InEKF} are no longer exact.}
      \label{fig:noisy_error_propagation}
  \end{minipage}
\end{figure}

We first analyzed the deterministic dynamics. Given an initial error vector, $\tangentError[0][\text{true}]$, the initial state estimate for the \ac{InEKF} was computed using the definition of right-invariant error \eqref{eq:invariant_error}, and the initial state estimate for the \ac{QEKF} was computed using equation \eqref{eq:quaternion_error_states}. The true state was initialized to the identity element. The true and estimated states for both filters were then propagated for 1 second (1000 time steps), using randomly sampled \ac{IMU} measurements. The error states for both the \ac{InEKF} and the \ac{QEKF} were also propagated using their respective linearized error dynamics. The resulting error, $\tangentError[1][\text{true}]$, between the final estimated and true states were computed and compared to the propagated error states, $\tangentError[1][\text{prop}]$ to yield a measure of linearization accuracy, $\lVert\tangentError[1][\text{true}]-\tangentError[1][\text{prop}]\rVert$. This test was performed multiple times while linearly scaling the initial orientation error from $\vector[0,0,0]$ to $\vector[\pi/2,\pi/2,\pi/2]$. The results are shown in Figure~\ref{fig:deterministic_error_propagation}.
% \begin{figure}[h!]
%   \centering
%     \includegraphics[width=\columnwidth]{graphics/deterministic_error_propagation}
%       \caption{Analyzing accuracy of the deterministic error dynamics. This Figure shows the difference between the true error and the propagated error as the initial true error increases. The state and errors were propagated for 1 second using randomly sampled \ac{IMU} measurements.}
% \label{fig:deterministic_error_propagation}
% \end{figure}

% \FloatBarrier
As expected, when the initial error is zero, the difference between the true and propagated error is zero. This indicates that the linearized error dynamics for both the \ac{InEKF} and the \ac{QEKF} are correct. As the initial error increases, the difference between the true and propagated error states for the \ac{QEKF} grows due to the decreased accuracy of the linearization. In contrast, the difference between the true and propagated error states for the \ac{InEKF} are always exactly zero regardless of the initial error. In other words, assuming the initial error is known, the true propagated state can be exactly recovered from solving the linearized error dynamics system; see Theorem~\ref{theorem:log_linear_error}. 

In the non-deterministic case (with sensor noise), Theorem~\ref{theorem:log_linear_error} no longer holds. This can be seen in Figure~\ref{fig:noisy_error_propagation}, where the same test was performed, but with sensor noise corrupting the propagated state estimate. The difference between the true and propagated error is no longer exactly zero for the \ac{InEKF}.
% \begin{figure}[h!]
%   \centering
%     \includegraphics[width=\columnwidth]{graphics/noisy_error_propagation}
%       \caption{Difference between the true and propagated errors when measurements contain noise. The log-linear error dynamics of the \ac{InEKF} are no longer exact.}
% \label{fig:noisy_error_propagation}
% \end{figure}
%\FloatBarrier
However, the \ac{InEKF} linearization remains more accurate due to the reduced sensitivity to initial state errors. This helps to further explain the improved convergence properties shown in Figure~\ref{fig:sim_comparison}.

%%%%%%%%%%%%%%%%%%%%%%%%%%%%%%%%%%%%%%%%%%%%%%%
\subsection{Covariance Ellipse Comparison}
The error states in both the \ac{QEKF} and the \ac{InEKF} are assumed to be zero-mean Gaussian random vectors. However, due to the differing choice of error variables, the state uncertainty will differ. In the \ac{QEKF}, all states and errors are decoupled \eqref{eq:quaternion_error_states}. For example, the true position only depends on the position estimate and the position error, $\p = \pE + \delta \p$. Therefore, the position estimate is a Gaussian centered at $\pE$. In contrast, when using the \ac{InEKF}, the position and orientation are actually coupled together, $\X = \exp(\tangentError) \XE$. Although $\tangentError$ is a Gaussian random vector, after applying the group's exponential map and matrix multiplication, the state estimate's uncertainty distribution is no longer Gaussian. This distribution is known as a \emph{concentrated Gaussian on a Lie group} \citep{wang2006error,wang2008nonparametric}. This type of distribution can often capture the underlying system uncertainty better than a standard Gaussian defined in Euclidean space \citep{long2013banana, barfoot2014associating}.

\begin{figure}[h!]
  \centering
    \includegraphics[trim={1cm 1cm 1cm 1cm},clip,width=0.65\columnwidth]{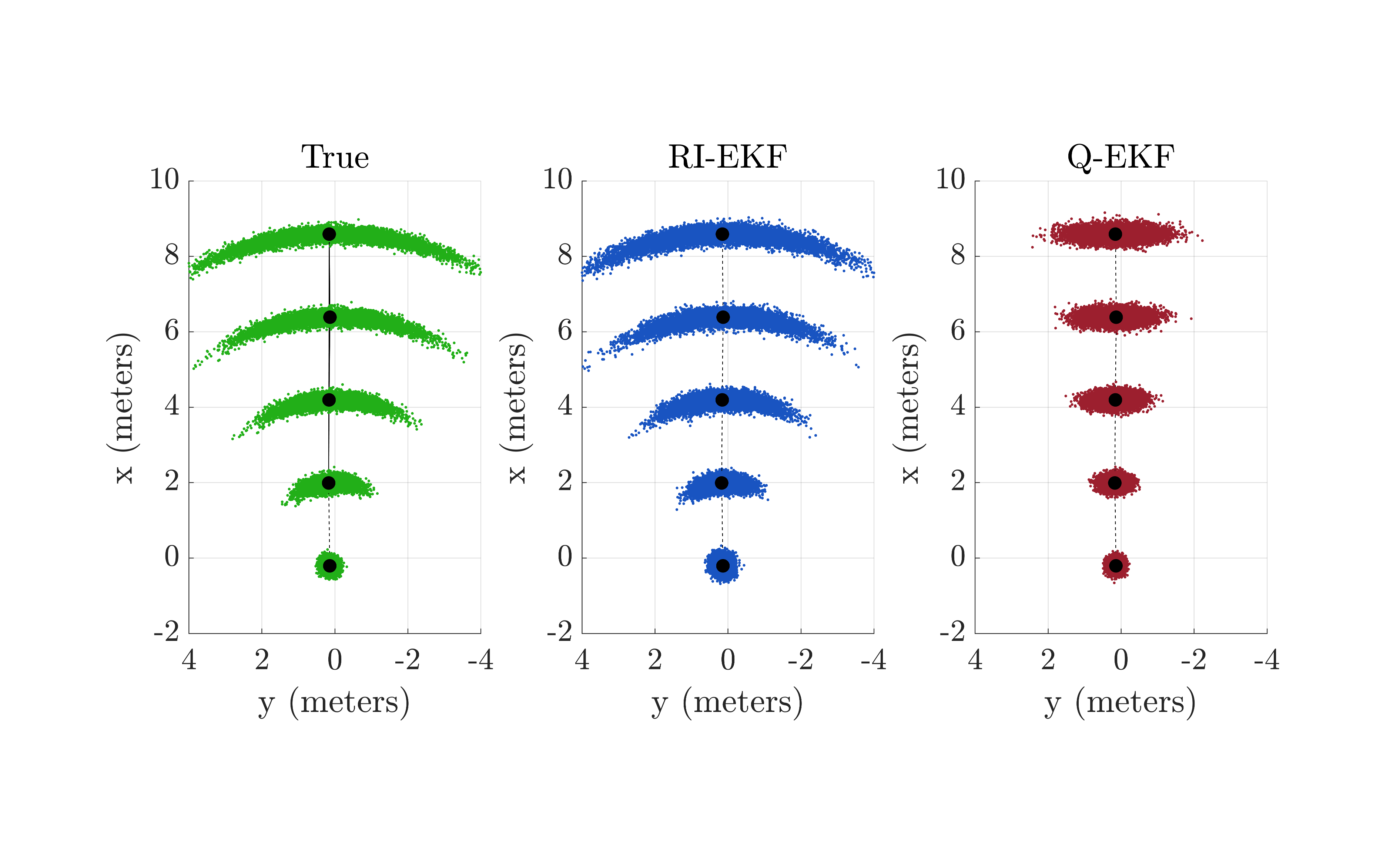}
      \caption{10,000 samples taken from the estimated filter covariances for a simulation where Cassie walked forward with an average speed of $1~m/s$. The position distributions at times 0, 2, 4, 6, and 8~$\sec$ are shown.}
\label{fig:wifi} 
\end{figure}

To demonstrate the difference, a simple simulation was performed where the Cassie robot walked forward for $8~\sec$ at an average speed of $1~m/s$. The standard deviation for the initial position uncertainty was set to $0.1~\m$ about each axis, while the standard deviation of the initial yaw uncertainty was set to $10~\deg$. A set of 10,000 particles sampled from this distribution were propagated forward to represent the robot's true uncertainty distribution. After running both the \ac{InEKF} and the \ac{QEKF}, particles were sampled from the resulting filter covariances to provide a picture of the estimated position uncertainties. This result is shown in Figure~\ref{fig:wifi}.
The curved position distribution comes from the initial yaw uncertainty that continually grows due to its unobservability (Section~\ref{sec:observability}). The \ac{InEKF} is able to closely match this distribution since the samples are taken in the Lie algebra and are mapped to the group through the exponential map. This couples the orientation and position errors leading to a curved position distribution. In contrast, the \ac{QEKF} position uncertainty can only have the shape of the standard Gaussian ellipse, which may not represent the true uncertainty well. 

\begin{figure}[h!]
  \centering
    \includegraphics[trim={1cm 0cm 1cm 0cm},clip,width=0.6\columnwidth]{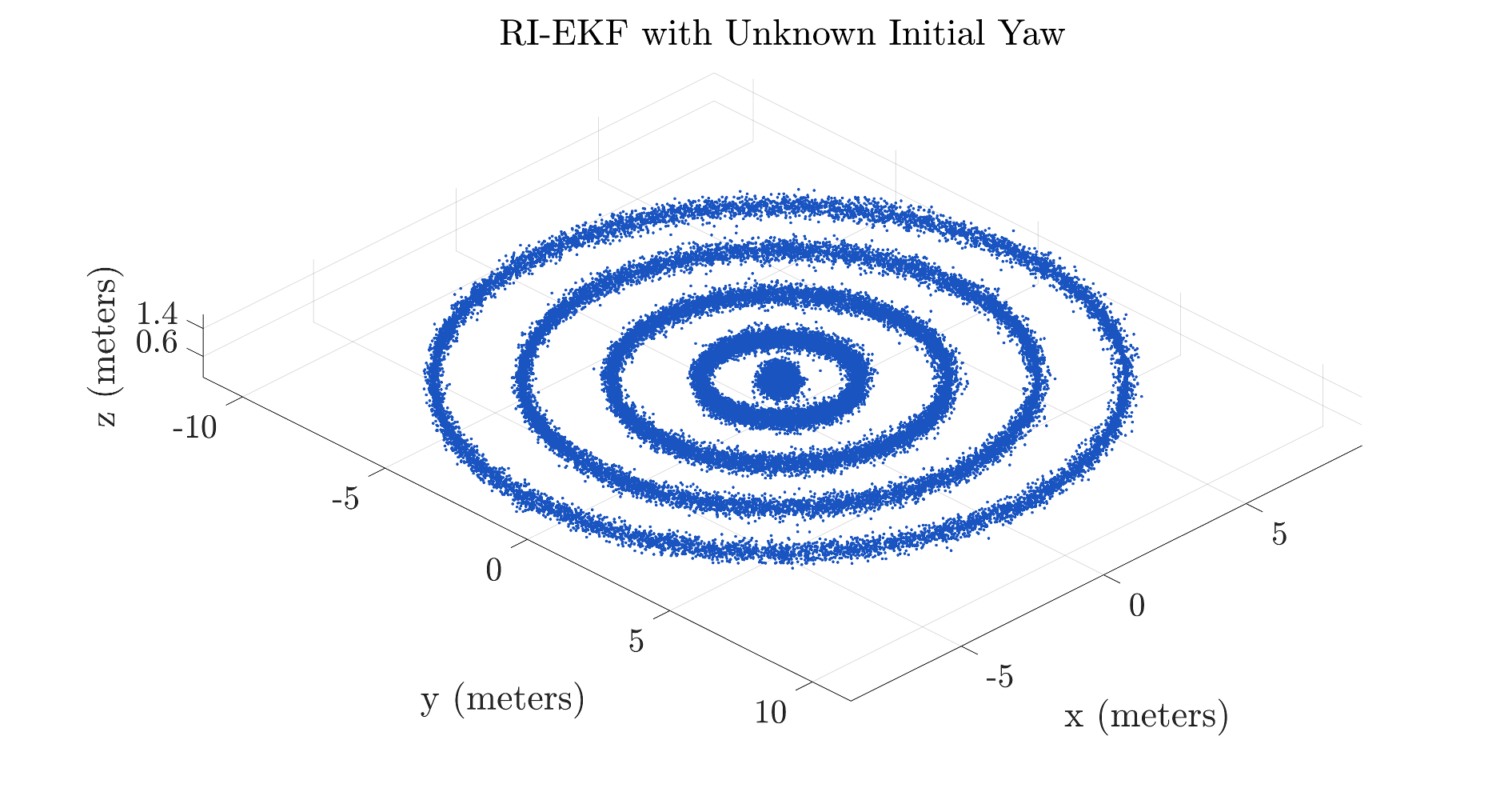}
      \caption{Samples taken from the \ac{InEKF}'s estimated position distribution for a walking simulation with a completely uncertain initial yaw angle. The robot moved forward at an average speed of 1~$m/s$. Each ring represents the position uncertainty at times 0, 2, 4, 6, and 8~$\sec$.}
\label{fig:bullseye}
\end{figure}

The \ac{InEKF} can even accurately model the case of complete yaw uncertainty. To demonstrate this, the initial yaw standard deviation was set to $360~\deg$, and the same $8~\sec$ simulation was performed. Each ring in Figure~\ref{fig:bullseye} shows the sample position distribution spaced $2~\sec$ apart. This type of uncertainty cannot be captured with a standard Gaussian covariance ellipse. These examples illustrate that even if the means are identical, the covariance estimate of the \ac{InEKF} can provide a more accurate representation of the state's uncertainty than the standard \ac{QEKF}. 

In these examples, the initial covariance of all states (except yaw) was small. It is interesting to note what happens if we remove the dynamics noise and set the covariance for some states to exactly zero (rank-deficient covariance matrix). In this case, the initial covariance is supported by a subgroup and the \ac{InEKF} will keep the state estimate within a time-dependent subset of the Lie group at all times. This theoretical result was proved by \citet{chauchat2017kalman,barrau2019extended} to create \acp{EKF} with state equality constraints.

%% file: bias.tex
\section{IMU bias augmentation}
\label{sec:bias}
Implementation of an IMU-based state estimator on hardware typically requires modeling additional states, such as gyroscope and accelerometer biases. Unfortunately, as noted in~\citet{barrau2015non}, there is no Lie group that includes the bias terms while also having the dynamics satisfy the group affine property \eqref{eq:group_affine}. Even though many of the theoretical properties of the  \ac{RIEKF} will no longer hold, it is possible to design an ``imperfect \ac{InEKF}'' that still outperforms the standard EKF~\citep{barrau2015non}.

\subsection{State Representation}
The IMU biases are slowly varying signals that corrupt the measurements in an additive manner:
\begin{alignat*}{2}
\wM[t] &= \w[t] + \b[t][g] + \noise[t][g], \qquad &&\noise[t][g] \sim \mathcal{GP}\left(\zeros[3,1], \Cov[][g]\,\delta(t - t^\prime)\right)  \\
\aM[t] &= \a[t] + \b[t][a] + \noise[t][a], \qquad &&\noise[t][a] \sim \mathcal{GP}\left(\zeros[3,1], \Cov[][a]\,\delta(t - t^\prime)\right). 
\end{alignat*}
These biases form a parameter vector that needs to be estimated as part of the \ac{RIEKF} state, 
\begin{equation}
% %\small
\begin{split}
\params[t] &\triangleq 
\begin{bmatrix}
\gyroscopeBias(t) \\
\accelerometerBias(t) \\
\end{bmatrix} 
\triangleq 
\begin{bmatrix}
\gyroscopeBias[t] \\
\accelerometerBias[t] \\
\end{bmatrix} \in \realnumbers^6.
\end{split}
\end{equation}
The model's state now becomes a tuple of our original matrix Lie group and the parameter vector, \mbox{$(\X[t], \params[t]) \in \lieGroup \times \realnumbers^6$}. The augmented right-invariant error is now defined as
\begin{equation} \label{eq:augmented_error}
\e[t][r] \triangleq (\XE[t] \X[t][-1], \paramsE[t] - \params[t]) \triangleq (\groupError[t][r], \paramError[t]).
\end{equation}
Written explicitly, the right-invariant error is
\begin{equation*}
\groupError[t][r] = 
\begin{bmatrix}
\RE[t] \R[t][\transpose] 
& \vE[t] - \RE[t] \R[t][\transpose] \v[t] 
& \pE[t] - \RE[t] \R[t][\transpose] \p[t] 
& \dE[t] - \RE[t] \R[t][\transpose] \d[t] \\
\zeros[1,3] & 1 & 0 & 0 \\
\zeros[1,3] & 0 & 1 & 0 \\
\zeros[1,3] & 0 & 0 & 1 \\
\end{bmatrix},
\end{equation*}
while the parameter vector error is given by
\begin{equation*}
\paramError[t] = 
\begin{bmatrix}
\gyroscopeBiasE[t] - \gyroscopeBias[t] \\
\accelerometerBiasE[t] - \accelerometerBias[t] \\
\end{bmatrix} 
\triangleq
\begin{bmatrix}
\paramError[t][g] \\
\paramError[t][a] 
\end{bmatrix}.
\end{equation*}

%%%%%%%%%%%%%%%%%%%%%%%%%%%%%%%%%%%%%%%%%%%%%%%%%%%%%%
\subsection{System Dynamics}

With \ac{IMU} biases included, the system dynamics are now expressed as
\begin{equation}\label{eq:world_dynamics_with_bias}
  \begin{split}
  \deriv \R[t] &= \R[t]\vectorToSkew[\wM[t] - \gyroscopeBias[t] - \noise[t][g]] \\ %, \quad 
  \deriv \v[t] &= \R[t](\aM[t] - \accelerometerBias[t] - \noise[t][a]) + \g \\
  \deriv \p[t] &= \v[t] \\ %, \qquad \qquad \qquad~~
  \deriv \d[t] &= \R[t] \FK[R](\encodersM[t])(-\noise[t][v]).
  \end{split}
\end{equation}

The IMU bias dynamics are modeled using the typical ``Brownian motion'' model, i.e., the derivatives are white Gaussian noise, to capture the slowly time-varying nature of these parameters,
\begin{equation}\label{eq:bias_dynamics}
\begin{aligned}
  \deriv \gyroscopeBias[t] &= \noise[t][bg] , \quad &&\noise[t][bg] \sim \mathcal{GP}(\zeros[3,1], \Cov[][bg]\,\delta(t - t^\prime))  \\
  \deriv \accelerometerBias[t] &= \noise[t][ba], \quad &&\noise[t][ba] \sim \mathcal{GP}(\zeros[3,1], \Cov[][ba]\,\delta(t - t^\prime)) . 
\end{aligned}
\end{equation}
The deterministic system dynamics now depend on both the inputs, $\u[t]$, and the parameters, $\params[t]$
\begin{equation*} 
f_{\u[t]}(\XE[t],\paramsE[t]) =
\begin{bmatrix}
  \RE[t]\vectorToSkew[\wB[t]] 
& \RE[t]\aB[t] + \g  
& \vE[t]
& \zeros[3,1] \\
\zeros[1,3] & 0 & 0 & 0  \\
\zeros[1,3] & 0 & 0 & 0  \\
\zeros[1,3] & 0 & 0 & 0  \\
\end{bmatrix},
\end{equation*}
where $\wB[t] \triangleq \wM[t] - \gyroscopeBiasE[t]$ and $\aB[t] \triangleq \aM[t] - \accelerometerBiasE[t]$ are the ``bias-corrected'' inputs. To compute the linearized error dynamics, the augmented right-invariant error \eqref{eq:augmented_error} is first differentiated with respect to time, 
\begin{equation}
\begin{split}
\deriv \e[t][r] &= \left( 
\deriv \groupError[t][r],
\begin{bmatrix}
\noise[t][bg] \\
\noise[t][ba] \\
\end{bmatrix}
\right).
\end{split}
\end{equation}
After carrying out the chain rule and making the first order approximation, \mbox{\small$\groupError[t][r] = \exp(\tangentError[t]) \approx \textbf{\textit{I}}_d + \vectorToAlgebra[\tangentError[t]]$}, the individual terms of the invariant error dynamics become
\begin{equation} \label{eq:augmented_error_dynamics}
  \begin{split}
  \deriv \left( \RE[t] \R[t][\transpose] \right) & \approx \vectorToSkew[\RE[t] \left(\noise[t][g] -\paramError[t][g]\right)] \\
  \deriv \left( \vE[t] - \RE[t] \R[t][\transpose] \v[t] \right) &\approx  \vectorToSkew[\g] \tangentError[t][R] + \vectorToSkew[\vE[t]] \RE[t](\noise[t][g]-\paramError[t][g]) + \RE[t](\noise[t][a]-\paramError[t][a])  \\
  \deriv \left( \pE[t] - \RE[t] \R[t][\transpose] \p[t] \right) &\approx \tangentError[t][v] + \vectorToSkew[\pE[t]] \RE[t](\noise[t][g]-\paramError[t][g])\\
  \deriv \left( \dE[t] - \RE[t] \R[t][\transpose] \d[t] \right) &\approx \vectorToSkew[\dE[t]] \RE[t](\noise[t][g]-\paramError[t][g]) + \RE[t] \FK[R](\encodersM[t]) \noise[t][v]. 
  \end{split}%
  \end{equation}
Importantly, the augmented invariant error dynamics only depends on the estimated trajectory though the noise and bias errors, $\paramError[t]$ (this is expected because when there are no bias errors, there is no dependence on the estimated trajectory). A linear system can now be constructed from \eqref{eq:augmented_error_dynamics} to yield,
\begin{equation*}
\deriv 
\begin{bmatrix}
\tangentError[t] \\
\paramError[t]
\end{bmatrix} 
= \A[t] \begin{bmatrix}
\tangentError[t] \\
\paramError[t]
\end{bmatrix} +
\begin{bmatrix}
\Adjoint[\XE[t]] & \zeros[12,6] \\
\zeros[6,12] & \I[6]
\end{bmatrix}
\noise[t],
\end{equation*}
where the noise vector is augmented to include the bias terms,
$$\noise[t] \triangleq \vector[\noise[t][g],\, \noise[t][a],\, \zeros[3,1], \FK[R](\encodersM[t]) \noise[t][v], \noise[t][bg], \noise[t][ba]].$$

\subsection{Forward Kinematic Measurements}
The forward kinematics position measurement \eqref{eq:fk_measurement} does not depend on the IMU biases. Therefore, the $\H[t]$ matrix can simply be appended with zeros to account for the augmented variables. The linear update equation becomes
\begin{equation*}
\begin{bmatrix}
\tangentError[t][+] \\
\paramError[t][+]
\end{bmatrix} =
\begin{bmatrix}
\tangentError[t] \\
\paramError[t]
\end{bmatrix}
- 
\begin{bmatrix}
\K[t][\tangentError] \\
\K[t][\paramError]
\end{bmatrix}
\left( \H[t]
\begin{bmatrix}
\tangentError[t] \\
\paramError[t]
\end{bmatrix}
- \RE[t] \left(\J[p](\encodersM[t]) \noise[t][\alpha]\right) \right).
  \end{equation*}

\subsection{Final Continuous  \ac{RIEKF} Equations}
The final ``imperfect'' \ac{RIEKF} equations that include IMU biases can now be written down. The estimated state tuple is predicted using the following set of differential equations:
\begin{equation*}
\deriv \left( \XE[t] \,, \paramsE[t] \right) = \left( f_{\u[t]}(\XE[t],\paramsE[t]), \zeros[6,1] \right).
\end{equation*}
The covariance of the augmented right invariant error dynamics is computed by solving the Riccati equation
\begin{equation*}
\deriv \P[t] = \A[t] \P[t] + \P[t] \A[t][\transpose] + \bar{\Q}_t,
\end{equation*}
where the matrices $\A[t]$ and $\bar{\Q}_t$ are now defined using \eqref{eq:augmented_error_dynamics},
\begin{equation} \label{eq:linearization_bias_right}
\begin{split}
\A[t] &= 
\begin{bmatrix}
\zeros & \zeros & \zeros & \zeros & -\RE[t] & \zeros \\
\vectorToSkew[\g] & \zeros & \zeros & \zeros & -\vectorToSkew[\vE[t]]\RE[t] & -\RE[t] \\
\zeros & \I & \zeros & \zeros &  -\vectorToSkew[\pE[t]]\RE[t] & \zeros \\
\zeros & \zeros & \zeros & \zeros & -\vectorToSkew[\dE[t]] \RE[t] & \zeros \\
\zeros & \zeros & \zeros & \zeros & \zeros & \zeros \\
\zeros & \zeros & \zeros & \zeros & \zeros & \zeros \\
\end{bmatrix}  \\
\bar{\Q}_t &= 
\begin{bmatrix}
\Adjoint[\XE[t]] & \zeros[12,6] \\
\zeros[6,12] & \I[6]
\end{bmatrix}
\text{Cov}(\noise[t])
\begin{bmatrix}
\Adjoint[\XE[t]] & \zeros[12,6] \\
\zeros[6,12] & \I[6]
\end{bmatrix}^\transpose. \\ 
\end{split}
\end{equation}
The estimated state tuple and its covariance are corrected though the update equations
\begin{equation} \label{eq:right_invariant_update}
\begin{split}
\left(\XE[t][+], \params[t][+]\right) &= \left( \exp\left( \K[t][\tangentError] \SelectionMatrix \XE[t] \Y[t] \right) \XE[t]\, , \;\; \paramsE[t] + \K[t][\paramError] \SelectionMatrix \XE[t] \Y[t] \right) \\
\P[t][+] &= (\I - \K[t] \H[t]) \P[t] (\I - \K[t] \H[t])^\transpose + \K[t] \bar{\N}_t \K[t][\transpose], 
\end{split}
\end{equation}
where the gains $\K[t][\tangentError]$ and $\K[t][\paramError]$ are computed from
\begin{align*}
\S[t] = \H[t] \P[t] \H[t][\transpose] + \bar{\N}_t
\qquad
\K[t] = 
\begin{bmatrix}
\K[t][\tangentError] \\ 
\K[t][\paramError]
\end{bmatrix} = \P[t] \H[t][\transpose] \S[t][-1],
\end{align*}
with the following measurement, output, and noise matrices,
\begin{equation}
\begin{split}
\Y[t][\transpose] &=
\begin{bmatrix}
\FK[p]^\transpose(\encodersM[t]) & 0 & 1 & -1
\end{bmatrix}, \\
\nonumber \H[t] &= 
\begin{bmatrix}
\zeros & \zeros & -\I & \I & \zeros & \zeros
\end{bmatrix}, \\
\bar{\N}_t &= \RE[t] \; \J[p](\encodersM[t]) \; \text{Cov}(\noise[t][\alpha]) \; \J[p][\transpose](\encodersM[t]) \; \RE[t][\transpose].
\end{split}
\end{equation}

\begin{remark}
The upper-right block of the new linearized dynamics matrix \eqref{eq:linearization_bias_right} is related to the adjoint of the current state estimate \eqref{eq:adjoint}. Intuitively, this maps the bias error (measured in the body frame) to the world frame.
\end{remark}

%% file: switch.tex
\section{Addition and Removal of Contact Points}
\label{sec:switchcontact}
Sections \ref{sec:riekf} and \ref{sec:bias} derived the equations for the \ac{RIEKF} under the assumption that the contact point is unchanging with time. However, for legged robots, contacts are discrete events that are created and broken as a robot navigates through the environment. Therefore, it is important to be able to conveniently add and remove contact points states to and from the observer's state.

\subsection{Removing Contact Points}
\label{sec:removecontact}
To remove a previous contact point from the state, we marginalize the corresponding state variable by simply removing the corresponding column and row from the matrix Lie group. The corresponding elements of the covariance matrix are also eliminated. This can be done through a simple linear transformation. For example, if the robot is going from one contact to zero contacts, then the newly reduced covariance would be computed by
\begin{equation}
\begin{split}
\begin{bmatrix}
\tangentError[t][R] \\
\tangentError[t][v] \\
\tangentError[t][p] \\
\end{bmatrix} &=
\begin{bmatrix}
\I & \zeros & \zeros & \zeros \\
\zeros & \I & \zeros & \zeros \\
\zeros & \zeros & \I & \zeros \\
\end{bmatrix}
\begin{bmatrix}
\tangentError[t][R] \\
\tangentError[t][v] \\
\tangentError[t][p] \\
\tangentError[t][d] \\
\end{bmatrix} \\
\tangentError[t][\mathrm{new}] &\triangleq \M \, \tangentError[t] \\
\implies \P[t][\mathrm{new}] &= \M \, \P[t] \, \M[][\transpose].
\end{split}
\end{equation}

\begin{remark}
This marginalization matrix, $\M$, does not depend on the choice of right or left invariant error.
\end{remark}

\subsection{Adding Contact Points}
When the robot makes a new contact with the environment, the state and covariance matrices need to be augmented. Special attention needs to be given to initialize the mean and covariance for the new estimated contact point. For example, if the robot is going from zero contacts to one contact, the initial mean is obtained though the forward kinematics relation
\begin{equation} \label{eq:contact_augmentation}
\dE[t] = \pE[t] + \RE[t] \FK[p](\encodersM[t]). 
\end{equation}
In order to compute the new covariance, we need to look at the right-invariant error, 
\begin{equation*}
\begin{split}
  \groupError[t][d] &= \dE[t] - \RE[t]\R[t][\transpose]\d[t] \\
   &= \pE[t] + \RE[t] \FK[p](\encodersM[t]) - \RE[t]\R[t][\transpose]\d[t]  \\
   &= \pE[t] + \RE[t] \FK[p](\encodersM[t]) - \RE[t]\R[t][\transpose] \left( \p[t] + \R[t] \FK[p](\encodersM[t] - \noise[t][\alpha]) \right) \\
   &\approx \groupError[t][p] + \RE[t] \J[p](\encodersM[t]) \noise[t][\alpha] \\
\implies \tangentError[t][d] &\approx \tangentError[t][p] + \RE[t] \J[p](\encodersM[t]) \noise[t][\alpha].
\end{split}
\end{equation*}
Therefore, covariance augmentation can be done using the following linear map,
\begin{equation}
\begin{split}
\begin{bmatrix}
\tangentError[t][R] \\
\tangentError[t][v] \\
\tangentError[t][p] \\
\tangentError[t][d] \\
\end{bmatrix} &=
\begin{bmatrix}
\I & \zeros & \zeros \\
\zeros & \I & \zeros \\
\zeros & \zeros & \I \\
\zeros & \zeros & \I \\
\end{bmatrix}
\begin{bmatrix}
\tangentError[t][R] \\
\tangentError[t][v] \\
\tangentError[t][p] \\
\end{bmatrix} + 
\begin{bmatrix}
\zeros \\
\zeros \\
\zeros \\
\RE[t] \J[p](\encodersM[t]) \\
\end{bmatrix} \noise[t][\alpha] \\
\tangentError[t][\mathrm{new}] &\triangleq \F[t] \, \tangentError[t] + \G[t] \noise[t][\alpha] \\
\implies \P[t][\mathrm{new}] &= \F[t] \, \P[t] \, \F[t][\transpose] + \G[t] \, \text{Cov}(\noise[t][\alpha]) \, \G[t][\transpose]. 
\end{split}
\end{equation}

\begin{remark}
The error augmentation matrix, $\F[t]$, and the noise matrix, $\G[t]$, will depend on the choice of error variable. Here they are derived for the right invariant error case. The matrices will differ in the left invariant error formulation, as detailed in Section \ref{sec:liekf}.
\end{remark}

%% file: experimental_results.tex
\section{Experimental Results on Cassie Robot}
\label{sec:experimental_results}
We now present an experimental evaluation of the proposed contact-aided \ac{RIEKF} observer using a 3D bipedal robot. The Cassie-series robot, shown in Figure~\ref{fig:cassie}, developed by Agility Robotics, has 20 degrees of freedom coming from the body pose, 10 actuators, and 4 springs. The robot is equipped with an IMU along with 14 joint encoders that can measure all actuator and spring angles. The proposed and baseline algorithms (along with the robot's feedback controller) are implemented in MATLAB (Simulink Real-Time). The IMU (VectorNav-100) is located in the robot's torso and provides angular velocity and linear acceleration measurements at $800 \Hz$. The encoders provide joint angle measurements at $2000 \Hz$. The robot has two springs on each leg that are compressed when the robot is standing on the ground. The spring deflections are measured by encoders and serve as a binary contact sensor. The controller used for these experiments was developed by~\citet{gong2019feedback}.

% \rha{Insert picture of cassie with torso}
\subsection{Convergence Comparison}
\begin{figure*}[t!]
    \centering
      \includegraphics[width=1\textwidth]{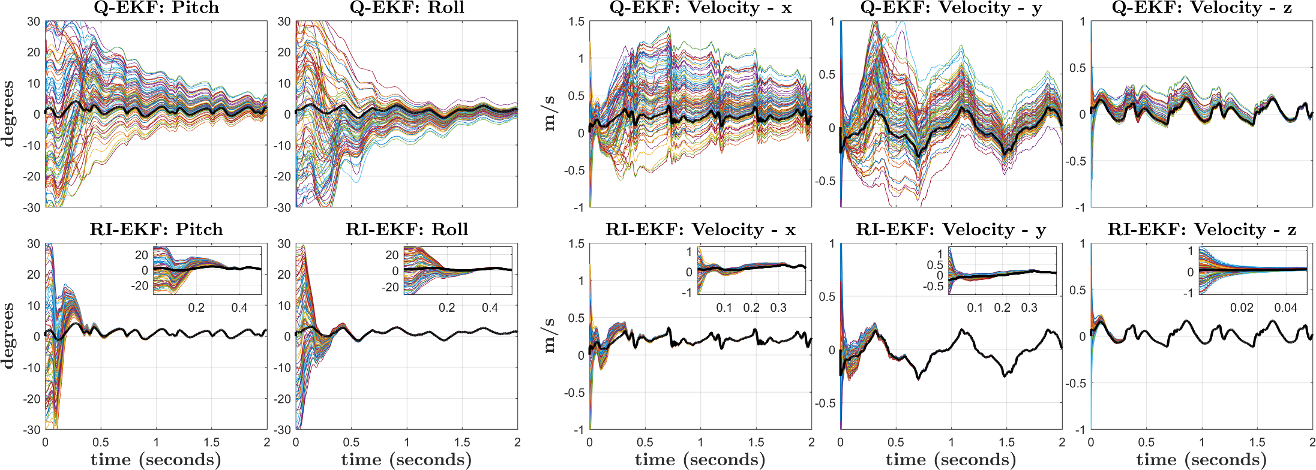}
      \caption{An experiment was performed where an actual Cassie-series robot slowly walked forward at approximately $0.3~\m/\sec$. The noisy measurements came from the on-board \ac{IMU} (VN-100) and the robot's joint encoders. The \acf{QEKF} and the proposed \acf{RIEKF} were run (off-line) 100 times using the same measurements, noise statistics, and initial covariance, but with random initial orientations and velocities. The black line represents the filter state estimates when initialized with a good estimate. The \ac{RIEKF} (bottom row) converges considerably faster than the \ac{QEKF} (top row) for all observable states. Zoomed-in plots of the \ac{RIEKF} performance is provided in the top-right corner.}
  \label{fig:exp_comparison_good_bias}
  \squeezeup\squeezeup
\end{figure*}

An experiment was performed where the robot walked forwards at approximately $0.3~\m/\sec$. The \ac{QEKF} and the proposed \ac{RIEKF} were run (off-line) 100 times using the same logged measurements, noise statistics, and initial covariance with random initial orientations and velocities. The noise statistics and initial covariance estimates are provided in Table~\ref{tab:params}. As with the simulation comparison presented in Section~\ref{sec:sim}, the initial mean estimate for the Euler angles were uniformly sampled from $-30\deg$ to $30\deg$ and the initial mean estimate for velocities were sampled uniformly from $-1.0~\m/\sec$ to $1.0~\m/\sec$. Bias estimation was turned on and the initial bias estimate was obtained from processing the IMU data when the robot was static. The pitch and roll estimates as well as the (body frame) velocity estimates for both filters are shown in Figure~\ref{fig:exp_comparison_good_bias}. The ``ground truth'' trajectory estimates (black lines) were computed by initializing each filter with a good state estimate at a time before the beginning of the plot (to allow for convergence). The initial orientation was obtained from the VectorNav-100's onboard EKF and the initial velocity was obtained through kinematics alone.

The experimental results for comparing filter convergence matches those of the simulation. The proposed \ac{RIEKF} converges faster and more reliably in all 100 runs than the \ac{QEKF}; therefore, due to the convenience of initialization and reliability for tracking the developed \ac{RIEKF} is the preferred observer.

When the state estimate is initialized close to the true value, the \ac{RIEKF} and \ac{QEKF} have similar performance (black lines), because the linearization of the error dynamics accurately reflects the underlying nonlinear dynamics. However, when the state estimate is far from the true value, the simulation and experimental results show that \ac{RIEKF} consistently converges faster than the \ac{QEKF}. The relatively poor performance of the \ac{QEKF}  is due to the error dynamics being linearized around the wrong operating point, in which case the linear system does not accurately reflect the nonlinear dynamics. In addition, when bias estimation is turned off, the invariant error dynamics of the \ac{RIEKF} do not depend on the current state estimate. As a result, the linear error dynamics can be accurately used even when the current state estimate is far from its true value, leading to better performance over the \ac{QEKF}. Although this theoretical advantage is lost when bias estimation is turned on, the experimental results (shown in Figure \ref{fig:exp_comparison_good_bias}) indicate that the \ac{RIEKF} still is the preferred observer due to less sensitivity to initialization.

%%%%%%%%%%%%%%%%%%%%%%%%%%%%%%%%%%%%%%%%%%%%%%%%%%%%
\subsection{Motion Capture Experiment}
\begin{figure}[h!]
    \centering
      \includegraphics[trim={0.5cm 0cm 0cm 0cm},clip,width=0.6\columnwidth]{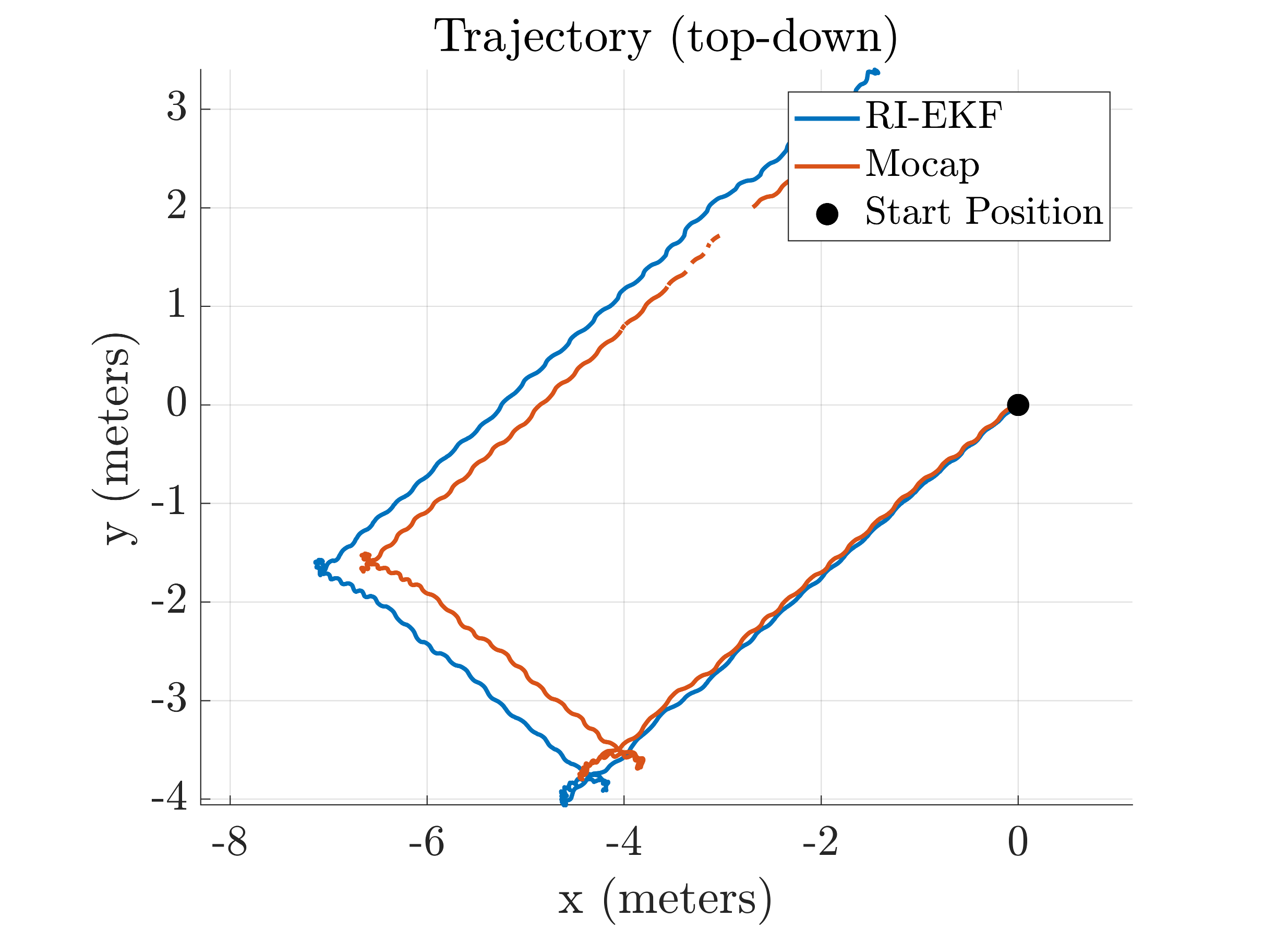}
      \caption{Top-down view of the \ac{InEKF}'s estimated trajectory for a motion capture experiment. The position drift is unobservable, however, the final drift is less than $5\%$ of the distance traveled.}
  \label{fig:mocap_trajectory}
\end{figure}

In order to verify the accuracy of the \ac{InEKF} state estimate, we performed a motion capture experiment in the University of Michigan's M-Air facility. This outdoor space is equipped with 18 Qualisys cameras that allows for position tracking. We had the Cassie robot walk untethered for $60~\sec$ along an approximately 15~$\m$ path. A top-down view of the estimated trajectory is shown in Figure~\ref{fig:mocap_trajectory}.
Although there is noticeable drift due to the unobservability of the position and yaw, the final position error accounts for less than $5\%$ of the distance traveled. This drift error is due to a combination of sensor noise and imperfect modeling of the robot's kinematics which may introduce biases to the forward kinematic measurements. The orientation, velocity, and position estimates along with their $3\sigma$ covariance hulls are shown in Figure~\ref{fig:mocap}. Due to the unobservability of the yaw angle, the velocity estimate is given in the body frame instead of the world frame. The orientation is plotted using exponential coordinates, $\exp(\boldsymbol{\phi})=\R$. Due to an inaccurate orientation estimate from the motion capture system, the ``ground truth'' for the orientation is given by the VectorNav-100, which runs a state-of-the-art \ac{QEKF} that fuses angular velocity, linear acceleration, and magnetometer measurements to estimate orientation only. 
%\newpage
\begin{figure*} \label{fig:mocap}
    \centering
    \begin{subfigure}[b]{0.99\textwidth}
        \includegraphics[trim={2cm 0cm 2cm 0cm},clip,width=\textwidth]{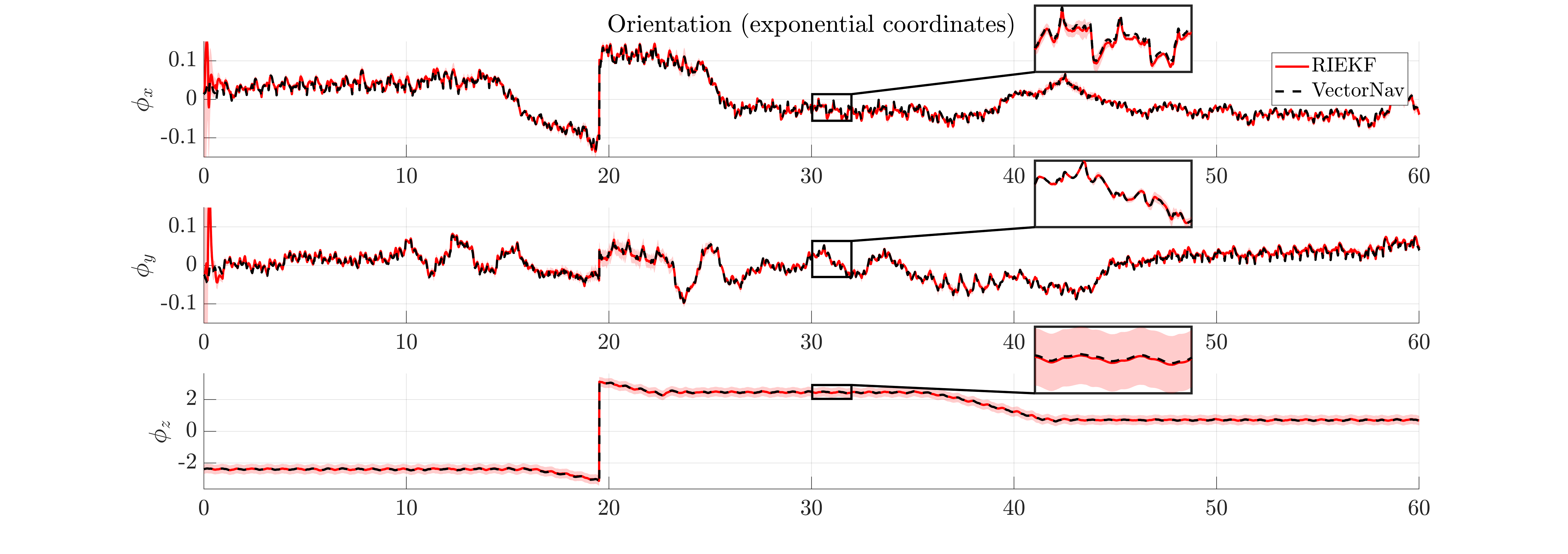}
    \end{subfigure}

    \begin{subfigure}[b]{0.99\textwidth}
        \includegraphics[trim={1cm 0cm 1cm 0cm},clip,width=\textwidth]{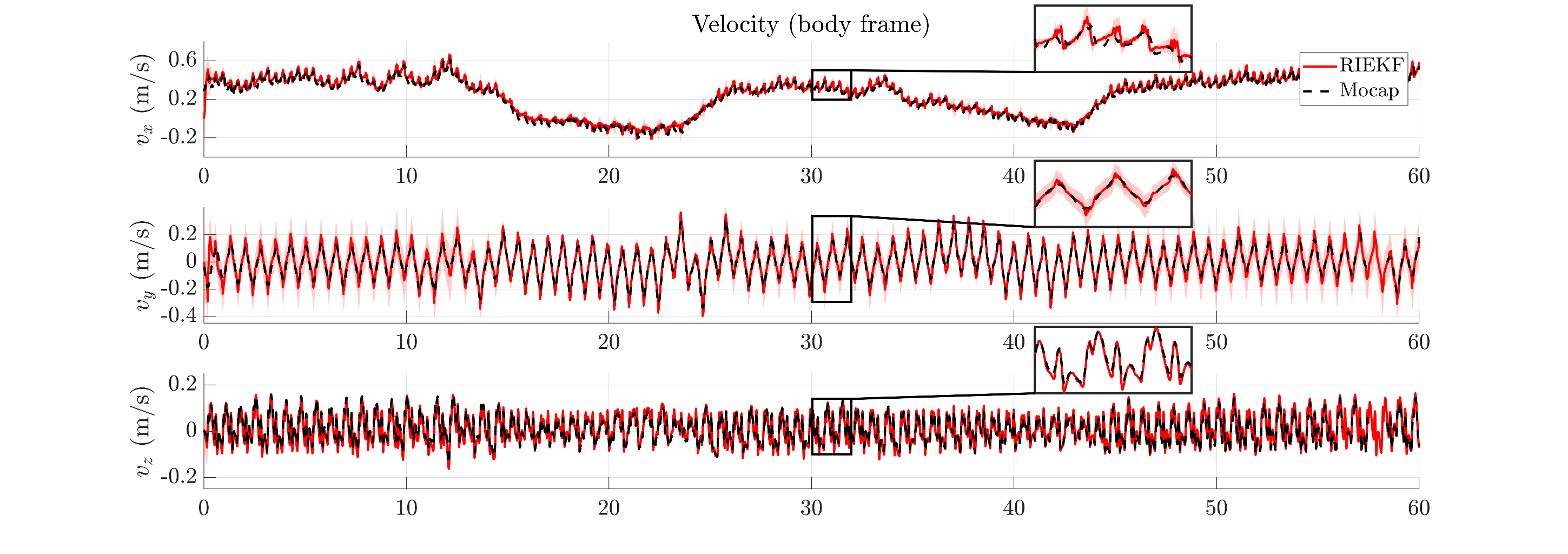}
    \end{subfigure}

    \begin{subfigure}[b]{0.99\textwidth} 
        \includegraphics[trim={4cm 0cm 4cm 0cm},clip,width=\textwidth]{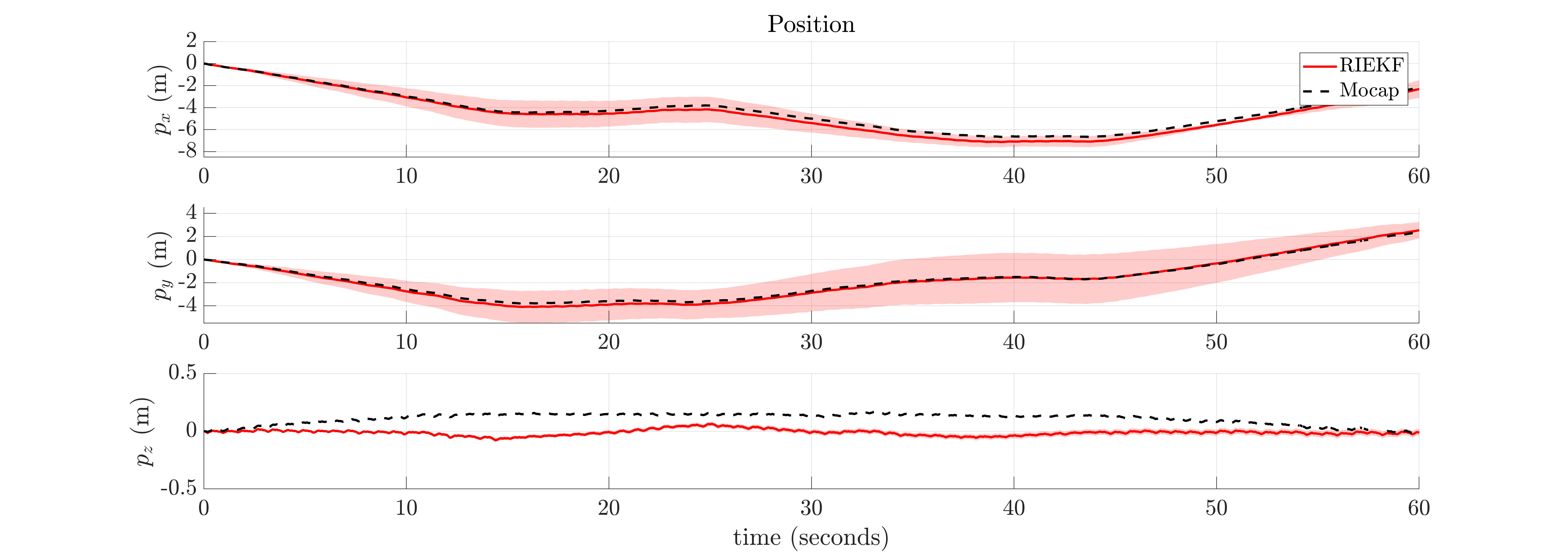}
    \end{subfigure}   
    \caption{Motion capture experiment conducted in the University of Michigan's M-Air facility. The dashed black line represents ground truth, the solid red line is the right-invariant EKF estimate, and the red shaded area represents the $3\sigma$ covariance hull. The ground truth for position and velocity were obtained using 18 Qualisys cameras. Due to poor orientation estimates from the motion capture system, the ``ground truth'' for orientation was obtained from the VectorNav-100, which runs a highly accurate on-board \ac{QEKF}. The orientation data is plotted using the exponential coordinates.}\label{fig:mocap}  
\end{figure*} 
As expected, the error for all observable states remains small. The absolute position and the orientation about the gravity vector are unobservable, so some drift will occur for these states. While the drift is largely imperceptible for $\boldsymbol{\phi}_z$, it is interesting to note that the covariance slowly grows over time due to this unobservability. More significant drift occurs on the absolute position states. This drift can be attributed to a combination of sensor noise, foot slip, and kinematic modeling errors.

In order to plot the $3\sigma$ covariance hull, the right-invariant error covariance needed to be converted to a covariance where the error is defined by Euclidean distance. Up to a first-order approximation, this mapping is done using:
\begin{equation} \label{eq:euclidean_error}
\begin{bmatrix}
    \delta \boldsymbol{\phi}_t \\
    \delta \v[t] \\
    \delta \p[t] 
\end{bmatrix} = 
\begin{bmatrix}
    -\Gam[1][-1](\bar{\boldsymbol{\phi}}_t) & \zeros & \zeros \\
    \vectorToSkew[\vE[t]] & -\I & \zeros \\
    \vectorToSkew[\pE[t]] & \zeros & -\I \\
\end{bmatrix} 
\begin{bmatrix}
    \tangentError[t][R] \\
    \tangentError[t][v] \\
    \tangentError[t][p] \\
\end{bmatrix},
\end{equation}
where the ``Euclidean orientation error'' is defined as $\delta \boldsymbol{\phi}_t \triangleq \boldsymbol{\phi}_t - \bar{\boldsymbol{\phi}}_t$, and the velocity and position errors match the \ac{QEKF} error states \eqref{eq:quaternion_error_states}. The matrix $\Gam[1](\bar{\boldsymbol{\phi}}_t)$ is known as the left Jacobian of $\SO(3)$ and has an analytical form. Further explanation and the derivation of the above equation is given in Appendix~\ref{appx:error_conversions}.

%%%%%%%%%%%%%%%%%%%%%%%%%%%%%%%%%%%%%%%%%%%%%%%%%%%%
\subsection{Long Odometry Experiment} 
\begin{figure}[h!]
    \centering
    \includegraphics[trim={5cm 3cm 3cm 2cm},clip,width=0.6\columnwidth]{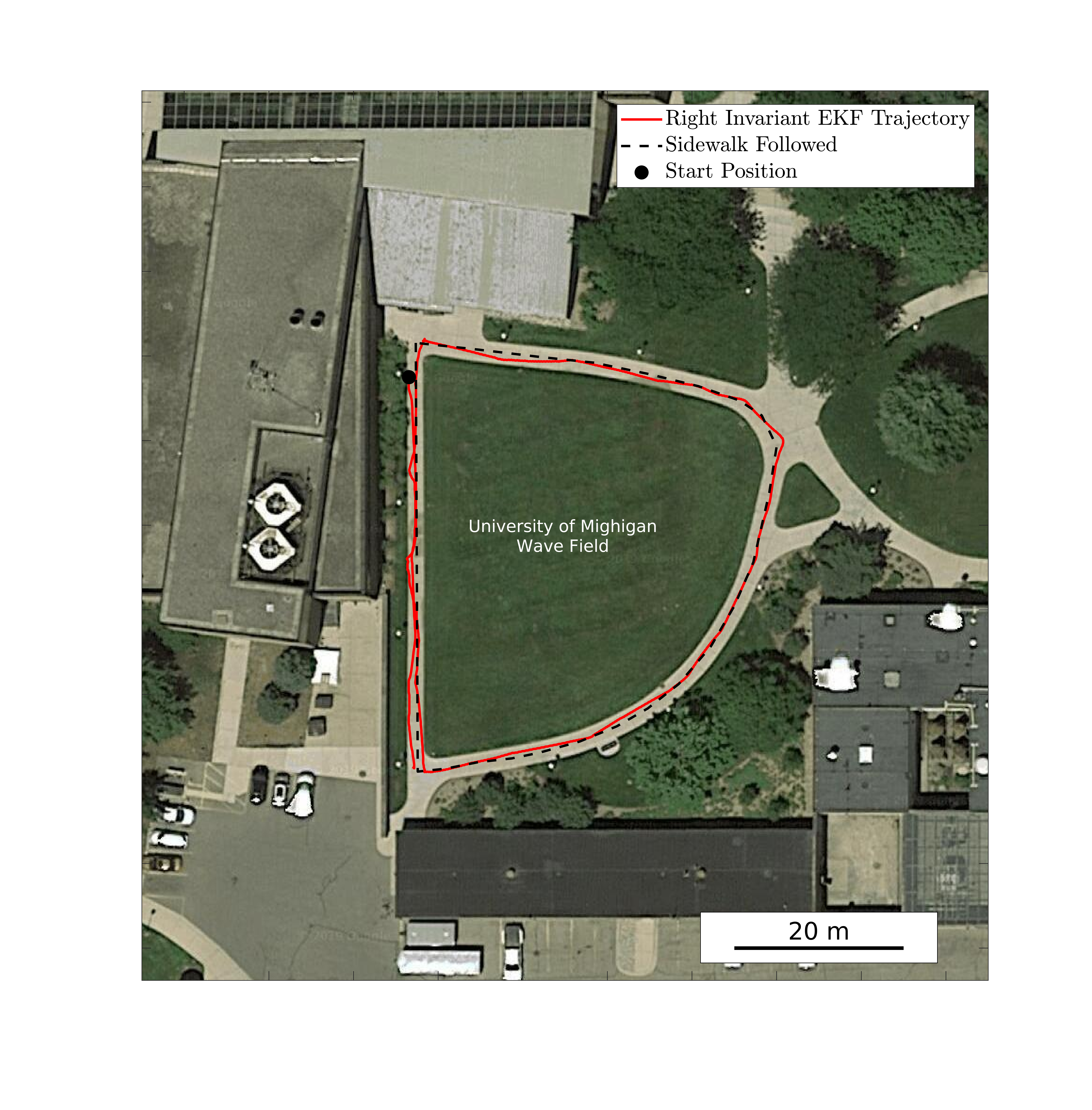}
    \caption{Long outdoor odometry experiment where Cassie walked roughly $200~\m$ along a sidewalk over 7 minutes and 45 seconds.}
  \label{fig:wave_field_odometry}
\end{figure}

In addition to providing accurate estimates of states vital for legged robot control (orientation and velocity), this \ac{InEKF} can also provide reliable odometry for a higher-level mapping or \ac{SLAM} system. To demonstrate the accuracy of long-term odometry, we had Cassie walk about $200~\m$ along a sidewalk around the University of Michigan's Wave Field. In total, the walk took 7 minutes and 45 seconds. The estimated path from the \ac{InEKF} overlaid onto Google Earth imagery is shown in Figure~\ref{fig:wave_field_odometry}. A video of this experiment can be found at \url{https://youtu.be/jRUltB_dMlo}. 

Even though the absolute position is unobservable, the odometry estimate from the \ac{InEKF} contains low enough drift to keep the estimate on the sidewalk for the duration of the experiment. The final position estimate is within a few meters of the true position, and the yaw drift is imperceptible. This odometry estimate is readily available, as it only depends on inertial, contact, and kinematic data, which barring sensor failure, always exist. It does not require the use of any vision systems that may be susceptible to changes in environment or lighting conditions.
% \FloatBarrier
% \afterpage{\FloatBarrier}

%%%%%%%%%%%%%%%%%%%%%%%%%%%%%%%%%%%%%%%%%%%%%%%%%%%%
\subsection{LiDAR Mapping Application}
\begin{figure*}[t!] 
    \centering  
    \begin{subfigure}[b]{0.49\textwidth}
        \includegraphics[trim={0cm 10mm 0cm 0cm},clip, width=\textwidth]{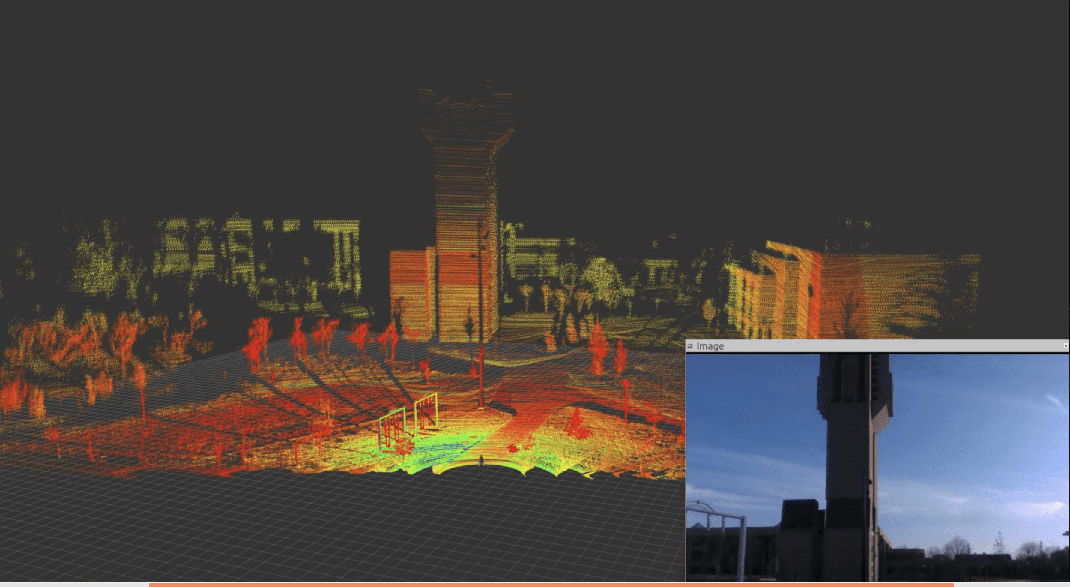}
        \caption{University of Michigan's North Campus with the bell tower}
    \end{subfigure}
    ~ 
    \begin{subfigure}[b]{0.49\textwidth}
        \includegraphics[trim={0cm 0mm 0cm 15mm},clip, width=\textwidth]{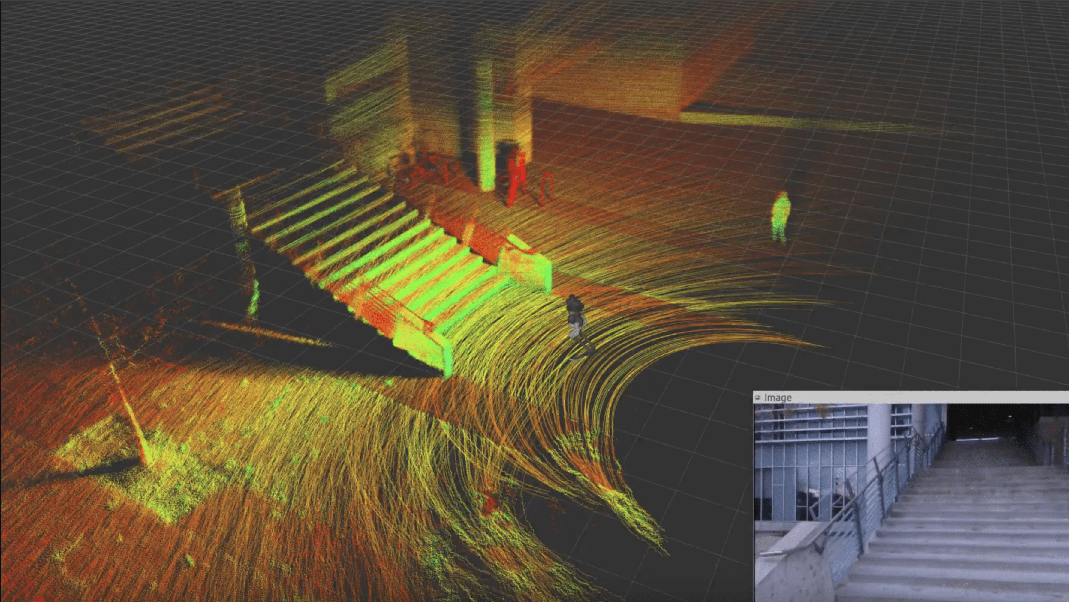}
        \caption{Looking towards a staircase} 
    \end{subfigure} 
    \\
    \begin{subfigure}[b]{0.49\textwidth}
        \includegraphics[trim={0cm 1mm 0cm 0cm},clip, width=\textwidth]{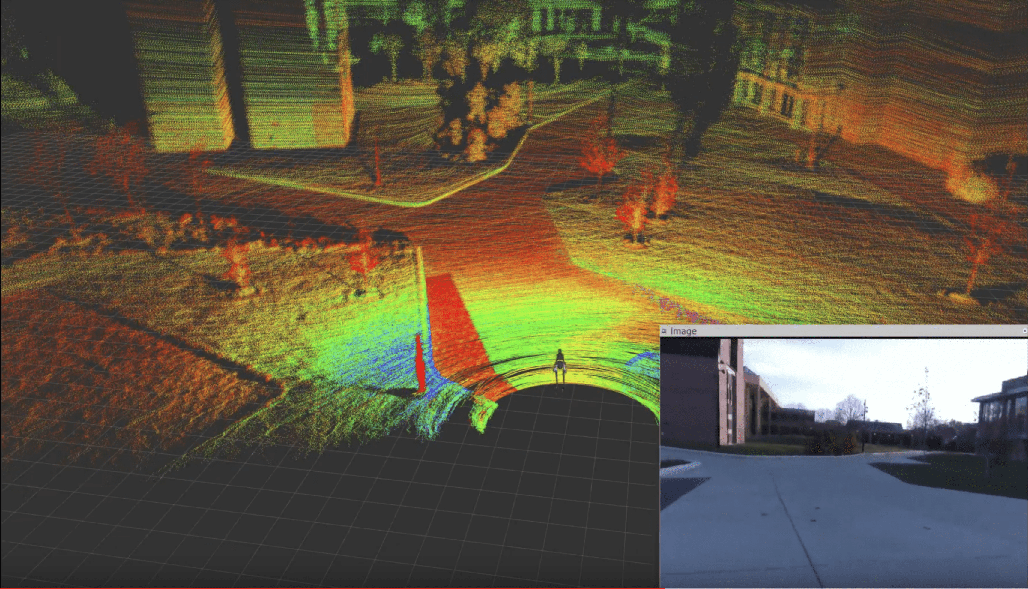}
        \caption{Walking along a sidewalk}
    \end{subfigure}
    ~ 
    \begin{subfigure}[b]{0.49\textwidth}
        \includegraphics[trim={0cm 1mm 0cm 0cm},clip, width=\textwidth]{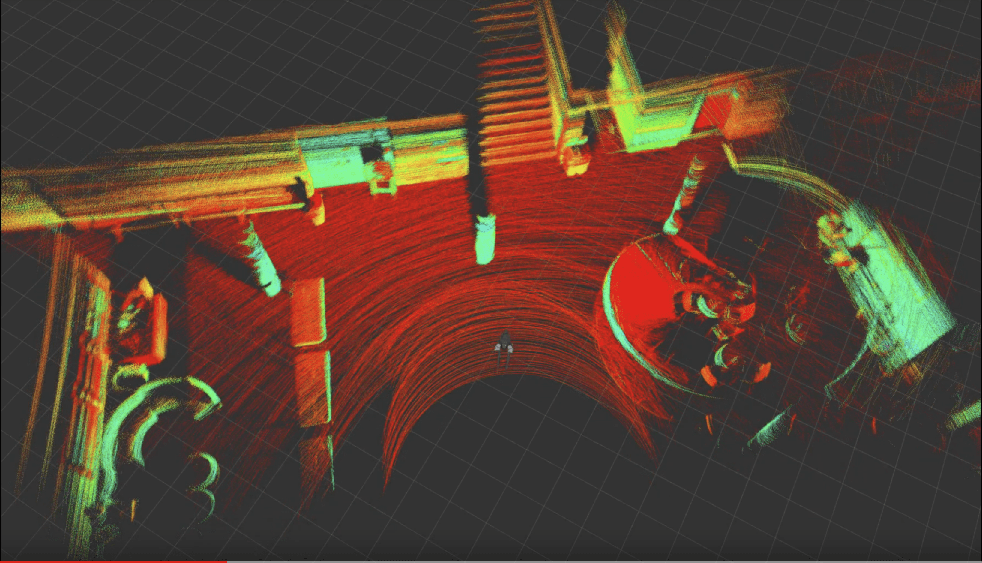}
        \caption{Inside the Bob and Betty Beyster Building}
    \end{subfigure}
    \caption{LiDAR maps created by transforming 10 seconds of point cloud data onto the pose trajectory estimated by the InEKF. The high frequency odometry estimate allows for motion compensation within a single scan of the LiDAR (10Hz); \url{https://youtu.be/pNyXsZ5zVZk} and \url{https://youtu.be/nbQTQw0gJ-k}.}\label{fig:lidar}
\end{figure*}

One application for the \ac{InEKF} odometry is the building of local maps of the environment. We equipped the Cassie-series robot with a new torso that houses a Velodyne VLP-32C LiDAR. With the filter running, we can project each received packet of point cloud data into the world frame based on the current state estimate. This point cloud data can then be accumulated to create a map of the environment. Figure~\ref{fig:lidar} shows a few still frames from several LiDAR mapping experiments. Videos of these results can be viewed at \url{https://youtu.be/pNyXsZ5zVZk} and
\url{https://youtu.be/nbQTQw0gJ-k}.

%% file: liekf.tex
\section{Alternative Left-Invariant Formulation} \label{sec:liekf}
For the derivations in Sections \ref{sec:riekf}-\ref{sec:switchcontact}, we were assuming the use of the right-invariant error. This choice was due to the forward kinematic measurement having the right-invariant observation form. However, it is possible to derive a left-invariant form of this filter, which may be more appropriate to use when dealing with left-invariant observations. For example, GPS measurements are left-invariant observations for the world-centric observer; see Section~\ref{sec:additionalmeasurements}. Written explicitly, the left-invariant error is
\begin{equation*}
    \begin{split}
    \groupError[t][l] &\triangleq \X[t][-1] \XE[t] =
    \begin{bmatrix}
    \R[t][\transpose]\RE[t]
    & \R[t][\transpose](\vE[t]-\v[t])
    & \R[t][\transpose](\pE[t]-\p[t]) 
    & \R[t][\transpose](\dE[t]-\d[t]) \\
    \zeros[1,3] & 1 & 0 & 0 \\
    \zeros[1,3] & 0 & 1 & 0 \\
    \zeros[1,3] & 0 & 0 & 1 \\
    \end{bmatrix}.
    \end{split}
    \end{equation*}
After carrying out the chain rule and making the first order approximation, \mbox{\small$\groupError[t][l] = \exp(\tangentError[t]) \approx \textbf{\textit{I}}_d + \vectorToAlgebra[\tangentError[t]]$}, the individual terms of the left-invariant error dynamics become:
\begin{equation*}
\begin{split}
    \deriv \R[t][\transpose]\RE[t] &\approx\vectorToSkew[ -\vectorToSkew[\wM[t]-\paramError[t][g]] \tangentError[t][R] - \paramError[t][g] + \noise[t][g] ] \\
    \deriv \R[t][\transpose](\vE[t]-\v[t]) &\approx -\vectorToSkew[\aM[t]-\accelerometerBiasE[t]]\tangentError[t][R] - \vectorToSkew[\wM[t]-\gyroscopeBiasE[t]]\tangentError[t][v] - \paramError[t][a] + \noise[t][a] \\
    \deriv \R[t][\transpose](\pE[t]-\p[t]) &\approx \tangentError[t][v] -  \vectorToSkew[\wM[t]-\gyroscopeBiasE[t]] \tangentError[t][p] \\
    \deriv \R[t][\transpose](\dE[t]-\d[t]) &\approx -\vectorToSkew[\wM[t]-\gyroscopeBiasE[t]] \tangentError[t][d] + \FK[R](\encodersM[t])\noise[t][v].\\
\end{split}
\end{equation*}
Using these results, the log-linear left-invariant dynamics can be expressed using the following linear system
\begin{equation*}
\begin{split}
    \deriv \tangentError[t] &= \A[t] \tangentError[t] + \noise[t] \\
    \implies \deriv \P[t] &= \A[t] \P[t] + \P[t] \A[t][\transpose] + \bar{\Q}_t,
\end{split}
\end{equation*}
where the dynamics and noise matrices are
\begin{equation}
\begin{split}
\A[t][l] &= 
\begin{bmatrix}
-\vectorToSkew[\wB[t]] & \zeros & \zeros & \zeros & -\I & \zeros \\
-\vectorToSkew[\aB[t]] & -\vectorToSkew[\wB[t]] & \zeros & \zeros & \zeros & -\I \\
\zeros & \I & -\vectorToSkew[\wB[t]] & \zeros & \zeros & \zeros\\
\zeros & \zeros & \zeros & -\vectorToSkew[\wB[t]] & \zeros & \zeros\\
\zeros & \zeros & \zeros & \zeros & \zeros & \zeros \\
\zeros & \zeros & \zeros & \zeros & \zeros & \zeros \\
\end{bmatrix} \\ 
\bar{\Q}_t &= 
\text{Cov}(\noise[t]).
\end{split}
\end{equation}
Similar to the right-invariant case, the dynamics only depend on the state through the IMU bias. When a left-invariant observation comes in, the state estimate is corrected using
\begin{equation} \label{eq:left_invariant_update}
    \left(\XE[t][+], \params[t][+]\right) = \left( \XE[t] \exp\left( \K[t][\tangentError] \SelectionMatrix \XE[t][-1] \Y[t] \right)\, , \;\; \paramsE[t] + \K[t][\paramError] \SelectionMatrix \XE[t][-1] \Y[t] \right),
\end{equation}
where the exponential map is now multiplied on the right side~\citep{barrau2017invariant}.

\subsection{Switching Between Left and Right-Invariant Errors}
Because forward kinematic measurements have the right invariant observation form, the innovation equations are only autonomous when the right invariant error is used. Fortunately, it is possible to switch between the left and right error forms through the use of the adjoint map.
\begin{equation*}
\begin{split}
    \groupError[t][r] &= \XE[t]\X[t][-1] = \XE[t]\groupError[t][l]\XE[t][-1] \\
    \implies \exp(\tangentError[t][r]) &= \XE[t]\exp(\tangentError[t][l])\XE[t][-1] = \exp(\Adjoint[\XE[t]] \tangentError[t][l]) \\
    \implies \tangentError[t][r] &= \Adjoint[\XE[t]] \tangentError[t][l] \\
\end{split}
\end{equation*}
This transformation is exact, which means that we can easily switch between the covariance of the left and right invariant errors using
\begin{equation} \label{eq:covariance_switch}
    \P[t][r] = \Adjoint[\XE[t]] \, \P[t][l] \, \Adjoint[\XE[t]]^\transpose .
\end{equation}

Therefore, when handling a right-invariant observation, we can map the propagated left-invariant covariance to the right-invariant covariance temporarily, apply the right-invariant update equations \eqref{eq:right_invariant_update}, then map the corrected covariance back to the left-invariant form.

% \mgj{make it a proposition} 
It is also possible to compute the log-linear left-invariant dynamics starting from the right-invariant form. Substituting \eqref{eq:covariance_switch} into the (right-invariant) covariance propagation equation \eqref{eq:propagation} and solving for the left-invariant covariance yields
\begin{equation*}
\begin{split}
    %\small
    \deriv \P[t][l] = &\left( \Adjoint[\XE[t][-1]] \A[t][r] \Adjoint[\XE[t]] - \Adjoint[\XE[t][-1]]\deriv\left(\Adjoint[\XE[t]]\right) \right) \P[t][l] \\
    &+ \P[t][l] \left(\Adjoint[\XE[t]]^\transpose  \A[t][r\transpose] \Adjoint[\XE[t][-1]]^\transpose - \deriv\left(\Adjoint[\XE[t]]^\transpose\right)\Adjoint[\XE[t][-1]]^\transpose \right) \\
    &+ \Adjoint[\XE[t][-1]] \bar{\Q}^r_t \Adjoint[\XE[t][-1]]^\transpose .
\end{split}
\end{equation*}
Therefore, we learn that the right and left dynamics and noise matrices are related by the following expressions
\begin{equation}
    \begin{split}
        \A[t][l] &\triangleq \Adjoint[\XE[t][-1]] \A[t][r] \Adjoint[\XE[t]] - \Adjoint[\XE[t][-1]]\deriv\left(\Adjoint[\XE[t]]\right) \\
        \bar{\Q}^l_t &\triangleq \Adjoint[\XE[t][-1]] \bar{\Q}^r_t \Adjoint[\XE[t][-1]]^\transpose .
    \end{split}
\end{equation}

\begin{remark}
    Intuitively, the left-invariant error represents an error measured in the body frame of the robot, while the right-invariant error represents an error measured in the world or spatial frame. The frame of measurement dictates whether the exponential map appears on the right or left in the update equations. The error can be moved between these two frames using the adjoint map of the Lie group.
\end{remark}

\subsection{Adding New Contact Points}
The process for removing a contact point from the state (marginalization) is identical to the right-invariant error case, described in Section \ref{sec:removecontact}. Likewise, when a new contact is detected, the state can be augmented using the same kinematics relation \eqref{eq:contact_augmentation} as before. However, due to the change in error variable, the process for augmenting the covariance will be different. 

In order to compute the new covariance, we need to look at the left-invariant error, 
\begin{equation*}
\begin{split}
  \groupError[t][d] &= \R[t][\transpose](\dE[t] - \d[t]) \\
   &= \R[t][\transpose]\left(\pE[t] + \RE[t] \FK[p](\encodersM[t])\right) - \R[t][\transpose]\left( \p[t] + \R[t] \FK[p](\encodersM[t] - \noise[t][\alpha]) \right)  \\
   &\approx \groupError[t][p] + \groupError[t][R]\FK[p](\encodersM[t]) - \FK[p](\encodersM[t]) + \J[p](\encodersM[t]) \noise[t][\alpha] \\
\implies \tangentError[t][d] &\approx \tangentError[t][p] - \vectorToSkew[\FK[p](\encodersM[t])]\tangentError[t][R] + \J[p](\encodersM[t]) \noise[t][\alpha].
\end{split}
\end{equation*}
Therefore, covariance augmentation can be done using the following linear map,
\begin{equation}
\begin{split}
\begin{bmatrix}
\tangentError[t][R] \\
\tangentError[t][v] \\
\tangentError[t][p] \\
\tangentError[t][d] \\
\end{bmatrix} &=
\begin{bmatrix}
\I & \zeros & \zeros \\
\zeros & \I & \zeros \\
\zeros & \zeros & \I \\
\vectorToSkew[-\FK[p](\encodersM[t])] & \zeros & \I \\
\end{bmatrix}
\begin{bmatrix}
\tangentError[t][R] \\
\tangentError[t][v] \\
\tangentError[t][p] \\
\end{bmatrix} + 
\begin{bmatrix}
\zeros \\
\zeros \\
\zeros \\
\J[p](\encodersM[t]) \\
\end{bmatrix} \noise[t][\alpha] \\
\tangentError[t][\mathrm{new}] &\triangleq \F[t] \, \tangentError[t] + \G[t] \noise[t][\alpha] \\
\implies \P[t][\mathrm{new}] &= \F[t] \, \P[t] \, \F[t][\transpose] + \G[t] \, \text{Cov}(\noise[t][\alpha]) \, \G[t][\transpose]. 
\end{split}
\end{equation}

%% file: robocentric.tex
\section{Robo-centric Estimator} \label{sec:robocentric}
In this section, we derive a ``robot-centric'' version of the contact-aided \ac{InEKF} where the estimated state is measured in the robot's base (\ac{IMU}) frame. When switching to a robot-centric model, the forward kinematics measurements take the left-invariant observation form. In addition, the left/right-invariant error dynamics equations are identical to the world-centric form, albeit swapped. The right-invariant error dynamics for the world-centric estimator are equivalent to the left-invariant error dynamics for the robo-centric estimator.

The properties of this filter are identical to the \ac{RIEKF} derived in Section \ref{sec:riekf}. However, in some cases this filter may be preferred as it directly estimates states that are useful for controlling a legged robot (namely the velocity measured in the body frame).

%%%%%%%%%%%%%%%%%%%%%%%%%%%%%%%%%%%%%%%%
\subsection{State and Dynamics}
We are interested in estimating the same states as before, though measured in the robot's body frame. Again, the state variables can form a matrix Lie group, $\lieGroup$. Specifically, for $N$ contact points, $\X[t] \in \SE_{N+2}(3)$ can be represented by the following matrix (which is simply the inverse of the world-centric state)\footnote{The negative sign on body velocity appears when inverting the world-centric state; $-\linearVelocity[B][B]=-\orientation[WB]^\transpose\,\linearVelocity[B][W]$. This sign is removed on the position vectors by swapping the start and end points, $\position[BW][B]=-\orientation[WB]^\transpose\,\position[WB][B]$.}:
\begin{equation*}
  \textbf{X}_t ~\redefine~ 
  \begin{bmatrix}
    \orientation[BW](t) & -\linearVelocity[B][B](t) & \position[BW][B](t) & {}_\text{B}\p[\text{C}_1\text{W}](t) & \cdots & {}_\text{B}\p[\text{C}_N\text{W}](t)  \\
    \zeros[1 \times 3] & 1 & 0 & 0 & \cdots & 0 \\
    \zeros[1 \times 3] & 0 & 1 & 0 & \cdots & 0 \\
    \zeros[1 \times 3] & 0 & 0 & 1 & \cdots & 0 \\
    \vdots & \vdots & \vdots & \vdots & \ddots & \vdots \\ 
    \zeros[1 \times 3] & 0 & 0 & 0 & \cdots & 1 \\
  \end{bmatrix}
\end{equation*} 
Without loss of generality, assume a single contact point. Furthermore, for the sake of readability, we redefine our shorthand notation to be body-centric states,
\begin{equation}
    \X[t] ~~\redefine~~
    \begin{bmatrix}
      \R[t] & \v[t] & \p[t] & \d[t] \\
      \zeros[1 \times 3] & 1 & 0 & 0 \\
      \zeros[1 \times 3] & 0 & 1 & 0 \\
      \zeros[1 \times 3] & 0 & 0 & 1 \\
    \end{bmatrix}.
\end{equation}

Using the bias corrected \ac{IMU} measurements, the individual terms of the new robot-centric system dynamics can be derived as~\citep{bloesch2013state}
\begin{equation} \label{eq:body_dynamics_with_bias}
    \begin{split}
        \deriv \R[t] &= -\vectorToSkew[\wB[t] - \noise[t][g]] \R[t] \\
        \deriv \v[t] &= -(\aB[t] - \noise[t][a]) - \R[t]\,\g - \vectorToSkew[\wB[t] - \noise[t][g]]\v[t] \\
        \deriv \p[t] &= \v[t] - \vectorToSkew[\wB[t] - \noise[t][g]] \p[t] \\
        \deriv \d[t] &= -\vectorToSkew[\wB[t] - \noise[t][g]]\d[t] - \FK[R](\encodersM[t])\noise[t][v] .
    \end{split}
\end{equation}
Written in matrix form, this becomes
\begin{equation}
  \resizebox{\hsize}{!}{$
    \begin{split}
        \deriv \X[t] &= 
        \begin{bmatrix}
            -\vectorToSkew[\wB[t]] \R[t]
            & -\aB[t] - \R[t]\g - \vectorToSkew[\wM[t]] \v[t] 
            & \v[t] - \vectorToSkew[\wM[t]] \p[t] 
            & -\vectorToSkew[\wB[t]]\d[t] \\
            \zeros[1,3] & 0 & 0 & 0  \\
            \zeros[1,3] & 0 & 0 & 0  \\
            \zeros[1,3] & 0 & 0 & 0  \\
        \end{bmatrix} - 
        \begin{bmatrix}
            \vectorToSkew[\noise[t][g]] 
            & \noise[t][a]
            & \zeros[3,1]
            & \FK[R](\encodersM[t]) \noise[t][v]  \\
            \zeros[1,3] & 0 & 0 & 0  \\
            \zeros[1,3] & 0 & 0 & 0  \\
            \zeros[1,3] & 0 & 0 & 0  \\
        \end{bmatrix} 
        \begin{bmatrix}
            \R[t] & \v[t] & \p[t] & \d[t] \\
            \zeros[1,3] & 1 & 0 & 0 \\
            \zeros[1,3] & 0 & 1 & 0 \\
            \zeros[1,3] & 0 & 0 & 1 \\
        \end{bmatrix} \\
        &\triangleq f_u(\X[t], \params[t]) - \vectorToAlgebra[\noise[t]] \X[t]
    \end{split}
    $}
\end{equation}
with $\noise[t] \triangleq \vector[\noise[t][g],\, \noise[t][a],\, \zeros[3,1], \FK[R](\encodersM[t]) \noise[t][v]]$. The deterministic dynamics function $f_u(\cdot)$ can be shown to satisfy the group affine property \eqref{eq:group_affine}. Therefore, the left (and right) invariant error dynamics depends solely on the invariant error.

We can derive the log-linear dynamics matrices for the body-centric estimator following a similar derivation process to the world-centric version in Section \ref{sec:riekf}. In fact, without IMU bias, the linearization of the left-invariant dynamics for the body-centric estimator is the same as the right-invariant dynamics for the world-centric estimator; 
\begin{equation}
\begin{split}
\A[t][l] \; \textit{(body-centric)} &= \A[t][r] \; \textit{(world-centric)} \\
\A[t][r] \; \textit{(body-centric)} &= \A[t][l] \; \textit{(world-centric)}. \\
\end{split}
\end{equation}
When IMU bias is included, the above relation still holds, though with the bias terms negated\footnote{If definition of bias error is negated, even these terms would remain the same.}. The noise covariance matrices for the body-centric left/right-invariant propagation models are given by:
\begin{equation}
\begin{split}
\bar{\Q}^l_t &= 
\begin{bmatrix}
\Adjoint[\XE[t][-1]] & \zeros[15,6] \\
\zeros[6,15] & \I[6]
\end{bmatrix}
\text{Cov}(\noise[t])
\begin{bmatrix}
\Adjoint[\XE[t][-1]] & \zeros[15,6] \\
\zeros[6,15] & \I[6]
\end{bmatrix}^\transpose \\
\bar{\Q}^r_t &= \text{Cov}(\noise[t]),                                                            
\end{split}
\end{equation}
which are also swapped versions of the world-centric noise matrices, after accounting for the redefinition of $\XE[t]$ as its inverse. A comparison of the world-centric and body-centric equations are given in Tables 2 and 3.%\ref{tab:world_centric_estimator} and \ref{tab:robo_centric_estimator}.

%%%%%%%%%%%%%%%%%%%%%%%%%%%%%%%%%%%%%%%% 
\subsection{Left-Invariant Forward Kinematic Measurement Model}
We use forward-kinematics to measure the relative position of the contact point with respect to the body, $\FK[p](\encodersM[t])$. Using the new robot-centric state variables, the measurement model \eqref{eq:forward_kinematics_position_measurement} becomes 
\begin{equation}
\FK[p](\encodersM[t]) = \p[t] - \d[t] + \J[p](\encodersM[t]) \noise[t].
\end{equation}
Re-written in matrix form, this measurement will have now have the left-invariant observation form \eqref{eq:invariant_observations},
\begin{equation}
\underbrace{
\begin{bmatrix}
  \FK[p](\encodersM[t]) \\ 0 \\ 1 \\ -1
\end{bmatrix} 
}_{\textstyle \Y[t]}
= 
\underbrace{
\begin{bmatrix}
  \R[t] & \v[t] & \p[t] & \d[t] \\
  \zeros[1,3] & 1 & 0 & 0 \\
  \zeros[1,3] & 0 & 1 & 0 \\
  \zeros[1,3] & 0 & 0 & 1 \\
\end{bmatrix}
}_{\textstyle \X[t]}
\underbrace{
\begin{bmatrix}
  0 \\ 0 \\ 1 \\ -1
\end{bmatrix} 
}_{\textstyle \b}
+
\underbrace{
\begin{bmatrix}
  \J[p](\encodersM[t]) \noise[t][\alpha] \\ 0 \\ 0 \\ 0 \\
\end{bmatrix}
}_{\textstyle \V[t]}.
\end{equation}    
The state update equation will take the left-invariant form \eqref{eq:left_invariant_update}, where the matrices $\H[t]$ and $\bar {\N}_t$ can be derived to be:
\begin{equation} \label{eq:left_invariant_kinematics_linearization}
  \begin{split}
    \H[t] &= 
    \begin{bmatrix}
      \zeros[3,3] & \zeros[3,3] & \I & -\I 
    \end{bmatrix}, \\
    \bar{\N}_t &= \bar{\R}_t^\transpose \; \J[p](\encodersM[t]) \; \text{Cov}(\noise[t][\alpha]) \; \J[p][\transpose](\encodersM[t]) \; \bar{\R}_t.
  \end{split}
\end{equation}
It is important to note that the forward kinematics measurement is a right-invariant observation for the world-centric estimator, while the same measurement becomes a left-invariant observation for the body-centric version. Also, the above linearization \eqref{eq:left_invariant_kinematics_linearization} is similar to the world-centric, right-invariant one \eqref{eq:right_invariant_kinematics_linearization}. Namely, $\H[t]$ is simply negated, while $\bar{\N}_t$ is identical after accounting for the redefinition of the state as its inverse.

%% file: additional_measurements.tex
\section{Additional Sensor Observations} \label{sec:additionalmeasurements}
This article extends upon our conference paper results~\citep{hartley2018contact} where we originally presented the contact-aided \ac{InEKF}. As described in earlier sections, this filter uses an inertial-contact dynamics model with corrections coming from forward kinematics. In addition to forward kinematics, other groups have discovered that a number of measurements common to robotics can also fit the invariant observation model~\eqref{eq:invariant_observations}. \citet{bonnabel2007left} developed an invariant observer that uses magnetometer and acceleration measurements to solve the attitude estimation problem. \citet{barczyk2011invariant,barrau2015non} described methods for invariant observer design for GPS and magnetometer-aided navigation. \citet{wu2017invariant} developed an \ac{InEKF} to solve visual-inertial navigation. It has also been shown that an \ac{InEKF} can be used for \ac{SLAM} \citep{bonnabel2012symmetries,barrau2015ekf,zhang2017convergence}.

In particular, it is interesting to note the similarities between our contact-aided \ac{InEKF} and landmark-based \ac{SLAM}. In the simplest case, this \ac{SLAM} problem involves jointly estimating the robots state along with the position of static landmarks in the environment. The robot is often assumed to have a sensor capable of measuring the position of the landmark relative to the robot. This formulation is identical to our developed \ac{InEKF} with the contact positions acting as landmarks and forward kinematics measuring the relative translation between the base and contact frames. This similarity was also mentioned by \citet{bloesch2012state}. However, there are a few notable differences. The contact frame velocity is assumed to be white noise to allow for foot slip, while landmarks are usually treated as static. Forward kinematics measurements often come at high frequencies ($2000~\Hz$ on Cassie). In contrast, landmarks measurements are often at a much lower frequency. Finally, with landmark observations, a data association problem often has to be solved which associates the measurement with a particular landmark state. This problem does not exist with forward kinematic measurements. 

%\mgj{Also, once the contact is broken it's not possible to reobserve it while landmarks can be reobserved (known as loop-closures)?}
%\rha{Its not that we can't reobserve the contact, it is that we no longer have a propagation model for it, necessitating marginalization. I don't think this is necessary to explain here.}

Due to the similarities between contacts and landmarks, in our state matrix, the contact states could be easily replaced with landmark states, ${}_\text{W}\p[\text{WL}_i](t)$,
\begin{equation*}
    \nonumber \textbf{X}_t \triangleq
    \begin{bmatrix}
    \orientation[WB](t) & \linearVelocity[B][W](t) & \position[WB][W](t) & {}_\text{W}\p[\text{WL}_1](t) & \cdots & {}_\text{W}\p[\text{WL}_N](t)  \\
    \zeros[1,3] & 1 & 0 & 0 & \cdots & 0 \\
    \zeros[1,3] & 0 & 1 & 0 & \cdots & 0 \\
    \zeros[1,3] & 0 & 0 & 1 & \cdots & 0 \\
    \vdots & \vdots & \vdots & \vdots & \ddots & \vdots \\ 
    \zeros[1,3] & 0 & 0 & 0 & \cdots & 1 \\
    \end{bmatrix},
\end{equation*}
or a combination of contact and landmark positions. In this way, it is possible to develop an observer that contains no unobservable states. Of course, like \ac{EKF}-\ac{SLAM}, the filter can become too computationally expensive to run in real-time if the number of landmarks grows too large. 

Tables 2 and 3 give a summary of the left/right world-centric and robo-centric \ac{InEKF} equations assuming a single contact and landmark position. The linearized observation matrix and observation type for several different sensors are also provided. In these tables, $\l[t]$ is shorthand for the true landmark position, and $\textbf{m}$ denotes the true magnetic field vector. Using these tables, it is clear to see the relation between the left/right invariant error dynamics and the world/robo-centric formulations. All of these dynamics and observations are supported in an open-source C++ Library released alongside this article; available at: \href{https://github.com/RossHartley/invariant-ekf}{\footnotesize\url{https://github.com/RossHartley/invariant-ekf}}. 

\begin{table*}[t] 
    \label{tab:world_centric_estimator}
    \caption{Summary of World-centric State Estimator}
    \begin{tabularx}{1.0\textwidth}[t]{p{7cm}X}
        \toprule
        \footnotesize State Definition & \footnotesize Deterministic Nonlinear Dynamics \\  
        \midrule
        \scriptsize
        $\X[t] \triangleq
        \begin{bmatrix}
        \orientation[WB] & \linearVelocity[B][W] & \position[WB][W] & {}_\text{W}\p[\text{WC}] & {}_\text{W}\p[\text{WL}] \\
        \zeros[1\times3] & 1 & 0 & 0 & 0 \\
        \zeros[1\times3] & 0 & 1 & 0 & 0 \\
        \zeros[1\times3] & 0 & 0 & 1 & 0 \\
        \zeros[1\times3] & 0 & 0 & 0 & 1 \\
        \end{bmatrix}$ 
        & 
        \scriptsize
        $f_{\u[t]}(\XE[t],\paramsE[t]) = 
        \begin{bmatrix}
        \RE[t]\vectorToSkew[\wB[t]] 
        & \RE[t]\aB[t] + \g  
        & \vE[t]
        & \zeros[3\times1] 
        & \zeros[3\times1] \\
        \zeros[1\times3] & 0 & 0 & 0 & 0 \\
        \zeros[1\times3] & 0 & 0 & 0 & 0  \\
        \zeros[1\times3] & 0 & 0 & 0 & 0  \\
        \zeros[1\times3] & 0 & 0 & 0 & 0  \\
        \end{bmatrix}$
    \end{tabularx}
    \begin{tabularx}{1.0\textwidth}[t]{p{7cm}X}
        \toprule
        \footnotesize Log-Linear Right-Invariant Dynamics & \footnotesize Log-Linear Left-Invariant Dynamics \\
        \midrule
        \scriptsize
        $\A[t][r] = 
        \begin{bmatrix}
        \zeros & \zeros & \zeros & \zeros & \zeros & -\RE[t] & \zeros \\
        \vectorToSkew[\g] & \zeros & \zeros & \zeros & \zeros & -\vectorToSkew[\vE[t]]\RE[t] & -\RE[t] \\
        \zeros & \I & \zeros & \zeros & \zeros &  -\vectorToSkew[\pE[t]]\RE[t] & \zeros \\
        \zeros & \zeros & \zeros & \zeros & \zeros & -\vectorToSkew[\dE[t]] \RE[t] & \zeros \\
        \zeros & \zeros & \zeros & \zeros & \zeros & -\vectorToSkew[\lE[t]] \RE[t] & \zeros \\
        \zeros & \zeros & \zeros & \zeros & \zeros & \zeros & \zeros \\
        \zeros & \zeros & \zeros & \zeros & \zeros & \zeros & \zeros \\
        \end{bmatrix}$ 
        &
        \scriptsize
        \resizebox{\linewidth}{!}{
        $\A[t][l] = 
        \begin{bmatrix}
        -\vectorToSkew[\wB[t]] & \zeros & \zeros & \zeros & \zeros & -\I & \zeros \\
        -\vectorToSkew[\aB[t]] & -\vectorToSkew[\wB[t]] & \zeros & \zeros & \zeros & \zeros & -\I \\
        \zeros & \I & -\vectorToSkew[\wB[t]] & \zeros & \zeros & \zeros & \zeros\\
        \zeros & \zeros & \zeros & -\vectorToSkew[\wB[t]] & \zeros & \zeros & \zeros\\
        \zeros & \zeros & \zeros & \zeros & -\vectorToSkew[\wB[t]] & \zeros & \zeros \\
        \zeros & \zeros & \zeros & \zeros & \zeros & \zeros & \zeros \\
        \zeros & \zeros & \zeros & \zeros & \zeros & \zeros & \zeros \\
        \end{bmatrix}$} 
        \\
        \scriptsize
        $\hat{\Q}^r_t = 
        \begin{bmatrix}
        \Adjoint[\XE[t]] & \zeros[15,6] \\
        \zeros[6,15] & \I[6]
        \end{bmatrix}
        \text{Cov}(\noise[t])
        \begin{bmatrix}
        \Adjoint[\XE[t]] & \zeros[15,6] \\
        \zeros[6,15] & \I[6]
        \end{bmatrix}^\transpose$
        &
        \scriptsize
        $\hat{\Q}^l_t = 
        \text{Cov}(\noise[t])$ \\ 
    \end{tabularx}
    \begin{tabularx}{1.0\textwidth}[t]{XXX}
        \toprule
        \footnotesize Measurement & \footnotesize Observation Matrix, $\H$  & \footnotesize Observation Type \\
        \midrule
    \footnotesize Forward Kinematic & \scriptsize $\begin{bmatrix} \zeros & \zeros & -\I & \I & \zeros & \zeros & \zeros \end{bmatrix}$ & \footnotesize Right-Invariant \\ 
    \footnotesize Relative Landmark Position & \scriptsize $\begin{bmatrix} \zeros & \zeros & -\I & \zeros & \I & \zeros & \zeros \end{bmatrix}$ & \footnotesize Right-Invariant \\
    \footnotesize Absolute Landmark Position & \scriptsize $\begin{bmatrix} \vectorToSkew[\l] & \zeros & -\I & \zeros & \zeros & \zeros & \zeros \end{bmatrix}$ & \footnotesize Right-Invariant \\
    \footnotesize Magnetometer & \scriptsize $\begin{bmatrix} \vectorToSkew[\textbf{m}] & \zeros & \zeros & \zeros & \zeros & \zeros & \zeros \end{bmatrix}$ & \footnotesize Right-Invariant \\
    \footnotesize GPS Position & \scriptsize $\begin{bmatrix} \zeros & \zeros & \I & \zeros & \zeros & \zeros & \zeros \end{bmatrix}$ & \footnotesize Left-Invariant \\
        \bottomrule
    \end{tabularx}
    \vspace{0.5cm}
    \label{tab:robo_centric_estimator}
    \caption{Summary of Robo-centric State Estimator}
    \begin{tabularx}{1.0\textwidth}[t]{p{5.5cm}X}
        \toprule
        \footnotesize State Definition & \footnotesize Deterministic Nonlinear Dynamics \\
        \midrule
        \scriptsize
        \resizebox{\linewidth}{!}{
        $\X[t] \triangleq
        \begin{bmatrix}
        \orientation[BW] & -\linearVelocity[B][B] & \position[BW][B] & {}_\text{B}\p[\text{CW}] & {}_\text{B}\p[\text{LW}] \\
        \zeros[1\times3] & 1 & 0 & 0 & 0 \\
        \zeros[1\times3] & 0 & 1 & 0 & 0 \\
        \zeros[1\times3] & 0 & 0 & 1 & 0 \\
        \zeros[1\times3] & 0 & 0 & 0 & 1 \\
        \end{bmatrix}$}
        & 
        \scriptsize
        \resizebox{\linewidth}{!}{
        $f_{\u[t]}(\XE[t],\paramsE[t]) = 
        \begin{bmatrix}
        -\vectorToSkew[\wB[t]] \R[t]
        & -\aB[t] - \R[t]\g - \vectorToSkew[\wB[t]] \v[t] 
        & \v[t] - \vectorToSkew[\wB[t]] \p[t] 
        & -\vectorToSkew[\wB[t]]\d[t]
        & -\vectorToSkew[\wB[t]]\l[t] \\
        \zeros[1\times3] & 0 & 0 & 0 & 0  \\
        \zeros[1\times3] & 0 & 0 & 0 & 0  \\
        \zeros[1\times3] & 0 & 0 & 0 & 0  \\
        \zeros[1\times3] & 0 & 0 & 0 & 0 \\
        \end{bmatrix}$}
    \end{tabularx}
    \begin{tabularx}{1.0\textwidth}[t]{Xp{6.8cm}}
        \toprule
        \footnotesize Log-Linear Right-Invariant Dynamics & \footnotesize Log-Linear Left-Invariant Dynamics \\
        \midrule
        \scriptsize
        \resizebox{\linewidth}{!}{
        $\A[t][r] = 
        \begin{bmatrix}
        -\vectorToSkew[\wB[t]] & \zeros & \zeros & \zeros & \zeros & \I & \zeros \\
        -\vectorToSkew[\aB[t]] & -\vectorToSkew[\wB[t]] & \zeros & \zeros & \zeros & \zeros & \I \\
        \zeros & \I & -\vectorToSkew[\wB[t]] & \zeros & \zeros & \zeros & \zeros\\
        \zeros & \zeros & \zeros & -\vectorToSkew[\wB[t]] & \zeros & \zeros & \zeros\\
        \zeros & \zeros & \zeros & \zeros & -\vectorToSkew[\wB[t]] & \zeros & \zeros \\
        \zeros & \zeros & \zeros & \zeros & \zeros & \zeros & \zeros \\
        \zeros & \zeros & \zeros & \zeros & \zeros & \zeros & \zeros \\
        \end{bmatrix}$} 
        &
        \scriptsize
        $\A[t][l] = 
        \begin{bmatrix}
        \zeros & \zeros & \zeros & \zeros & \zeros & \RE[t] & \zeros \\
        \vectorToSkew[\g] & \zeros & \zeros & \zeros & \zeros & \vectorToSkew[\vE[t]]\RE[t] & \RE[t] \\
        \zeros & \I & \zeros & \zeros & \zeros &  \vectorToSkew[\pE[t]]\RE[t] & \zeros \\
        \zeros & \zeros & \zeros & \zeros & \zeros & \vectorToSkew[\dE[t]] \RE[t] & \zeros \\
        \zeros & \zeros & \zeros & \zeros & \zeros & \vectorToSkew[\lE[t]] \RE[t] & \zeros \\
        \zeros & \zeros & \zeros & \zeros & \zeros & \zeros & \zeros \\
        \zeros & \zeros & \zeros & \zeros & \zeros & \zeros & \zeros \\
        \end{bmatrix}$ 
        \\
        \scriptsize
        $\hat{\Q}^r_t = 
        \text{Cov}(\noise[t])$ 
        &
        \scriptsize
        $\hat{\Q}^l_t = 
        \begin{bmatrix}
        \Adjoint[\XE[t][-1]] & \zeros[15,6] \\
        \zeros[6,15] & \I[6]
        \end{bmatrix}
        \text{Cov}(\noise[t])
        \begin{bmatrix}
        \Adjoint[\XE[t][-1]] & \zeros[15,6] \\
        \zeros[6,15] & \I[6]
        \end{bmatrix}^\transpose$ \\ 
    \end{tabularx}
    \begin{tabularx}{1.0\textwidth}[t]{XXX}
        \toprule
        \footnotesize Measurement & \footnotesize Observation Matrix, $\H$  & \footnotesize Observation Type \\
        \midrule
    \footnotesize Forward Kinematic & \scriptsize $\begin{bmatrix} \zeros & \zeros & \I & -\I & \zeros & \zeros & \zeros \end{bmatrix}$ & \footnotesize Left-Invariant \\ 
    \footnotesize Relative Landmark Position & \scriptsize $\begin{bmatrix} \zeros & \zeros & \I & \zeros & -\I & \zeros & \zeros \end{bmatrix}$ & \footnotesize Left-Invariant \\
    \footnotesize Absolute Landmark Position & \scriptsize $\begin{bmatrix} -\vectorToSkew[\l] & \zeros & \I & \zeros & \zeros & \zeros & \zeros \end{bmatrix}$ & \footnotesize Left-Invariant \\
    \footnotesize Magnetometer & \scriptsize $\begin{bmatrix} -\vectorToSkew[\textbf{m}] & \zeros & \zeros & \zeros & \zeros & \zeros & \zeros \end{bmatrix}$ & \footnotesize Left-Invariant \\
    \footnotesize GPS Position & \scriptsize $\begin{bmatrix} \zeros & \zeros & -\I & \zeros & \zeros & \zeros & \zeros \end{bmatrix}$ & \footnotesize Right-Invariant \\
        \bottomrule
    \end{tabularx}
\end{table*} 

%% file: conclusion.tex
\section{Conclusion and future Work}
\label{sec:conclusion}

Using recent results on a group-invariant form of the extended Kalman filter (EKF), this article derived an observer for a contact-aided inertial navigation system for a 3D legged robot. Contact and IMU sensors are available on all modern bipedal robots; therefore, the developed system has the potential to become an essential part of such platforms since an observer with a large basin of attraction can improve the reliability of perception and control algorithms. We also included IMU biases in the state estimator and showed that, while some of the theoretical guarantees are lost, in real experiments, the proposed system has better convergence performance than that of a commonly used quaternion-based EKF. Although the latter is a discrete EKF on a Lie group, it does not exploit symmetries present in the system dynamics and observation models, namely, invariance of the estimation error under a group action. In addition to the original right-invariant form of the filter, the left-invariant dynamics are discussed as well as a robo-centric version of the filter.

A series of experiments were conducted while running the filter on a Cassie-series biped robot. The accuracy of the invariant-EKF was demonstrated through a motion capture comparison, while long-term odometry was compared by overlaying the trajectory on Google Earth imagery. In addition to improving signals used for feedback control, the pose estimate from this filter can be used alongside a vision sensor to build maps of the environment. This was demonstrated on Cassie by building LiDAR-based point cloud maps while walking. 

Future work includes developing an invariant smoother based on this filter. This would utilize the framework developed by \citet{chauchat2018invariant} to potentially improve \ac{IMU} preintegration \citep{lupton2012visual,forster2016manifold,eckenhoff2018closed} as well as contact preintegration \citep{hartley2018hybrid} to perform \ac{SLAM}. One interesting extension would be online estimation of kinematic parameters, which may help remove biases in the forward kinematic measurements. Another possibility for improvement of legged robot odometry is the detection of mode changes such as standing, flat-ground walking, and turning. During these modes, additional constraints may be inferred, which could improve state estimation \citep{brossard2019rins}. Finally, additional experiments need to be conducted to explore the potential for visual-inertial-contact odometry using an \ac{InEKF} as well as incorporating prior terrain information into the filter.

%% file: appendix.tex
\newpage
\FloatBarrier
\begin{appendices}

\section{Discretization of Filter Equations} \label{appx:discretization}
In the preceding sections, the filters equations were in continuous time. However, in order to implement these filters using software and physical sensors, these equations need to be discretized. For our implementation, we assumed a zero-order hold on the inertial measurements, and performed analytical integration \citep{eckenhoff2018closed,huai2018robocentric}. In particular, analytical integration was important for the resulting error dynamics to satisfy Theorem~\ref{theorem:log_linear_error}. 

\subsection{World-centric Dynamics}
This section demonstrates how to derive the deterministic, discrete time (world-centric) dynamics through analytical integration of the continuous state \eqref{eq:world_dynamics_with_bias} and bias \eqref{eq:bias_dynamics} dynamics. The contact, landmark, and bias dynamics are simply gaussian noise. Therefore, the discrete, deterministic dynamics are simply:
\begin{equation} 
  \begin{split}
    \dE[t_{k+1}] &= \dE[t_k], \qquad  
    \gyroscopeBiasE[t_{k+1}] = \gyroscopeBiasE[t_k], \qquad
    \accelerometerBiasE[t_{k+1}] = \accelerometerBiasE[t_k].
  \end{split} 
\end{equation}
Assuming a zero-order hold on the incoming IMU measurements between times $t_k$ and $t_{k+1}$, the orientation can be updated using the exponential map of $\SO(3)$:
\begin{equation*} 
      \RE[t_{k+1}] =  \int_{t_k}^{t_{k+1}} \RE[t_k] \vectorToSkew[\wB[t]] \, dt = \RE[t_k] \exp\left(\wB[t_k] \Delta t\right)
\end{equation*}
where $\Delta t \triangleq t_{k+1}-t_k$. Integrating the velocity dynamics yields an equation that involves the integral of the exponential map:
\begin{equation*}
    \vE[t_{k+1}] = \vE[t_k] + \int_{t_k}^{t_{k+1}} \RE[t] \aB[t] + \g \, dt = \vE[t_k] + \g \Delta t + \RE[t_k] \left( \int_{t_k}^{t_{k+1}}  \exp\left(\wB[t_k] t\right) \, dt \right) \aB[t_k]. \\
\end{equation*}
Likewise, analytically solving for the discrete position dynamics involves computing the double integral:
\begin{equation*}
    \pE[t_{k+1}] = \pE[t_k] + \vE[t_k] \Delta t + \dfrac{1}{2} \g \Delta t^2  + \RE[t_k] \left( \int_{t_k}^{t_{k+1}} \int_{t_k}^{\tau} \exp\left(\wB[t_k] t\right) \, dt\,d\tau \right) \aB[t_k]. \\
\end{equation*}
To solve these integrals, it is useful to define an auxiliary function \citep{bloesch2012state}:
\begin{equation}
  \Gam[m](\boldsymbol{\phi}) \triangleq \left(\sum_{n=0}^{\infty} \dfrac{1}{(n+m)!} \vectorToSkew[\boldsymbol{\phi}]^n \right),
\end{equation}
which allows integrals to be easily expressed and computed using the taylor series form of the $\SO(3)$ exponential map.
\begin{equation*}
  % \resizebox{\hsize}{!}{$
  \begin{split}
  \int_{t_k}^{t_{k+1}} &\exp\left( \w t \right) dt = \int_{t_k}^{t_{k+1}} \Gam[0](\wB[t_k] t) \; dt = \left(\sum_{n=0}^{\infty} \dfrac{1}{(n+1)!} \vectorToSkew[\wB[t_k] \Delta t]^n \right) \Delta t = \Gam[1](\w \Delta t) \Delta t \\ 
  \int_{t_k}^{t_{k+1}} \int_{t_k}^{\tau} &\exp\left(\wB[t_k] t\right) \, dt\,d\tau = \int_{t_k}^{t_{k+1}} \Gam[1](\wB[t_k] t) \, t \; dt = \left(\sum_{n=0}^{\infty} \dfrac{1}{(n+2)!} \vectorToSkew[\wB[t_k] \Delta t]^n \right) \Delta t^2 = \Gam[2](\w \Delta t) \Delta t^2
  \end{split}%$}
  \end{equation*}
Closed form expressions also exist, allowing fast and easy computation of these quantities~\citep{sola2017quaternion}.
\begin{equation}  
    \begin{split}
    \Gam[0](\boldsymbol{\phi}) &= \I[3] 
    + \dfrac{\sin(||\boldsymbol{\phi}||)}{||\boldsymbol{\phi}||} \vectorToSkew[\boldsymbol{\phi}]
    + \dfrac{1-\cos(||\boldsymbol{\phi}||)}{||\boldsymbol{\phi}||^2} \vectorToSkew[\boldsymbol{\phi}]^2 \\
    \Gam[1](\boldsymbol{\phi}) &= \I[3] 
    + \dfrac{1-\cos(||\boldsymbol{\phi}||)}{||\boldsymbol{\phi}||^2} \vectorToSkew[\boldsymbol{\phi}]
    + \dfrac{||\boldsymbol{\phi}||-\sin(||\boldsymbol{\phi}||)}{||\boldsymbol{\phi}||^3} \vectorToSkew[\boldsymbol{\phi}]^2 \\
    \Gam[2](\boldsymbol{\phi}) &= \dfrac{1}{2} \I[3] 
    + \dfrac{||\boldsymbol{\phi}||-\sin(||\boldsymbol{\phi}||)}{||\boldsymbol{\phi}||^3} \vectorToSkew[\boldsymbol{\phi}]
    + \dfrac{||\boldsymbol{\phi}||^2+2\cos(||\boldsymbol{\phi}||)-2}{2||\boldsymbol{\phi}||^4} \vectorToSkew[\boldsymbol{\phi}]^2 \\
    \end{split}
\end{equation}
\begin{remark}
$\Gam[0](\boldsymbol{\phi})$ is simply the exponential map of $\SO(3)$, while $\Gam[1](\boldsymbol{\phi})$ is also known as the left Jacobian of $\SO(3)$ \citep{chirikjian2011stochastic,barfoot2017state}. 
\end{remark}

Using these expressions, we can write down the discrete dynamics for the rotation, velocity, and positions states as:
\begin{equation}
\begin{split}
\RE[k+1] &= \RE[k] \; \Gam[0](\wB[k] \Delta t) \\
\vE[k+1] &= \vE[k] + \RE[k]\Gam[1](\wB[k] \Delta t)\aB[k] \Delta t + \g \Delta t \\
\pE[k+1] &= \pE[k] + \vE[k] \Delta t + \RE[k]\Gam[2](\wB[k] \Delta t)\aB[k] \Delta t^2 + \frac{1}{2} \g \Delta t^2, \\
\end{split}
\end{equation}
where the $t$ is dropped from the subscript for readability. These discrete dynamics are an exact integration of the continuous-time system under the assumption that the \ac{IMU} measurements are constant over $\Delta t$.

%%%%%%%%%%%%%%%%%%%%%%%%%%%%%%%%%%%%%%%%%%%%%%%5
% \subsection{Body-centric Dynamics}
% To compute the body-centric dynamics simply invert

% \begin{equation*}
%   \begin{split}
%   \RE[k+1] &= \Gam[0](-\wB[k] \Delta t) \; \RE[k]  \\
%   \vE[k+1] &= \Gam[0](-\wB[k] \Delta t) \left(\vE[k] - \Gam[1](\wB[k] \Delta t)\aB[k] \Delta t - \RE[k] \g \Delta t\right) \\
%   \pE[k+1] &= \Gam[0](-\wB[k] \Delta t) \Big(\pE[k] + \vE[k] \Delta t \\
%   &\qquad - \Gam[2](\wB[k] \Delta t)\aB[k] \Delta t^2 - \frac{1}{2} \RE[k] \g \Delta t^2\Big), \\
%   \end{split}
% \end{equation*}

%%%%%%%%%%%%%%%%%%%%%%%%%%%%%%%%%%%%%%%%%%%%%%%5
\subsection{Covariance Propagation}

In order to propagate the covariance, a continuous-time Riccati equation needs to be solved.
\begin{equation*}
  \deriv \P[t] = \A[t] \P[t] + \P[t] \A[t][\transpose] + \bar{\Q}_t
\end{equation*}
The analytical solution to the differential equation above is given by \citep{maybeck1982stochastic}:
\begin{equation}
  \P[t_{k+1}] = \stateTransition(t_{k+1},t_{k}) \P[t_{k}] \stateTransition(t_{k+1},t_{k})^\transpose + \bar{\Q}_d, \\
\end{equation} 
where the discrete noise covariance matrix is computed by
\begin{equation} \label{eq:discrete_noise_covariance}
  \bar{\Q}_d =  \int_{t_{k}}^{t_{k+1}} \stateTransition(t_{k+1},t) \bar{\Q}_t \stateTransition(t_{k+1},t)^\transpose dt,
\end{equation}
and the state transition matrix, $\stateTransition(t_{k+1},t_{k})$, satisfies 
\begin{equation} \label{eq:state_transition_ode}
  \deriv \stateTransition(t,t_{k}) = \A[t] \stateTransition(t,t_{k}) ~~\text{with}~~ \stateTransition(t_{k},t_{k}) = \I. 
\end{equation}

The (world-centric) left-invariant error dynamics matrix only depends on the IMU inputs and the estimated bias terms, see Table 2. Since both are assumed to be constant between times $t_k$ and $t_{k+1}$, the state transition matrix can be simply computed from the matrix exponential. 
\begin{equation}
  \stateTransition[][l](t_{k+1},t_{k}) = \exp_m(\A[t][l]\Delta t)
\end{equation}
This state transition matrix also has an analytical solution of the form:
\begin{equation}
  \stateTransition[][l](t_{k+1},t_{k}) = 
  \begin{bmatrix}
    \stateTransition[11][l] & \zeros & \zeros & \zeros & \stateTransition[15][l] & \zeros \\ 
    \stateTransition[21][l] & \stateTransition[22][l]  & \zeros & \zeros & \stateTransition[25][l] & \stateTransition[26][l]  \\ 
    \stateTransition[31][l] & \stateTransition[32][l]  & \stateTransition[33][l] & \zeros & \stateTransition[35][l] & \stateTransition[36][l] \\
    \zeros & \zeros & \zeros & \stateTransition[44][l] & \zeros & \zeros \\ 
    \zeros & \zeros & \zeros & \zeros & \I & \zeros \\ 
    \zeros & \zeros & \zeros & \zeros & \zeros & \I \\ 
  \end{bmatrix}
\end{equation}
where the individual terms are
\begingroup
\allowdisplaybreaks
\begin{flalign*}
  \stateTransition[11][l] &= \Gam[0][\transpose](\wB[k] \Delta t) \\
  \stateTransition[21][l] &= -\Gam[0][\transpose](\wB[k] \Delta t) \vectorToSkew[\Gam[1](\wB[k] \Delta t)\aB[k]] \Delta t \\
  \stateTransition[31][l] &= -\Gam[0][\transpose](\wB[k] \Delta t) \vectorToSkew[\Gam[2](\wB[k] \Delta t)\aB[k]] \Delta t^2 \\
  \stateTransition[22][l] &= \Gam[0][\transpose](\wB[k] \Delta t) \\
  \stateTransition[32][l] &= \Gam[0][\transpose](\wB[k] \Delta t) \Delta t \\
  \stateTransition[33][l] &= \Gam[0][\transpose](\wB[k] \Delta t) \\
  \stateTransition[44][l] &= \Gam[0][\transpose](\wB[k] \Delta t) \\
  \stateTransition[15][l] &= -\Gam[0][\transpose](\wB[k] \Delta t)\Gam[1](\wB[k] \Delta t) \Delta t \\
  \stateTransition[25][l] &=  \Gam[0][\transpose](\wB[k] \Delta t) \boldsymbol{\Psi_1} \\
  \stateTransition[35][l] &=  \Gam[0][\transpose](\wB[k] \Delta t) \boldsymbol{\Psi_2} \\
  \stateTransition[26][l] &= -\Gam[0][\transpose](\wB[k] \Delta t)\Gam[1](\wB[k] \Delta t) \Delta t \\
  \stateTransition[36][l] &= -\Gam[0][\transpose](\wB[k] \Delta t)\Gam[2](\wB[k] \Delta t) \Delta t^2. \\
\end{flalign*}
\endgroup
The matrices $\boldsymbol{\Psi_1}$ and $\boldsymbol{\Psi_2}$ involve computing the solution to a more complicated integral. However, these integrals still have analytical solutions which can be expressed easier after defining $\phi \triangleq \lVert \wB[k] \rVert$ and $\theta \triangleq \phi \Delta t$.
\begin{equation}
  \small
  \begin{split}
  \boldsymbol{\Psi_1} &\triangleq \int_{t_k}^{t_{k+1}} \vectorToSkew[\Gam[0](\wB[k] t) \aB[k]] \Gam[1](\wB[k] t) \, t \; dt \\
  &= \vectorToSkew[\aB[k]] \Gam[2](-\wB[k] \Delta t) \Delta t^2 \\
  &\quad \Big( \frac{\sin(\theta)-\theta\cos(\theta)}{\phi^3} \vectorToSkew[\wB[k]] \vectorToSkew[\aB[k]] \\ 
  &\quad+ \frac{\cos(2\theta)-4\cos(\theta)+3}{4\phi^4} \vectorToSkew[\wB[k]] \vectorToSkew[\aB[k]] \vectorToSkew[\wB[k]] \\ 
  &\quad+ \frac{4\sin(\theta)+\sin(2\theta)-4\theta\cos(\theta)-2\theta}{4\phi^5} \vectorToSkew[\wB[k]] \vectorToSkew[\aB[k]] \vectorToSkew[\wB[k]]^2 \\ 
  &\quad+ \frac{\theta^2-2\theta\sin(\theta)-2\cos(\theta)+2}{2\phi^4} \vectorToSkew[\wB[k]]^2 \vectorToSkew[\aB[k]] \\ 
  &\quad+ \frac{6\theta-8\sin(\theta)+\sin(2\theta)}{4\phi^5} \vectorToSkew[\wB[k]]^2 \vectorToSkew[\aB[k]] \vectorToSkew[\wB[k]] \\ 
  &\quad+ \frac{2\theta^2-4\theta\sin(\theta)-\cos(2\theta)+1}{4\phi^6} \vectorToSkew[\wB[k]]^2 \vectorToSkew[\aB[k]] \vectorToSkew[\wB[k]]^2 \Big)\\ 
  \end{split}
  \end{equation}
  \begin{equation}
  \small
  \begin{split}
  \boldsymbol{\Psi_2} &\triangleq \int_{t_k}^{t_{k+1}} \Gam[0](\wB[k] t) \stateTransition[25][l](t, t_k) \; dt \\
  &= \vectorToSkew[\aB[k]] \Gam[3](-\wB[k] \Delta t) \Delta t^3 \\
  & \Big( \frac{\theta\sin(\theta)+2\cos(\theta)-2}{\phi^4} \vectorToSkew[\wB[k]] \vectorToSkew[\aB[k]] \\ 
  &+ \frac{6\theta-8\sin(\theta)+sin(2\theta)}{8\phi^5} \vectorToSkew[\wB[k]] \vectorToSkew[\aB[k]] \vectorToSkew[\wB[k]] \\ 
  &+ \frac{2\theta^2+8\theta\sin(\theta)+16\cos(\theta)+\cos(2\theta)-17}{8\phi^6} \vectorToSkew[\wB[k]] \vectorToSkew[\aB[k]] \vectorToSkew[\wB[k]]^2 \\ 
  &+ \frac{\theta^3+6\theta-12\sin(\theta)+6\theta\cos(\theta)}{6\phi^5} \vectorToSkew[\wB[k]]^2 \vectorToSkew[\aB[k]] \\ 
  &+ \frac{6\theta^2+16\cos(\theta)-\cos(2\theta)-15}{8\phi^6} \vectorToSkew[\wB[k]]^2 \vectorToSkew[\aB[k]] \vectorToSkew[\wB[k]] \\ 
  &+ \frac{4\theta^3+6\theta-24\sin(\theta)-3\sin(2\theta)+24\theta\cos(\theta)}{24\phi^7} \vectorToSkew[\wB[k]]^2 \vectorToSkew[\aB[k]] \vectorToSkew[\wB[k]]^2 \Big)\\ 
  \end{split}
  \end{equation}

The (world-centric) right-invariant error dynamics matrix depends on the the state estimates, $\RE[t]$, $\vE[t]$, and $\pE[t]$, which will change between times $t_k$ and $t_{k+1}$, see Table 2. Therefore, the state transition matrix will not simply be the matrix exponential, as in the left-invariant case. Solving \eqref{eq:state_transition_ode} yields a state transition matrix of the form:
\begin{equation}
  \stateTransition[][r](t_{k+1},t_{k}) = 
  \begin{bmatrix}
    \I & \zeros & \zeros & \zeros & \stateTransition[15][r] & \zeros \\ 
    \stateTransition[21][r] & \I  & \zeros & \zeros & \stateTransition[25][r] & \stateTransition[26][r]  \\ 
    \stateTransition[31][r] & \stateTransition[32][r]  & \I & \zeros & \stateTransition[35][r] & \stateTransition[36][r] \\
    \zeros & \zeros & \zeros & \I & \stateTransition[45][r] & \zeros \\ 
    \zeros & \zeros & \zeros & \zeros & \I & \zeros \\ 
    \zeros & \zeros & \zeros & \zeros & \zeros & \I \\ 
  \end{bmatrix}
\end{equation}
where the individual terms can be analytically computed as
\begingroup
\allowdisplaybreaks
\begin{flalign*}
  \stateTransition[21][r] &= \vectorToSkew[\g] \Delta t\\
  \stateTransition[31][r] &= \frac{1}{2} \vectorToSkew[\g] \Delta t^2\\
  \stateTransition[32][r] &= \I \Delta t \\
  \stateTransition[15][r] &= -\RE[k] \Gam[1](\wB[k] \Delta t) \Delta t \\
  \stateTransition[25][r] &= -\vectorToSkew[\vE[k+1]]\RE[k]\Gam[1](\wB[k] \Delta t) \Delta t + \RE[k]\boldsymbol{\Psi_1} \\
  \stateTransition[35][r] &= -\vectorToSkew[\pE[k+1]]\RE[k]\Gam[1](\wB[k] \Delta t) \Delta t + \RE[k]\boldsymbol{\Psi_2} \\
  \stateTransition[45][r] &= -\vectorToSkew[\dE[k+1]]\RE[k]\Gam[1](\wB[k] \Delta t) \Delta t \\
  \stateTransition[26][r] &= -\RE[k] \Gam[1](\wB[k] \Delta t) \Delta t \\
  \stateTransition[36][r] &= -\RE[k] \Gam[2](\wB[k] \Delta t) \Delta t^2. \\
\end{flalign*}
\endgroup

Since the left/right-invariant errors are related through the adjoint, the two state transition and discrete noise matrices also satisfy the following relations \citep{barrau2015non}.
\begin{equation}
  \begin{split}
    \stateTransition[][r] &= \Adjoint[\XE[k+1]] \stateTransition[][l] \Adjoint[\XE[k][-1]] \\
    \bar{\Q}_d^r &= \Adjoint[\XE[k+1]] \bar{\Q}_d^l \Adjoint[\XE[k+1]]^\transpose \\
  \end{split}
\end{equation}
Therefore, the right-invariant state transition matrix can alternatively be computed using:
\begin{equation}
  \stateTransition[][r](t_{k+1},t_{k}) = \Adjoint[\XE[k+1]] \, \exp_m(\A[t][l]\Delta t) \, \Adjoint[\XE[k][-1]],
\end{equation}
which can simplify implementation since many software libraries already contain efficient methods to compute the matrix exponential.

Similar to the state transition matrices, the discrete noise covariance matrix \eqref{eq:discrete_noise_covariance} also has an analytical solution. In practice, this matrix is often approximated as:
\begin{equation}
  \bar{\Q}_d \approx \stateTransition \bar{\Q}_k \stateTransition[][\transpose] \Delta t .
\end{equation}
This approximated discrete noise matrix was used for all results in this article.

%%%%%%%%%%%%%%%%%%%%%%%%%%%%%%%%%%%%%%%%%%%%%%%5
%%%%%%%%%%%%%%%%%%%%%%%%%%%%%%%%%%%%%%%%%%%%%%%5
\section{Useful Lie Group Expressions}
\label{appx:formulas}

The matrix Lie group $\SEK(3)$ is known as the group of $K$ direct isometries \citep{barrau2015non}. This group is comprised of a rotation matrix, $\R\in\SO(3)$, and $K$ vectors in $\realnumbers^3$, $\p[1], \cdots, \p[K]$. Let $\X$ be an element of $\SEK(3)$, which can be written as a $(3+K)\times(3+K)$ matrix:
\begin{equation}
    \X \triangleq
    \begin{bmatrix}
        \R & \p[1] & \cdots & \p[K] \\
        \zeros[3,3] & 1 & \cdots & 0 \\
        \vdots & \vdots & \ddots & \vdots \\
        \zeros[3,3] & 0 & \cdots  & 1 \\
    \end{bmatrix}.
\end{equation}
The group action is matrix multiplication. The adjoint is a linear map that can be used to move vectors between the tangent spaces of two group elements. The matrix representation of the adjoint of $\SEK(3)$ is given by:
\begin{equation}
\Adjoint[\X] = 
\begin{bmatrix}
\R & \zeros & \cdots & \zeros \\
\vectorToSkew[\p[1]] \R & \R & \cdots & \zeros \\
\vdots & \vdots & \ddots & \vdots \\
\vectorToSkew[\p[K]] \R & \zeros & \cdots & \R \\
\end{bmatrix}.
\end{equation}
The matrix representation of a vector, $\tangentError \triangleq \vector[\boldsymbol{\phi}, \tangentError[1], \cdots, \tangentError[K]] \in \realnumbers^{3+3K}$, in the Lie algebra can obtained using the ``hat'' operator:
\begin{equation}
    \vectorToAlgebra[\tangentError] = 
    \begin{bmatrix}
       \vectorToSkew[\boldsymbol{\phi}] & \tangentError[1] & \cdots & \tangentError[K] \\
       \zeros[3,3] & 0 & \cdots & 0 \\
       \vdots & \vdots & \ddots & \vdots \\
       \zeros[3,3] & 0 & \cdots  & 0 \\
   \end{bmatrix} \in \sek(3),
\end{equation}
where $\vectorToSkew[\boldsymbol{\phi}]$ denotes the skew-symmetric matrix of a vector $\boldsymbol{\phi} = \vector[\boldsymbol{\phi}_1, \boldsymbol{\phi}_2, \boldsymbol{\phi}_3] \in \realnumbers^3$ .
\begin{equation}
    \vectorToSkew[\boldsymbol{\phi}] \triangleq
    \begin{bmatrix}
        0 & -\boldsymbol{\phi}_3 & \boldsymbol{\phi}_2 \\
        \boldsymbol{\phi}_3 & 0 & -\boldsymbol{\phi}_1 \\
        -\boldsymbol{\phi}_2 & \boldsymbol{\phi}_1 & 0
    \end{bmatrix} \in \mathfrak{so}(3)
\end{equation}
The same vector can be moved to the Lie group through the exponential map:
\begin{equation} \label{eq:exp_SEK3}
    \begin{split}
    \exp(\tangentError) &= 
    \begin{bmatrix}
        \Gam[0](\boldsymbol{\phi}) & \Gam[1](\boldsymbol{\phi}) \tangentError[1] & \cdots & \Gam[1](\boldsymbol{\phi}) \tangentError[K] \\
        \zeros[3,3] & 1 & \cdots & 0 \\
        \vdots & \vdots & \ddots & \vdots \\
        \zeros[3,3] & 0 & \cdots  & 1 \\
    \end{bmatrix},
    \end{split} 
\end{equation}
where $\Gam[0](\boldsymbol{\phi})$ is the exponential map of $\SO(3)$, and $\Gam[1](\boldsymbol{\phi})$ is the left Jacobian of $\SO(3)$.

%%%%%%%%%%%%%%%%%%%%%%%%%%%%%%%%%%%%%%%%5
\section{Error-state Conversions}  \label{appx:error_conversions}
Throughout this document, several versions of error states are used. These include the left/right invariant error~\eqref{eq:invariant_error}, the \ac{QEKF} error states~\eqref{eq:quaternion_error_states}, and the ``Euclidean orientation error'' used for plotting the covariance hull~\eqref{eq:euclidean_error}. It is often necessary to convert between these error states for plotting or when initializing the filters to provide fair comparisons. For example, if the \ac{QEKF} and the \ac{InEKF} are initialized with identical covariance matrices, the underlying distribution that they represent may be substantially different. This makes it impossible to show an accuracy comparison between the \ac{InEKF} and \ac{QEKF} with identical initial uncertainty. This section provides details on how to convert between these error states up to a first-order approximation.

When designing a \ac{QEKF}, the orientation error can be defined in either the local or global frame~\citep{sola2017quaternion}. These errors are equivalent to the left- and right-invariant errors for $\SO(3)$. In this document, the orientation error in the \ac{QEKF} was chosen to be the error defined in the local frame (left-invariant error). Since the invariant errors are related through the group's adjoint, the exact relation between right-invariant and \ac{QEKF} orientation errors is
\begin{equation}
  \exp(\tangentError[t][R]) = \exp(\RE[t] \delta \boldsymbol{\theta}_t).
\end{equation}
When plotting the individual axes of the orientation error covariance hull, a ``Euclidean orientation error'' should be used. Let $\delta \boldsymbol{\phi}_t \triangleq \boldsymbol{\phi}_t - \bar{\boldsymbol{\phi}}_t$ be this Euclidean error where $\exp(\boldsymbol{\phi}) \triangleq \R$ is the exponential coordinate representation of a particular orientation. When the errors are small, a first-order approximation can be used to find a mapping between the right-invariant orientation error and this ``Euclidean orientation error''. 
\begin{equation}
    \begin{split}
    \exp(\tangentError[t][R]) &= \RE[t] \R[t][\transpose] = \exp(\bar{\boldsymbol{\phi}}_t) \exp(-\bar{\boldsymbol{\phi}}_t-\delta\boldsymbol{\phi}) \\
    &\approx \exp(-\Gam[1](\bar{\boldsymbol{\phi}}_t)\delta\boldsymbol{\phi}) \\
    \end{split}
\end{equation}
A similar first-order approximation can be used to find the relation between the right-invariant error and the \ac{QEKF} velocity errors.
\begin{equation}
  \begin{split}
  \groupError[t][v] &= \vE[t] - \RE[t]\R[t][\transpose]\v[t] = \vE[t] - \exp(\tangentError[t][R])\v[t] \\
  &\approx \vE[t] - \v[t] - \vectorToSkew[\tangentError[t][R]]\v[t] = -\delta \v[t] + \vectorToSkew[\v[t]]\tangentError[t][R] \\
  \implies \tangentError[t][v] &\approx -\delta \v[t] + \vectorToSkew[\vE[t]]\tangentError[t][R] \\
  \end{split}
\end{equation}
The same process can be repeated for the position states.
\begin{equation}
  \begin{split}
    \tangentError[t][p] &\approx -\delta \p[t] + \vectorToSkew[\pE[t]]\tangentError[t][R] \\
    \tangentError[t][d] &\approx -\delta \d[t] + \vectorToSkew[\dE[t]]\tangentError[t][R].
\end{split}
\end{equation}

\end{appendices}